\documentclass{article}

\usepackage{microtype}
\usepackage{graphicx}
\usepackage{booktabs} 

\usepackage{xcolor}
\usepackage{hyperref}
\usepackage{bookmark}

\definecolor{linkcolor}{HTML}{1F77B4} 
\definecolor{citecolor}{HTML}{2CA02C} 
\definecolor{urlcolor}{HTML}{D62728}  

\hypersetup{
    colorlinks=true,
    linkcolor=linkcolor,
    citecolor=citecolor,
    urlcolor=urlcolor,
    bookmarksnumbered=true,
    bookmarksopen=true,
    bookmarksopenlevel=1,
    pdfpagemode=UseOutlines,
    pdfstartview=FitH,
    breaklinks=true,
    hyperindex=true,
    linktoc=all
}

\usepackage[accepted]{icml2024}

\usepackage{amsmath}
\usepackage{amssymb}
\usepackage{mathtools}
\usepackage{amsthm}

\usepackage[capitalize,noabbrev]{cleveref}

\usepackage{flushend}

\theoremstyle{plain}

\theoremstyle{definition}

\theoremstyle{remark}

\usepackage[textsize=tiny]{todonotes}

\usepackage{enumitem}
\usepackage[labelformat=simple]{subcaption}

\definecolor{redplot}{rgb}{0.81, 0.00, 0.00}
\definecolor{blueplot}{rgb}{0.157, 0.43, 0.678}
\definecolor{greenplot}{rgb}{0.32, 0.647, 0.223}

\icmltitlerunning{Approaching Deep Learning through the Spectral Dynamics of Weights}

\begin{document}

\twocolumn[
\icmltitle{Approaching Deep Learning through the Spectral Dynamics of Weights}



\icmlsetsymbol{equal}{*}

\begin{icmlauthorlist}
\icmlauthor{David Yunis}{ttic}
\icmlauthor{Kumar Kshitij Patel}{ttic}
\icmlauthor{Samuel Wheeler}{argonne}
\icmlauthor{Pedro Savarese}{ttic}\\
\icmlauthor{Gal Vardi}{weizmann}
\icmlauthor{Karen Livescu}{ttic}
\icmlauthor{Michael Maire}{uchicago}
\icmlauthor{Matthew R.~Walter}{ttic}
\end{icmlauthorlist}

\icmlaffiliation{ttic}{Toyota Technological Institute at Chicago, Chicago, IL, USA}
\icmlaffiliation{argonne}{Argonne National Laboratory, Lemont, IL, USA}
\icmlaffiliation{uchicago}{Department of Computer Science, University of Chicago, Chicago, IL, USA}
\icmlaffiliation{weizmann}{Weizmann Institute of Science, Israel, work done primarily at TTIC}

\icmlcorrespondingauthor{David Yunis}{dyunis@ttic.edu}

\icmlkeywords{Machine Learning, ICML}

\vskip 0.3in
]



\printAffiliationsAndNotice{}  

\begin{abstract}

We propose an empirical approach centered on the spectral dynamics of weights---the behavior of singular values and vectors during optimization---to unify and clarify several phenomena in deep learning. We identify a consistent bias in optimization across various experiments, from small-scale ``grokking'' to large-scale tasks like image classification with ConvNets, image generation with UNets, speech recognition with LSTMs, and language modeling with Transformers. We also demonstrate that weight decay enhances this bias beyond its role as a norm regularizer, even in practical systems. Moreover, we show that these spectral dynamics distinguish memorizing networks from generalizing ones, offering a novel perspective on this longstanding conundrum. Additionally, we leverage spectral dynamics to explore the emergence of well-performing sparse subnetworks (lottery tickets) and the structure of the loss surface through linear mode connectivity. Our findings suggest that spectral dynamics provide a coherent framework to better understand the behavior of neural networks across diverse settings.
\end{abstract}

\section{Introduction}

Interest in neural networks has exploded in the past decade. Capabilities are rapidly improving, and deployment is ever-increasing. Yet, although issues with these technologies now have social repercussions~\citep{bender2021dangers, bommasani2021opportunities}, many fundamental questions regarding their behavior remain unanswered.

\begin{figure}[t]
  \centering
  \begin{subfigure}[b]{0.60\columnwidth}
    \centering
    \includegraphics[width=\columnwidth]{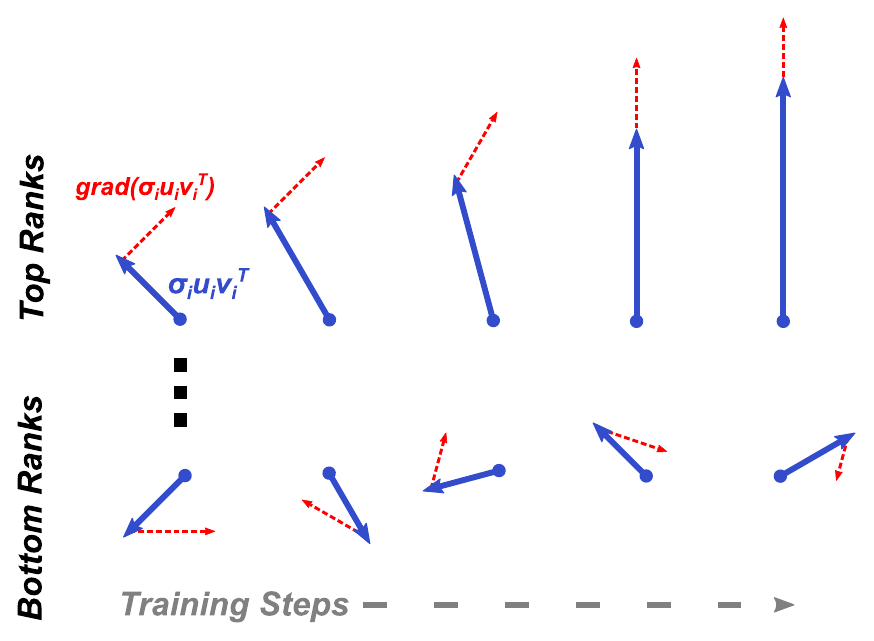}
    \caption{Schematic}
  \end{subfigure}
  \begin{subfigure}[b]{0.39\columnwidth}
    \centering
    \includegraphics[width=\columnwidth]{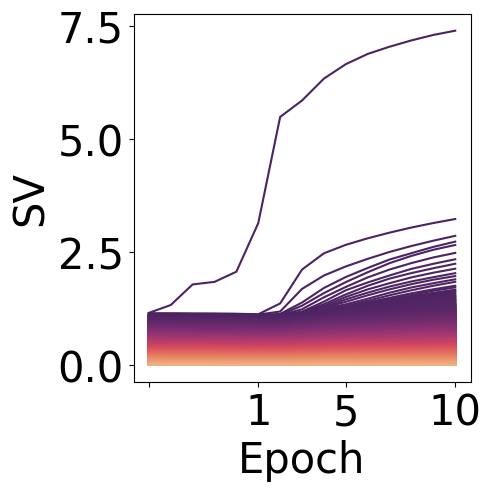}
    \caption{SVs}
  \end{subfigure}
  \caption{\textbf{Left:} Schematic for the spectral dynamics of a weight matrix. As training proceeds top singular vectors become stable and top singular values grow disproportionately large. \textbf{Right:} Singular value evolution for a single matrix in a Transformer, where each line is a single singular value and color represents rank. We see a disproportionate trend where large singular values grow larger faster. We explore these spectral dynamics of weights and connect them to generalization, regularization, and seemingly unrelated phenomena like linear mode connectivity.}\label{fig:teaser}
\end{figure}

For instance, despite extensive research, we still lack a complete understanding of the implicit biases~\citep{neyshabur2014search} of neural networks trained via stochastic optimization. Even basic questions regarding the role of regularization like weight decay~\citep{hanson1988comparing, krogh1991simple, zhang2018three} have only partial answers~\citep{van2017l2, andriushchenko2023we, yaras2023law}. Perhaps most vexing, we lack a complete explanation for how neural networks generalize, despite having the capacity to perfectly memorize the training data~\citep{zhang2021understanding}. Such an explanation may allow us to design better algorithms, however a lack of understanding makes the deployment of neural networks vulnerable to uninterpretable errors across fields~\citep{szegedy2013intriguing, ilyas2019adversarial, hendrycks2021natural, zou2023universal}.

Although theoretical explanations have been put forward to make analysis tractable, these studies often rely on special settings like deep linear networks~\citep{arora2018optimization, arora2019implicit} or infinite-width systems~\citep{jacot2018neural}, and arguments may rely on unsubstantiated or impractical assumptions like near zero initialization. On the empirical side, a growing body of work in interpretability has attempted to reverse-engineer neural networks~\citep{rahaman2019spectral, barak2022hidden, nanda2023progress}, but given the difficulty of the tasks, the systems of interest have been very small-scale, and the methodology for analysis quite bespoke and difficult to scale. A third category of work aims at understanding empirical behavior from a higher level~\citep{zhang2021understanding, huh2022low, yu2023compressing}, but while these works often study larger-scale systems, they often focus on more abstract objects like the gram matrix~\citep{huh2022low} or NTK~\citep{fort2020deep}, and thus do not have the granularity and predictive power of the previous two categories.

To bridge these gaps, we propose a task-agnostic, unifying perspective of many disparate phenomena in deep learning across many different practical tasks and architectures, including image classification with ConvNets, image generation with UNets, speech recognition with LSTMs and language modeling with Transformers. Through extensive experiments, we examine the dynamics of singular values and singular vectors of weight matrices and the spectral dynamics of weights. We show that such dynamics underlie many distinct phenomena and appear intimately tied to generalization. We are motivated to study these dynamics specifically as optimization is the fundamental process driving deep learning~\citep{nagarajan2019uniform, zhang2021understanding}, the matrix parameters form the core identity of any neural network, and the singular value decomposition is a fundamental way to view any matrix. We detail our specific contributions in the following paragraphs.

As a test bed for understanding generalization, \citet{power2022grokking} introduce the ``grokking'' phenomenon, where a small-scale model initially minimizes the training loss but performs poorly on validation data, then with much more training suddenly minimizes the validation loss. In particular, \citet{nanda2023progress} showed that in simple modular arithmetic tasks, the specific solution learned by optimization could be reverse-engineered from the weight matrices. Although this description is precise, in Section~\ref{sec:grokking}, we notice a task-agnostic view of grokking, observing that the drop in validation loss during grokking coincides with the simultaneous discovery of low-rank solutions across all weight matrices in the network. The connection between rank and generalization might intuitively be through Occam's Razor. Furthermore, echoing existing works~\cite{lyu2023dichotomy,liu2023omnigrok}, we find that weight decay clearly affects grokking and rank minimization: neither behavior occurs as strongly without it. Having said that, we demonstrate that increasing the training data can partially compensate for the absence of weight decay in terms of generalization and rank minimization, and in all cases, we see a correlation between low-rank matrices and generalization. Thus, we find that examining spectral dynamics provides a task-agnostic view of grokking.

Though this suggests a connection between rank and generalization, grokking is typically studied on synthetic tasks, very small-scale models like single-layer Transformers or small MLPs, and requires very particular hyperparameter settings~\citep{power2022grokking, nanda2023progress, gromov2023grokking, kumar2023grokking}. If our perspective is to be useful, it needs to scale to larger systems. Thus, we turn to common empirical tasks drawn from the literature like image classification, image generation, speech recognition and language modeling as well as varied and larger networks like VGG~\citep{simonyan2014very}, UNet~\citep{ronneberger2015u}, LSTM~\citep{hochreiter1997long} and Transformers~\citep{vaswani2017attention}.

In Section~\ref{sec:spectral_dynamics}, we demonstrate that the spectral dynamics are biased toward effective rank minimization across various practical neural networks in complex settings. Although this behavior echoes theoretical predictions in the deep linear setting, we find that the behavior of networks disagrees with a common theoretical assumption about low-rank dynamics: alignment of singular vectors in consecutive layers~\citep{saxe2014exact, arora2018optimization, arora2019implicit, milanesi2021implicit}. Thus, the rank minimization mechanism may differ from what the theory describes. It is notable too that our hyperparameter settings are drawn from existing literature, thus the trend toward rank minimization coincides with well-generalizing networks across settings.

Still, in these more practical settings, we do not see such an obvious rank minimization as in grokking, nor is there a sudden transition from memorization to generalization. One particularly notable ingredient for grokking was a very high level of weight decay. Weight decay has a long history as a regularizer explicitly penalizing parameter norm, which can be used for norm-based generalization bounds~\citep{bartlett1996valid}, but these bounds do not seem to explain the success of practical systems~\citep{nagarajan2019uniform, jiang2019fantastic}. Recently, others have suggested an implicit effect of weight decay on parameter rank in theoretical or small-scale empirical studies~\citep{galanti2022sgd, timor2023implicit} which may be connected to generalization~\citep{razin2020implicit}. As such, we explore this effect of weight decay in practical settings.

In Section~\ref{sec:weight-decay}, we empirically connect rank minimization to weight decay, showing that weight decay promotes rank minimization across architectures and tasks. In addition, in some cases it also appears to promote singular vector alignment in consecutive weights despite the nonlinearities between layers, which indicates further compression of the model. Although weight decay explicitly penalizes norm, studying spectral dynamics allows us to observe the implicit effect on rank. Such an effect can help in understanding generalization as norm regularization is insufficient.

Given the suggestive connection between rank and generalization, we turn to the classic memorization experiments of \citet{zhang2021understanding}. \citet{zhang2021understanding} demonstrated that even small networks can memorize random labels. Hence, any arguments about generalization need to take into account the structure of the data and the optimization process. In Section~\ref{sec:random_labels}, we show that training with random labels leads to high-rank solutions, while rank with true labels is much lower. We also find that while random labels do not align consecutive layers, true labels do, which is surprising given the non-linearities between layers. Through spectral dynamics, we see a clear difference between the optimization of generalizing and memorizing networks, which provides a foothold in the climb toward better theoretical understanding.

Our results suggest that viewing neural networks through the lens of spectral dynamics can shed light on several generalization-related phenomena, but we suspect there are broader connections. In the literature, many curious and unexplained phenomena regarding neural networks exist. We take two as case studies. First, the lottery ticket hypothesis (LTH)~\citep{frankle2018lottery}, which has found the existence of sparse sub-networks with similar performance. Such a phenomenon provides evidence that, despite ever-increasing parameter counts and energy costs, efficient smaller networks already exist. Understanding the source of such efficiency may help us alleviate deployment costs. Second, linear mode connectivity (LMC)~\citep{nagarajan2019uniform, frankle2020linear, neyshabur2020being}, which finds that models sharing a portion of the optimization trajectory can be averaged together in weight-space to yield a stronger model~\citep{wortsman2022model, ramesh2022hierarchical}. This phenomenon indicates that, after some training, the loss surface is quite convex in a subspace, even though the optimization problem is theoretically extremely nonconvex. As any finetuning from pre-trained models stays in this convex space~\citep{neyshabur2020being, li2022branch, sadrtdinov2023stay}, an explanation for what underlies model-averaging would help to clarify the role of pretraining, would shed light on the commonly-used low-rank adaptation~\citep{hu2021lora}, and could lead to better optimization.

In Section~\ref{sec:beyond-generalization}, we find that global magnitude pruning, a standard procedure for finding lottery tickets, preserves top singular vectors and acts like a low-rank pruning. We also see that the ability to interpolate between models in LMC strongly correlates with sharing top singular vectors. With these results, we note that the two phenomena can be seen as aspects of the spectral dynamics of weights and bring them under the umbrella of prior sections.

To summarize the discussion above, by studying the spectral dynamics of weights, we find:
\begin{itemize}
    \item Grokking is intimately linked to rank minimization;
    \item Rank minimization is a general phenomenon in more complex tasks;
    \item Weight decay acts implicitly as a low-rank regularizer;
    \item Generalizing solutions have a lower rank than memorizing ones; and
    \item Top singular vectors are preserved when performing magnitude pruning while linearly interpolating between connected modes.
\end{itemize}

All of these phenomena and effects have previously been studied in isolation to varying degrees, but by approaching deep learning through spectral dynamics, we aim at a common language for neural networks. Code for all experiments is released at \url{https://github.com/dyunis/spectral_dynamics}.

\section{Related Work}

\subsection{Grokking}

\citet{power2022grokking} first noticed a surprising phenomenon they called ``grokking'' where models quickly fit the training data on toy tasks, then after a long period of training, very quickly generalize on the validation data. Later, others found that this phenomenon can occur in a relaxed fashion~\citep{thilak2022slingshot} on very simple models and different datasets~\citep{liu2022towards, gromov2023grokking, kumar2023grokking, xu2023benign} and that weight decay seems critical to cause it~\citep{lyu2023dichotomy, liu2023omnigrok, tan2023understanding}. Some posit a transition from the kernel~\citep{jacot2018neural} to rich~\citep{atanasov2023onset} regime explains grokking~\citep{kumar2023grokking, mohamadi2023grokking}. There is also empirical evidence for a connection between double descent and grokking~\citep{davies2022unifying} the discovery of a sparse solution~\citep{merrill2023tale}, the simplification of decision boundaries~\citep{humayun2024deep} and the leading indicator of loss oscillation~\citep{thilak2022slingshot, notsawo2023predicting}. None of these works have explicitly examined the connection with rank, which we do in Section~\ref{sec:grokking}, and provides a common framework through which to view many of these results.

\subsection{Singular Value Dynamics}

Prior work on deep linear networks~\cite{arora2019implicit, milanesi2021implicit} suggests that rank minimization may describe implicit regularization in deep matrix factorization better than simple matrix norms. See \cite{arora2018optimization} (Appendix~A) for a detailed argument. However, a critical assumption used in these works is ``balanced initialization.'' This means that for consecutive matrices $W_i$ and $W_{i+1}$ in the product matrix $\prod_j W_j$, we have $W^\top_{i+1}W_{i+1} = W_iW_i^\top$ at initialization. Decomposing these matrices with SVDs and leveraging orthogonality, this simplifies to $V_{i+1}\Sigma_{i+1}^2V_{i+1}^\top = U_i\Sigma_i^2U_i^\top$ where $U_i$ and $V_{i+1}$ are orthogonal matrices. Since these are orthogonal decompositions of the same matrix, their diagonals must be equivalent, allowing for the permutation of elements with the same value. This leads to $U_i = V_{i+1} O$ up to signs, where $O$ is a block diagonal permutation matrix that may permute the rows of equivalent diagonal elements. Notably, if all diagonal elements are distinct and $U_i$ and $V_{i+1}$ are square matrices, then $U_i = V_{i+1}$ up to signs.

Under the balanced initialization assumption, all product matrices will be aligned. Consequently, the product of the diagonals will evolve in a closed-form manner, with larger singular values growing faster than smaller ones. As shown by~\cite{arora2019implicit}, this translates to rank-minimizing behavior with increasing depth in the matrix products. This formula is also empirically validated for linear matrix factorization problems. Similar results have been derived for tensor products and other structured settings~\cite{saxe2014exact, yaras2023invariant}. More generally, \citep{ji2019gradient} show that for deep linear networks with infinite training alignment between layers will happen. In Section~\ref{sec:spectral_dynamics}, we explore how these conclusions and assumptions hold for much larger, practical neural networks that are far from linear.

\subsection{Low-Rank Properties}\label{sec:low_rank}

Another line of research focuses on more general low-rank biases. Early work explored norms as an implicit bias~\citep{gunasekar2017implicit}. Theoretical analyses reveal that norms or closed-form functions of weights might be insufficient to explain implicit regularization, but they do not necessarily contradict the possibility of rank minimization~\citep{razin2020implicit, vardi2021implicit}. Numerous studies investigate low-rank biases in various matrices, including the Jacobian~\citep{pennington2018emergence}, weight matrices~\citep{le2021training, martin2020heavy, martin2021implicit, frei2022implicit, ongie2022role}, Gram matrix~\citep{huh2022low}, and features~\citep{yu2023compressing, feng2022rank}. Additionally, research suggests that dynamics influence the decay of rank~\citep{li2020towards, chen2023stochastic, wang2023implicit}. Some works establish connections between weight decay and rank minimization in idealized settings~\citep{ziyin2022exact, galanti2022sgd, zangrando2024neural, ergen2023path, parhi2023deep, shenouda2023vector}. We are particularly interested in how far these connections extend in practice. In Section~\ref{sec:weight-decay}, we present empirical evidence that sometimes disagrees with, but also expands, arguments from theory and small-scale systems to much larger ones.

\section{Grokking and Rank Minimization}\label{sec:grokking}

Motivated by theoretical work that proposes connections between rank and generalization~\citep{razin2020implicit} weight decay and rank~\citep{galanti2022sgd, timor2023implicit, yaras2023law, zangrando2024neural}, and the importance of weight decay for grokking~\citep{power2022grokking, lyu2023dichotomy, liu2023omnigrok} in simple settings, we evaluate the potential connection between rank and grokking in neural networks. Examining grokking through the lens of rank (and more generally spectral dynamics) offers a complementary perspective on grokking with other descriptions such as fourier decomposition~\citep{nanda2023progress}, the simplification of linear decision boundaries~\citep{humayun2024deep}, the connection to double descent~\citep{davies2022unifying}, and the discovery of a sparse solution~\citep{merrill2023tale}.

\begin{figure*}[!h]
  \centering
  \begin{subfigure}[b]{0.24\linewidth}
    \centering
    \includegraphics[width=\linewidth]{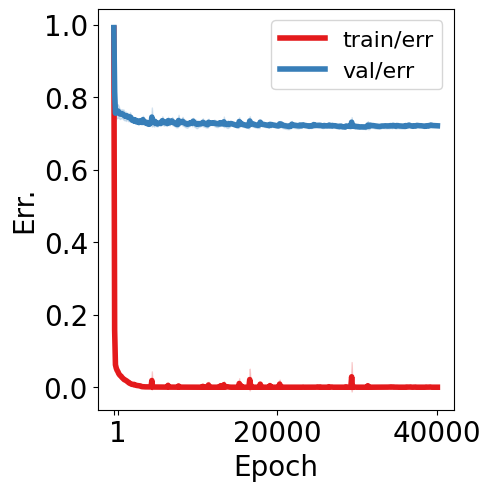}
    \\
    \includegraphics[width=\linewidth]{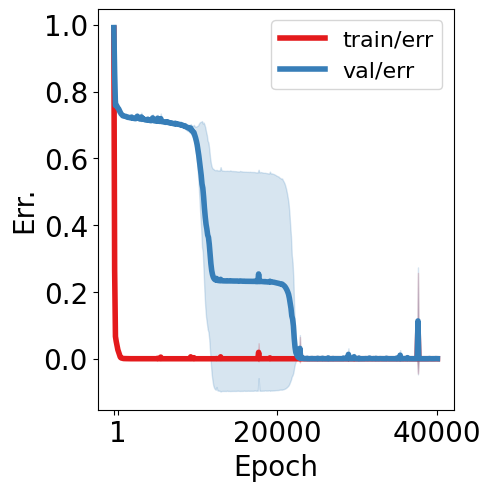}
    \\
    \includegraphics[width=\linewidth]{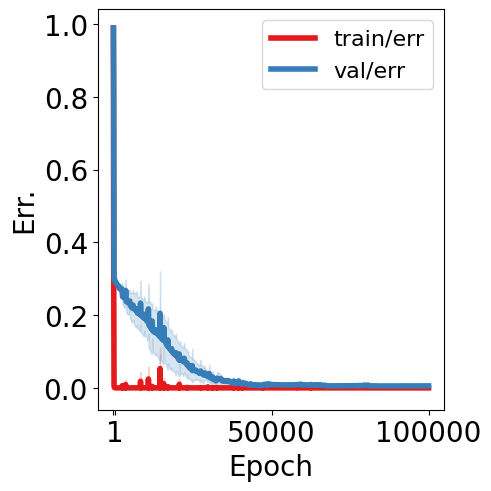}
    \\
    \includegraphics[width=\linewidth]{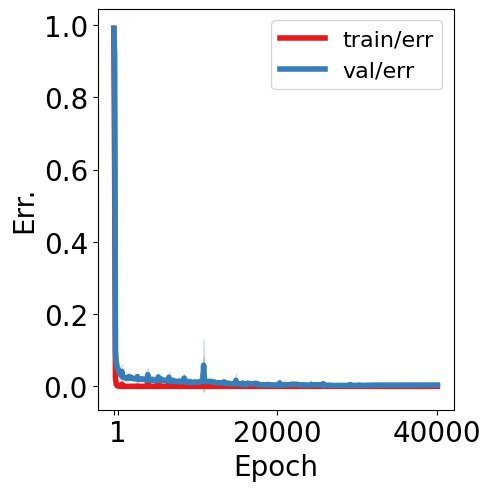}
    \caption{Error}
  \end{subfigure}\hfil
  \begin{subfigure}[b]{0.24\linewidth}
    \centering
    \includegraphics[width=\linewidth]{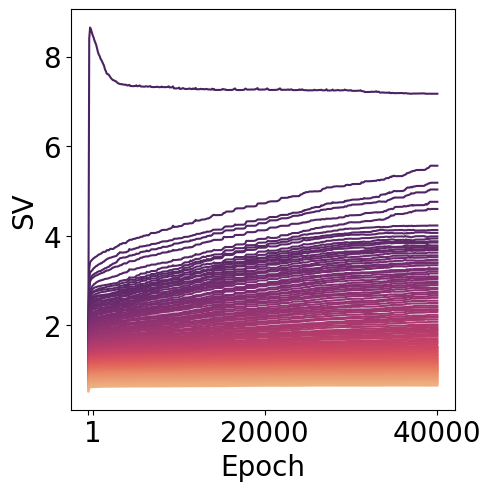}
    \\
    \includegraphics[width=\linewidth]{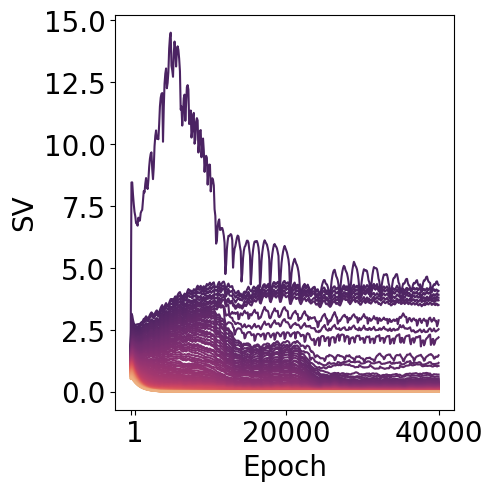}
    \\
    \includegraphics[width=\linewidth]{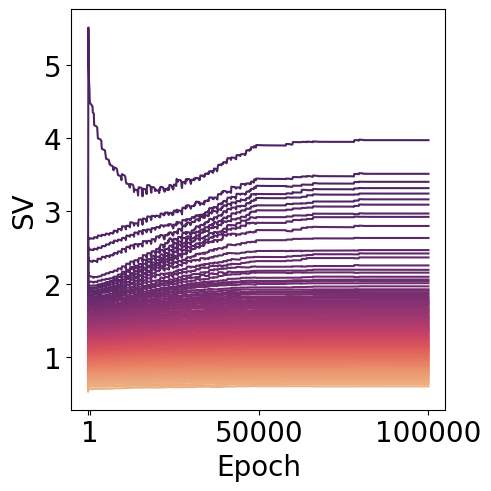}
    \includegraphics[width=\linewidth]{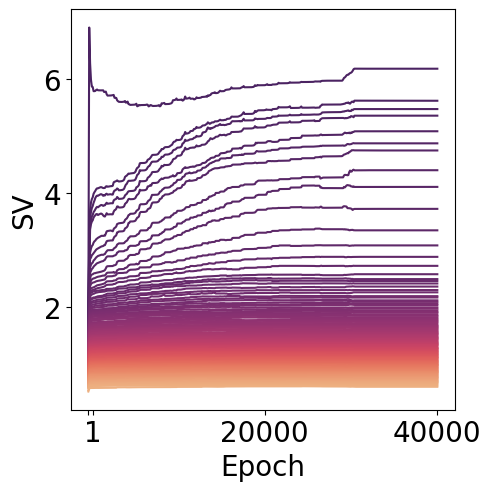}
    \caption{SV Evolution}
  \end{subfigure}\hfil
  \begin{subfigure}[b]{0.24\linewidth}
    \centering
    \includegraphics[width=\linewidth]{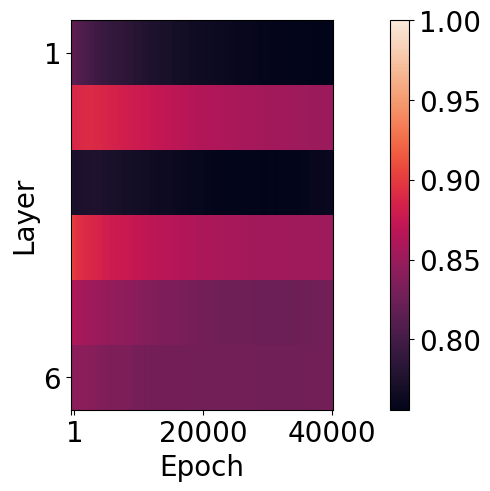}
    \\
    \includegraphics[width=\linewidth]{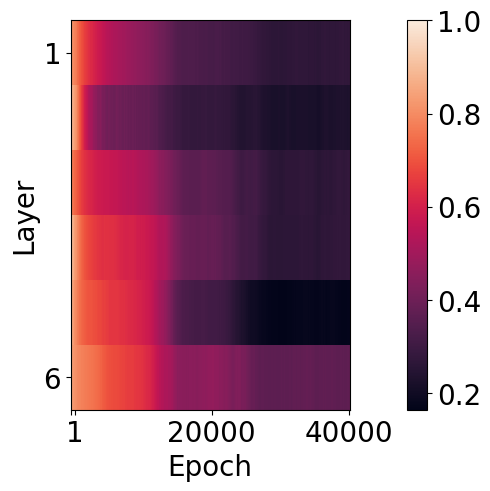}
    \\
    \includegraphics[width=\linewidth]{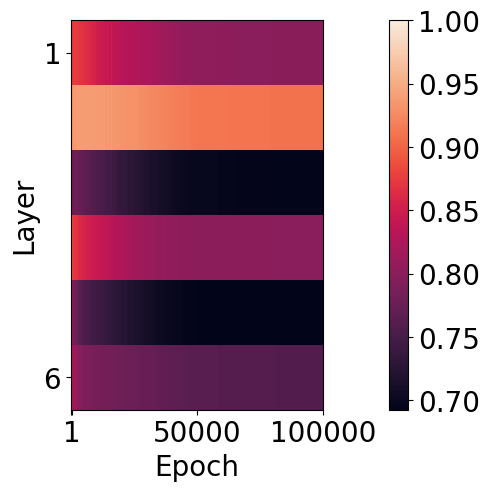}
    \\
    \includegraphics[width=\linewidth]{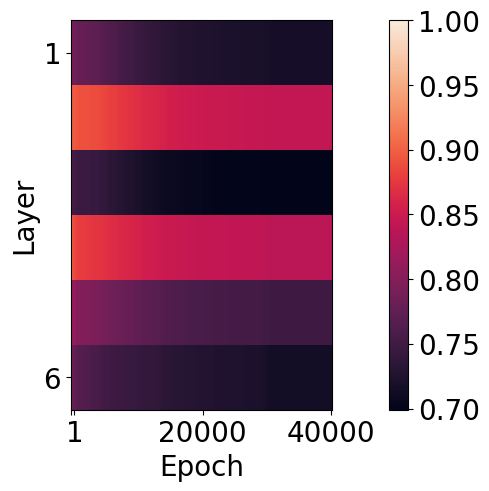}
    \caption{Effective Rank}
  \end{subfigure}\hfil
  \begin{subfigure}[b]{0.24\linewidth}
    \centering
    \includegraphics[width=\linewidth]{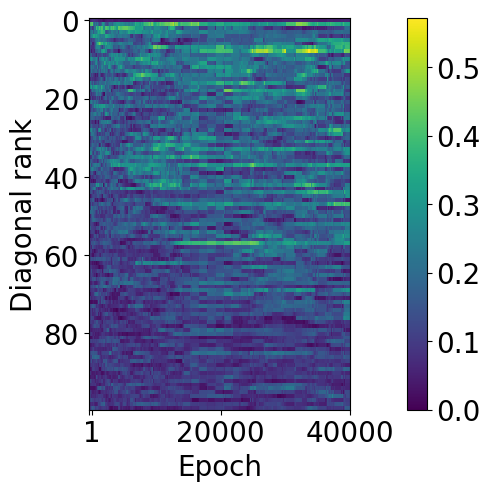}
    \\
    \includegraphics[width=\linewidth]{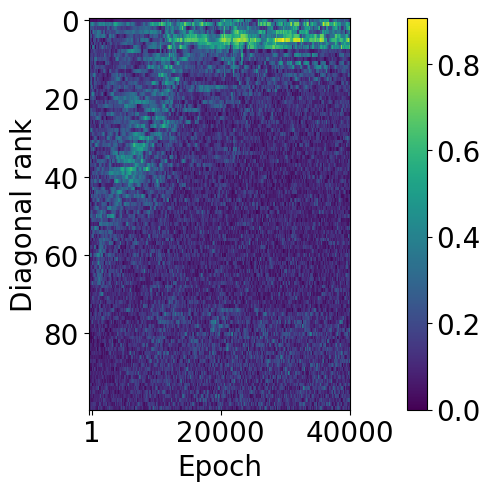}
    \\
    \includegraphics[width=\linewidth]{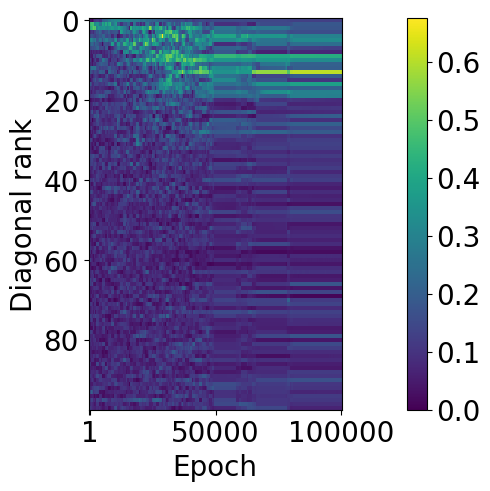}
    \\
    \includegraphics[width=\linewidth]{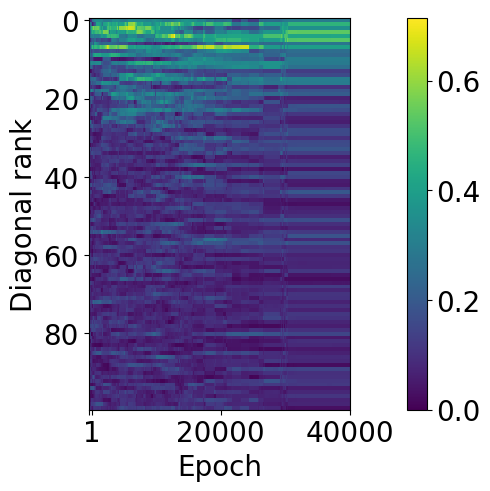}
    \caption{Alignment}
  \end{subfigure}
  \caption{\textbf{Grokking and Spectral Dynamics.} \textbf{Top row:} 30\% data and no weight decay. \textbf{2nd row:} 30\% data and weight decay 1.0 (grokking), using hyperparameters from \citet{nanda2023progress}. \textbf{3rd row:} 70\% data with no weight decay (slingshot), using hyperparameters from \citet{thilak2022slingshot}. \textbf{Bottom row:} 90\% data and no weight decay. \textbf{2nd column:} Singular value evolution is visualized for the first attention parameter, where each line represents a single singular value and the color represents the rank. \textbf{4th column:} Alignment (Eqn.~\ref{eqn:alignment-matrix}) between the embedding and the first attention parameter is also visualized, where the y-axis corresponds to index $i$ of the diagonal. \textbf{3rd column:} One can see that grokking co-occurs with low-rank weights (effective rank is Eqn.~\ref{eqn:normalized-effective-rank}). In addition, there is an alignment that begins early in training that evolves up the diagonal. Without weight decay and with less data, neither grokking nor the other phenomena occur during the entire training budget, but using more data, even without weight decay, leads to low-rank solutions from the beginning of training. The slingshot case follows a similar trend, though the validation loss is gradually fit. Across cases with good generalization, parameters are lower rank, and alignment is also more prevalent in the top ranks.}
  \label{fig:grokking}
\end{figure*}

We largely follow the setting of \citet{nanda2023progress}, optimizing a single-layer Transformer for modular addition (details in Appendix~\ref{app:experimental-details}). Inspired by work in the deep linear case~\citep{saxe2014exact, arora2019implicit, milanesi2021implicit, yaras2023law}, we track the evolution of singular values for individual weight matrices. To gain a high-level overview of all parameter evolutions, we compute the (normalized) effective rank of a matrix $W$~\citep{roy2007effective} with rank $R$ as
\begin{align}\label{eqn:normalized-effective-rank}
    & \text{EffRank}(W) := -\sum_{i=1}^R \frac{\sigma_i}{\sum_j \sigma_j} \log \frac{\sigma_i}{\sum_j \sigma_j}\enspace,\\
    & \text{NormEffRank}(W) := \frac{\text{EffRank}(W)}{R}\enspace,
\end{align}
where $\sigma_i$'s are the singular values of matrix $W$ and $\text{EffRank}(W)$ is the entropy of the normalized singular value distribution. As the probability mass concentrates, the effective rank decreases. We plot $\text{NormEffRank}(W)$ to compare across layers and time.

In addition, inspired by the assumptions of balancedness made by prior work~\citep{arora2018optimization, arora2019implicit}, we examine the alignment of consecutive weight matrices in the Transformer. To examine and quantify this alignment between consecutive matrices in a network at training time $t$, i.e., 
\begin{align*}
 W_i &= \sum_{j=1}^R \sigma_j(t) u_j(t) v_j(t)^\top\enspace,\\
 W_{i+1} &= \sum_{k=1}^R \sigma'_k(t) u'_k(t) {v'_k}(t)^\top\enspace, 
\end{align*}
we compute,
\begin{equation} \label{eqn:alignment-matrix}
    A(t)_{jk} = \lvert \langle u_{j}(t), v'_{k}(t) \rangle \rvert,
\end{equation}
where the absolute value is taken to ignore sign flips in the SVD computation. We then plot the diagonal of this matrix $A(t)_{ii}~\forall~i \leq 100$ over time. For exact details on how alignment is computed for different architectures and layers that are more complex than the fully connected case, please see Appendix~\ref{app:experimental-details}.

In Figure~\ref{fig:grokking}, we see a tight alignment: the sudden drop in validation loss coincides precisely with the onset of low-rank behavior in the singular values. Examining inter-layer alignment during training, we observe that the final low-rank solution gradually emerges from the model's middle ranks. Conversely, the grokking phenomenon is absent without weight decay, and no low-rank solution seems to develop. We also replicate the setting of \citet{thilak2022slingshot} who show a form of grokking without weight decay, but when plotted on a linear-scale x-axis the generalization appears much less sudden than the setting with weight decay. Additionally, when using 90\% of the data and no weight decay, generalization still coincides with effective rank minimization. The familiar reader will note that \citet{nanda2023progress} previously showed that the particular solution found in modular addition is a low rank fourier decomposition, so our observations on low rank weights will directly follow. While such a description for modular addition is impressively precise, it may be difficult to obtain for more complex tasks. In following sections we argue that rank minimization is a perspective that can apply in more complex settings when one does not know what to look for in the weights, and it may even be possible to eventually interpret the neural network via the top ranks~\citep{praggastis2022svd}.

We also want to highlight a complication between norm and generalization that our results indicate, in dialogue with \citet{liu2023omnigrok}. One can see that without weight decay but with 90\% of the data, the generalizing solution that is discovered has many singular values larger than 1, and the maximum singular value is also larger than that of the grokked solution, for which most singular values are near zero. Recall the norm of matrix $W$ is given by $\lVert W \rVert_2 = \sqrt{\sum_i \sigma_i^2}$ where $\sigma_i$ are the singular values. Thus, the large-data solution has a much higher norm than the grokked solution. Still, both settings generalize with a certain low-rank behavior. The confounding factor is that, with high weight decay, the smaller singular values disappear, while without, they do not. We currently lack a precise understanding of how low-rank behavior directly relates to generalization, and it might not always hold true – high-rank solutions might also generalize. Nonetheless, from Occam's Razor's perspective, a low-rank solution is conceptually ``simpler'' due to its lower dimensionality, making for a compelling explanation.

\section{Spectral Dynamics Across Tasks}\label{sec:spectral_dynamics}

Inspired by the results on grokking and prior work on deep linear networks, which studies the evolution of the SVD of the weight matrices~\citep{saxe2014exact, arora2018optimization, arora2019implicit, milanesi2021implicit, yaras2023invariant}, we apply the same analysis to larger, more practical systems. We show that the trends we saw in the analysis of grokking mostly hold true across networks and tasks at a much larger scale, even though our findings do occasionally deviate from theoretical predictions.

\subsection{Methodology}

Our experiments aim to examine reasonably sized neural networks across a variety of tasks. We select models and tasks that are representative of current applications. Specifically, we focus on:
\begin{itemize}
    \item \textbf{Image classification} with CNNs (VGG-16~\cite{simonyan2014very}) on CIFAR10~\cite{krizhevsky2009learning};
    \item \textbf{Image generation} through diffusion with UNets~\cite{ronneberger2015u} on MNIST~\cite{lecun1998mnist};
    \item \textbf{Speech recognition} with LSTMs~\cite{hochreiter1997long} on LibriSpeech~\cite{panayotov2015librispeech}; and
    \item \textbf{Language modeling} with Transformers~\cite{vaswani2017attention} on Wikitext-103~\cite{merity2016pointer}.
\end{itemize}

Training hundreds of runs for each of the above experiments is computationally expensive, limiting the scale of models we can explore. We primarily adopt hyperparameters from existing literature, with minor modifications for simplicity. This ensures that any correlations observed are likely a reflection of common practices, not introduced bias on our part. We hope that the broad scope of these experiments will allow for a more general perspective on neural network optimization.

The primary evidence in this section comes from computing the SVDs of weight matrices within the models. Consequently, we disregard 1D bias and normalization parameters in our analysis. However, previous research suggests that in some cases these parameters are not crucial for performance~\cite {zhang2018fixup,mohan2019robust,karras2023analyzing}. Due to the large number of matrices in these models, we present plots of individual layers' matrix parameters and statistics summarizing behavior across layers for conciseness of presentation. Hundreds of thousands of plots were generated for this study, making it impossible to include them all.  Full experimental details, including hyperparameters, are available in Appendix~\ref{app:experimental-details}, where we take hyperparameters from existing settings in the literature.

\subsection{Effective Rank Minimization}

\begin{figure}[!t]
  \centering
  \begin{subfigure}[b]{0.24\columnwidth}
    \centering
    \includegraphics[width=\columnwidth]{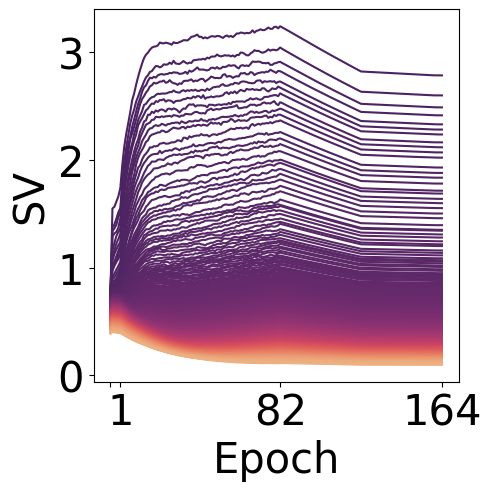}
    \\
    \includegraphics[width=\columnwidth]{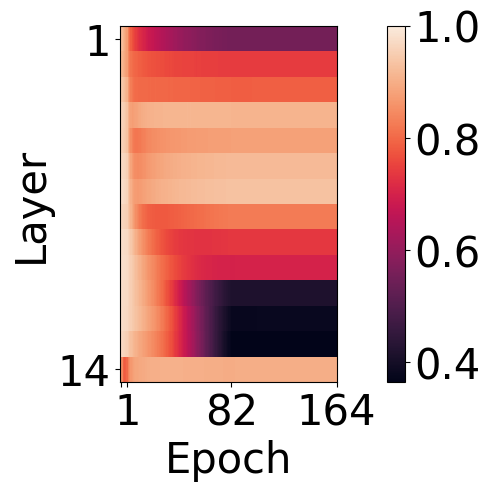}
    \caption{VGG}
  \end{subfigure}
  \begin{subfigure}[b]{0.24\columnwidth}
    \centering
    \includegraphics[width=\columnwidth]{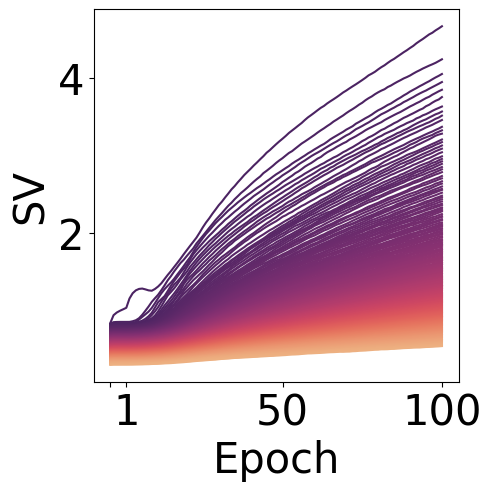}
    \\
    \includegraphics[width=\columnwidth]{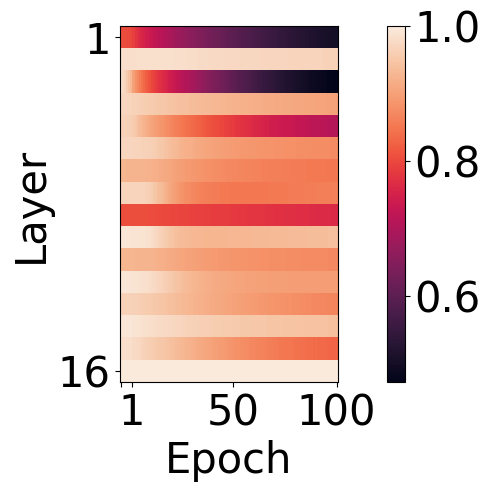}
    \caption{UNet}
  \end{subfigure}
  \begin{subfigure}[b]{0.24\columnwidth}
    \centering
    \includegraphics[width=\columnwidth]{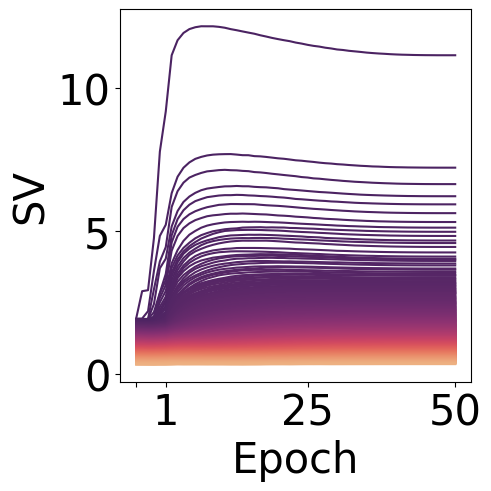}
    \\
    \includegraphics[width=\columnwidth]{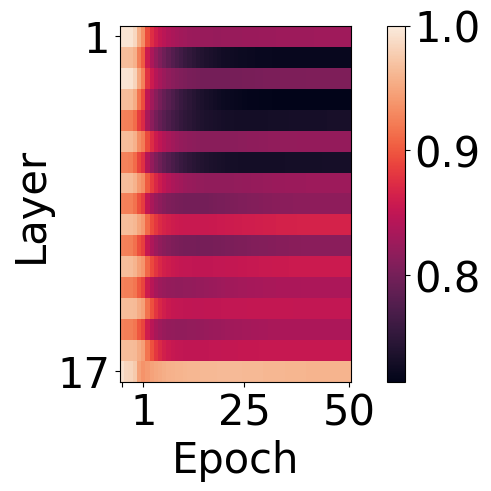}
    \caption{LSTM}
  \end{subfigure}
  \begin{subfigure}[b]{0.24\columnwidth}
    \centering
    \includegraphics[width=\columnwidth]{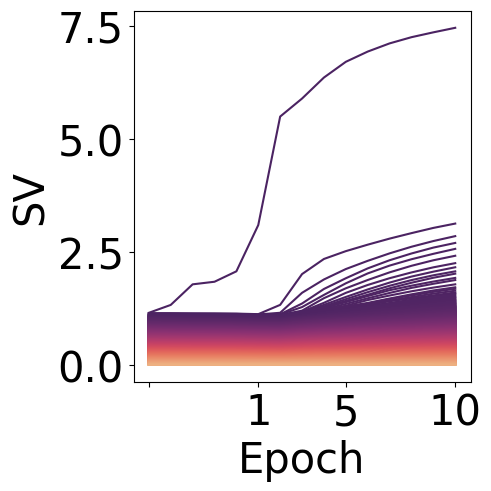}
    \\
    \includegraphics[width=\columnwidth]{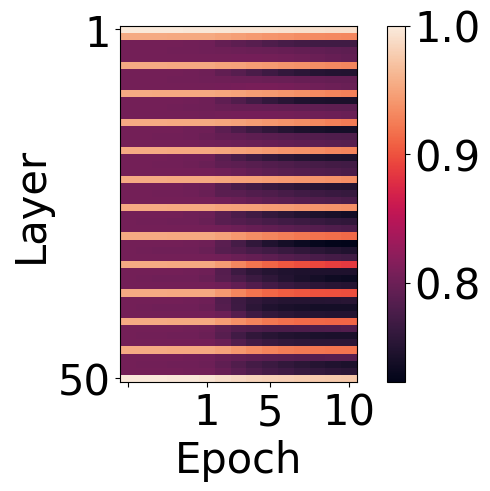}
    \caption{Transformer}
  \end{subfigure}
  \caption{\textbf{Top row:} Singular value evolution for a single matrix in the middle of each model. Each line represents a singular value, whereas color represents rank. Notice the unequal evolution where top singular values grow at a disproportionate rate. \textbf{Bottom row:} Normalized effective rank (Eqn.~\ref{eqn:normalized-effective-rank}) evolution visualized in color for different matrices across architectures and time. As we move down the $y$-axis, the depth of the parameters in the model increases, while the $x$-axis tracks training time. Notice decreasing effective rank across nearly all parameters, though the magnitude differs across layers. The block-like patterns in the VGG case are likely due to different channel dimension sizes. The banding in the UNet, LSTM, and Transformer cases is due to the differences between convolutional and linear layers, residual block connections, and attention and fully connected layers, respectively. The sharp transition midway through training in the VGG case is likely due to a 10$\times$ learning rate decay.}
  \label{fig:normalized-effective-rank}
\end{figure}

Building on theoretical~\citep{saxe2014exact, arora2019implicit, milanesi2021implicit, boix2023transformers, yaras2023invariant} and empirical~\citep{dittmer2019singular, martin2020heavy, martin2021implicit, boix2023transformers} findings, we investigate effective rank minimization across parameters in larger models and on a more diverse variety of tasks. Figure~\ref{fig:normalized-effective-rank} reveals a consistent trend: the effective rank of network parameters generally decreases throughout training, regardless of the specific parameter or network architecture. This suggests a progressive ``simplification'' of the network as training progresses, and echoes our previous findings in the high-data regime on modular addition.

\begin{figure}[!t]
  \centering
  \begin{subfigure}[b]{0.24\columnwidth}
    \centering
    \includegraphics[width=\columnwidth]{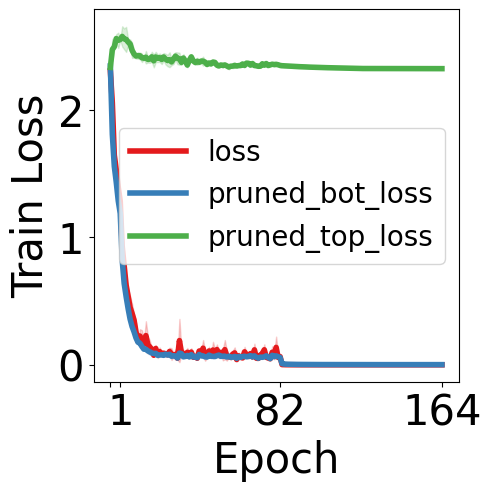}
    \\
    \includegraphics[width=\columnwidth]{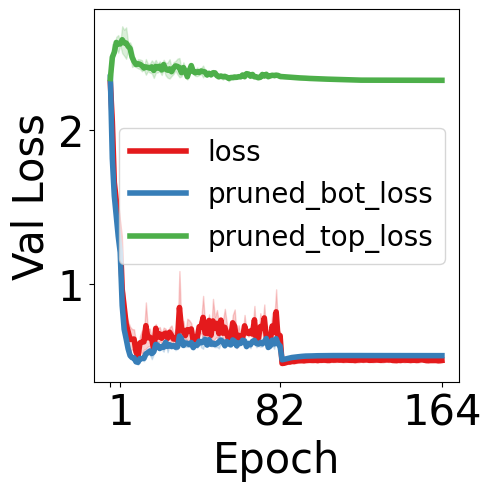}
    \caption{VGG}
  \end{subfigure}
  \begin{subfigure}[b]{0.24\columnwidth}
    \centering
    \includegraphics[width=\columnwidth]{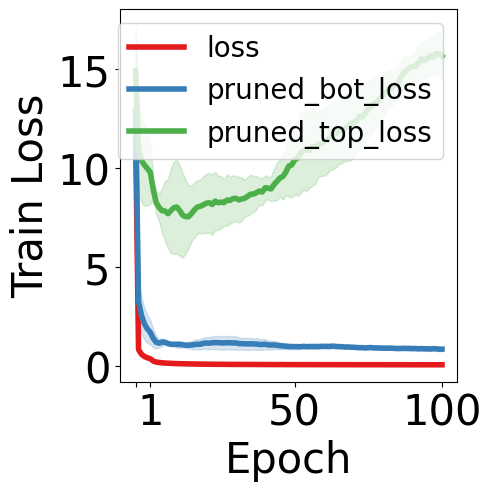}
    \\
    \includegraphics[width=\columnwidth]{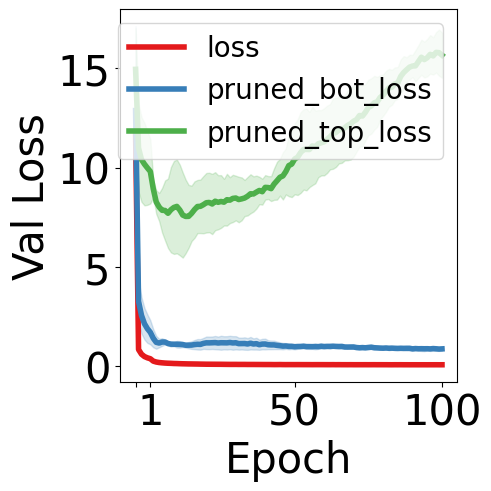}
    \caption{UNet}
  \end{subfigure}
  \begin{subfigure}[b]{0.24\columnwidth}
    \centering
    \includegraphics[width=\columnwidth]{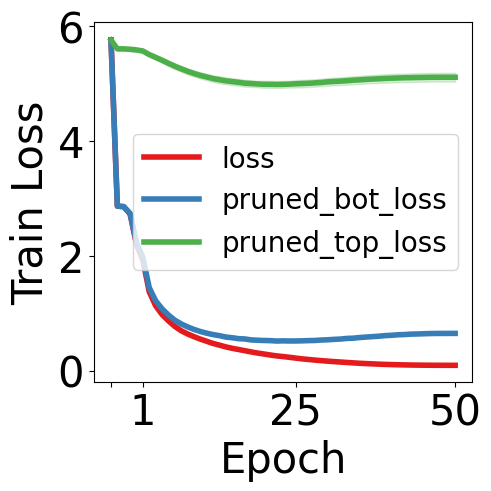}
    \\
    \includegraphics[width=\columnwidth]{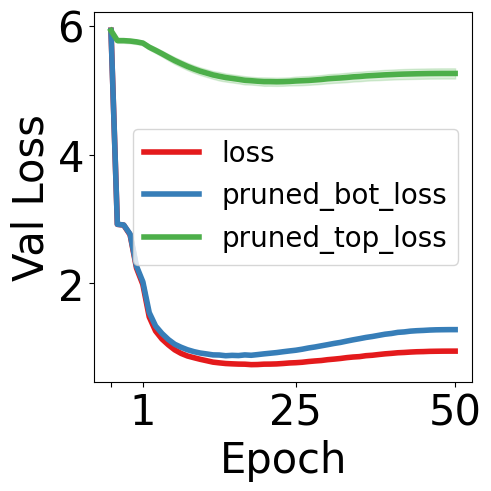}
    \caption{LSTM}
  \end{subfigure}
  \begin{subfigure}[b]{0.24\columnwidth}
    \centering
    \includegraphics[width=\columnwidth]{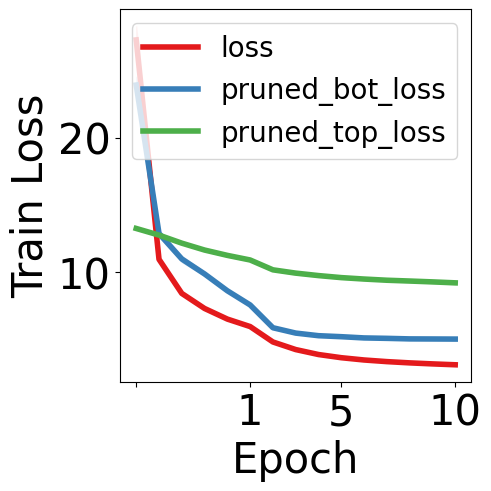}
    \\
    \includegraphics[width=\columnwidth]{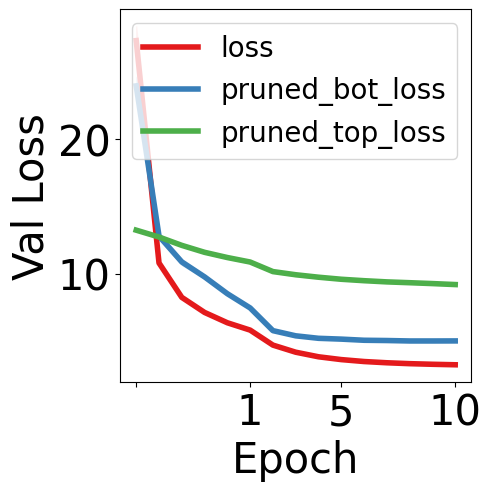}
    \caption{Transformer}
  \end{subfigure}
  \caption{\textbf{Top row:} Training losses for all tasks. \textbf{Bottom row:} Validation losses for all tasks. \textcolor{redplot}{Red} is the full model. \textcolor{blueplot}{Blue} is post-training pruning the bottom half of the SVD for every matrix in the model that is not the final layer. \textcolor{greenplot}{Green} is post-training pruning the top half of the SVD. Notice that for all models, keeping the top half of the SVD is close to the full model performance, supporting the idea that the top directions provide a better approximation to the function.}
  \label{fig:pruned-performance}
\end{figure}

We further conduct a singular-value pruning experiment to explore the relationship between low-rank behavior and model performance. We prune either the top or bottom half of the singular values for each weight matrix in the network and then evaluate the pruned model at each training step. Intuitively, we expect the top singular values to capture the most critical information for the network's function. Figure~\ref{fig:pruned-performance} confirms this hypothesis, demonstrating that the pruned parameters, without further training, can closely approximate the full model's performance. This may not have been the case. In particular, simultaneously pruning lower components across all layers may lead to losing some critical signal that must be passed between layers, or the large number of small magnitude singular values may provide some important noising effect, thus empirical confirmation is necessary. Another way to argue this is that, though each individual matrix is well-approximated in euclidean space by its top ranks, it is unclear the total effect on the neural network function from pruning these matrices simultaneously. We will rely in later sections on this observation that large singular values are more important to the function of the network.

\subsection{Alignment of Singular Vectors Between Layers}

Similar to the analysis of grokking, we investigate the alignment between consecutive layers in the larger neural networks considered in this section. We not only employ the alignment matrix defined in Eqn.~\ref{eqn:alignment-matrix} but also derive and plot a scalar measure for alignment based on the top diagonal entries:
\begin{equation}
\label{eqn:alignment-measure}
a(t) = \frac{1}{10} \sum_{i=1}^{10} A(t)_{ii}
\end{equation}
For specific details on calculating this measure in diverse architectures and complex layers (beyond fully connected layers), please refer to Appendix~\ref{app:experimental-details}.

\begin{figure}[!t]
  \centering
  \begin{subfigure}[b]{0.24\columnwidth}
    \centering
    \includegraphics[width=\columnwidth]{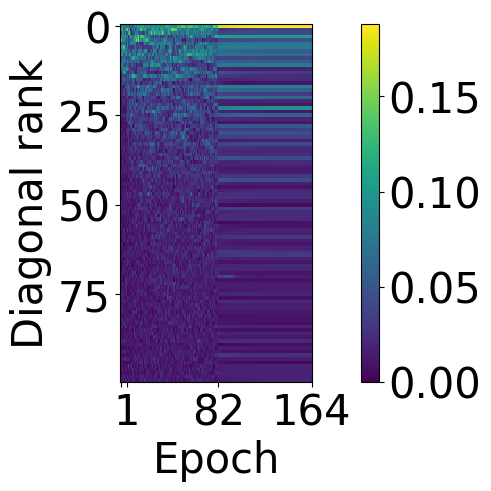}
    \\
    \includegraphics[width=\columnwidth]{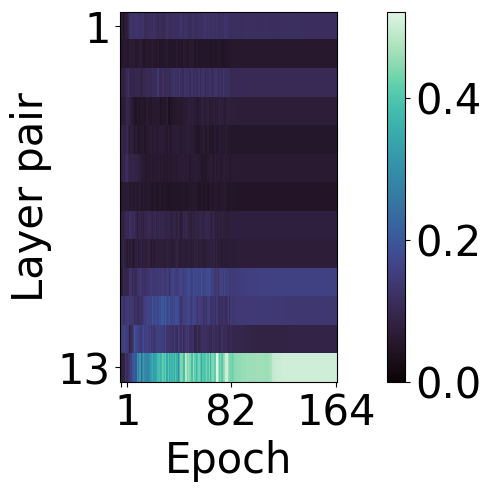}
    \caption{VGG}
  \end{subfigure}
  \begin{subfigure}[b]{0.24\columnwidth}
    \centering
    \includegraphics[width=\columnwidth]{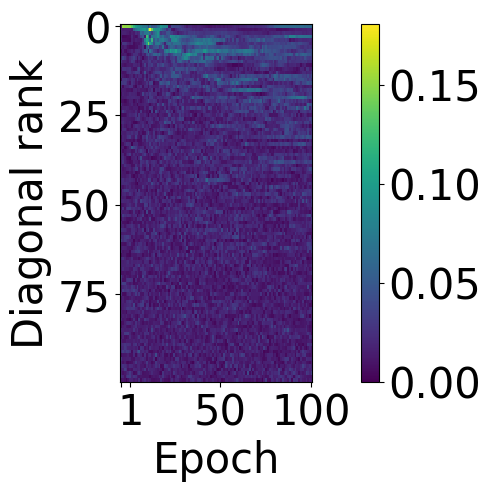}
    \\
    \includegraphics[width=\columnwidth]{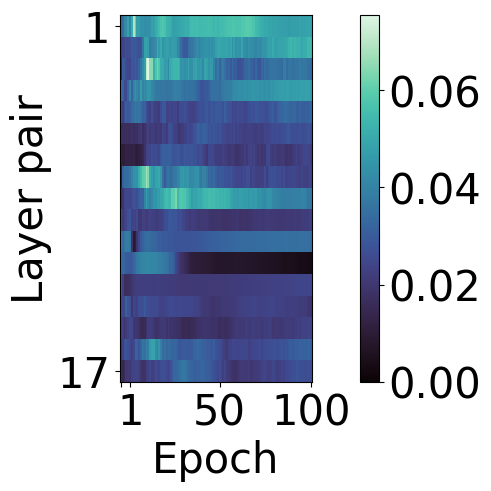}
    \caption{UNet}
  \end{subfigure}
  \begin{subfigure}[b]{0.24\columnwidth}
    \centering
    \includegraphics[width=\columnwidth]{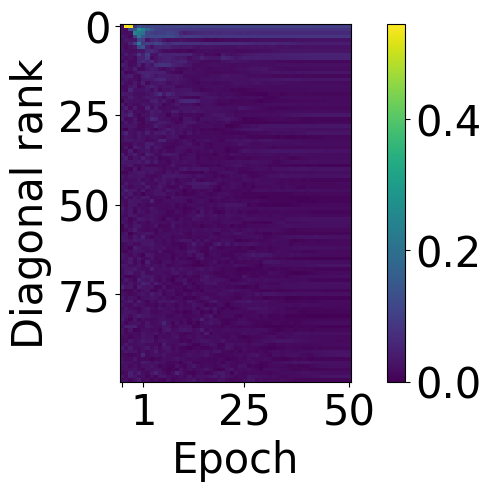}
    \\
    \includegraphics[width=\columnwidth]{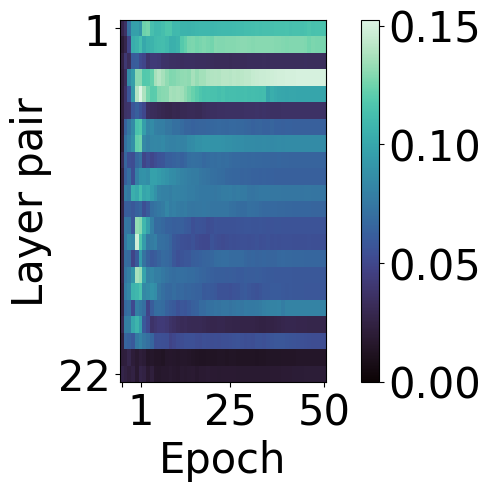}
    \caption{LSTM}
  \end{subfigure}
  \begin{subfigure}[b]{0.24\columnwidth}
    \centering
    \includegraphics[width=\columnwidth]{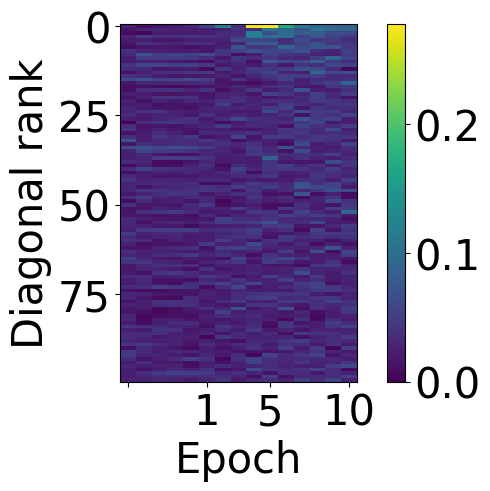}
    \\
    \includegraphics[width=\columnwidth]{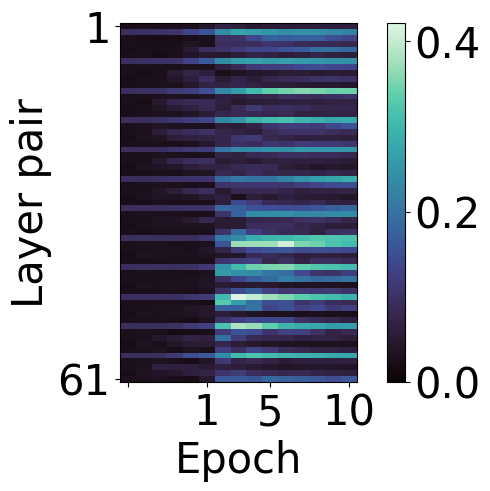}
    \caption{Transformer}
  \end{subfigure}
  \caption{Neighboring layer alignment of singular vectors. \textbf{Top row:} 
  The diagonal of the alignment matrix $A(t)_{ii}$ (Eqn.~\ref{eqn:alignment-matrix}) vs.\ training time for a single pair of matrices in the middle of each model. We see a small amount of alignment in the top ranks between layers shortly after training begins, but this becomes more diffuse over time. \textbf{Bottom row:} Alignment metric (Eqn.~\ref{eqn:alignment-measure}) for pairs of matrices for depth vs.\ training time. It is hard to make out a global trend across models, though the LSTM shows a weak signal around Epoch 1 when the initial alignment occurs, and the Transformer case has a banding pattern with depth due to alignment between the query and key matrices that have no nonlinearity in between.}
  \label{fig:layer-alignment}
\end{figure}

Figure~\ref{fig:layer-alignment} reveals a key finding: the theoretical assumption of \textbf{balanced initialization}, which posits aligned singular value decompositions (SVDs) between weight matrices~\citep{arora2018optimization, saxe2014exact}, does not hold true at the start of training in these larger networks. Additionally, unlike the linear case discussed in \citet{du2018algorithmic}, the alignment does not appear to remain static throughout training. However, a weak signal of alignment  in the top ranks develops. This trend is somewhat reminiscent of the theoretical result provided by \citet{mulayoff2020unique} for linear networks under the assumption of whitened input data. Still, the weakness of the observed signal means that existing theoretical models do not fully capture the complexities of real-world neural network training.

\section{The Effect of Weight Decay}\label{sec:weight-decay}

In light of the previously observed evolution of singular values, we investigate a proposed effect of weight decay. Though weight decay explicitly penalizes the norm of weights, there is evidence that complicates the connection between norm and generalization for neural networks~\citep{razin2020implicit, andriushchenko2023we}, meaning we do not have a full understanding as to why weight decay may be useful. Alternatively, some theoretical~\citep{boix2023transformers, razin2020implicit, yaras2023invariant, timor2023implicit, ongie2022role, galanti2022sgd, zangrando2024neural} and empirical works~\citep{galanti2022sgd, boix2023transformers} propose a connection with the rank of matrices in constrained settings. Still, a comprehensive connection to larger empirical networks has not yet been demonstrated.

\begin{figure*}[!t]
  \centering
  \begin{subfigure}[b]{0.24\linewidth}
    \centering
    \includegraphics[width=0.5\linewidth]{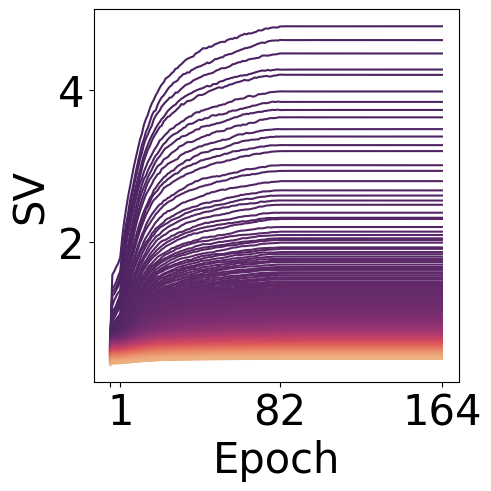}\hfil
    \includegraphics[width=0.5\linewidth]{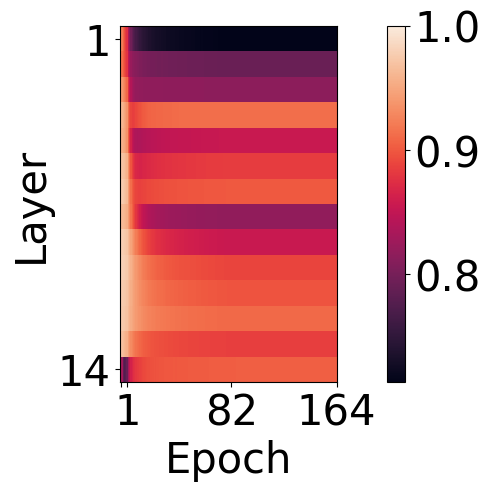}
    \\
    \includegraphics[width=0.5\linewidth]{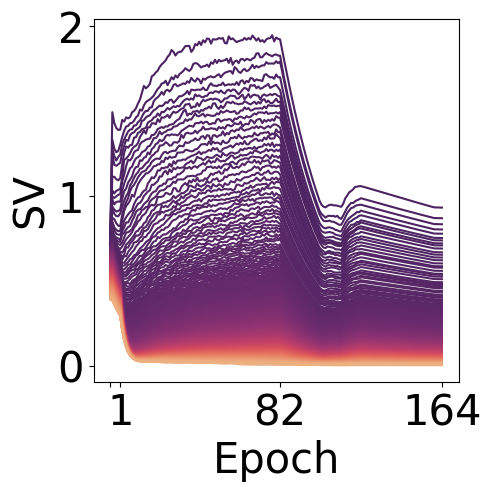}\hfil
    \includegraphics[width=0.5\linewidth]{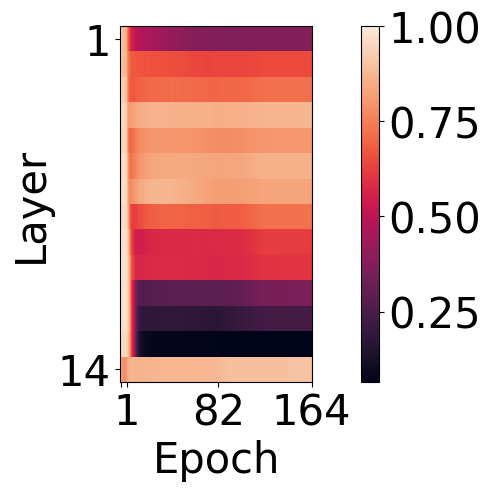}
    \\
    \includegraphics[width=0.5\linewidth]{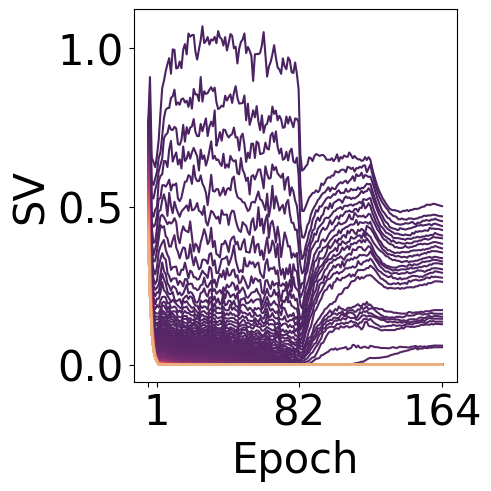}\hfil
    \includegraphics[width=0.5\linewidth]{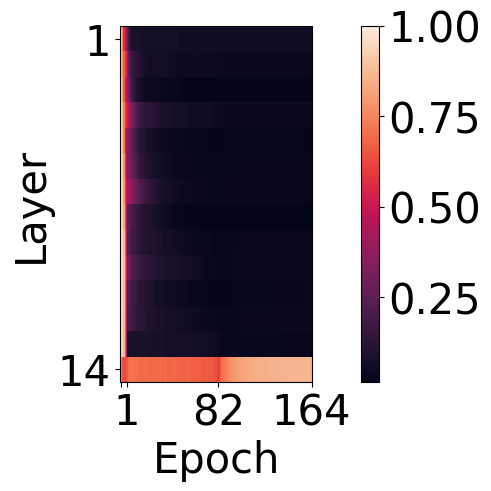}
    \\
    \includegraphics[width=0.5\linewidth]{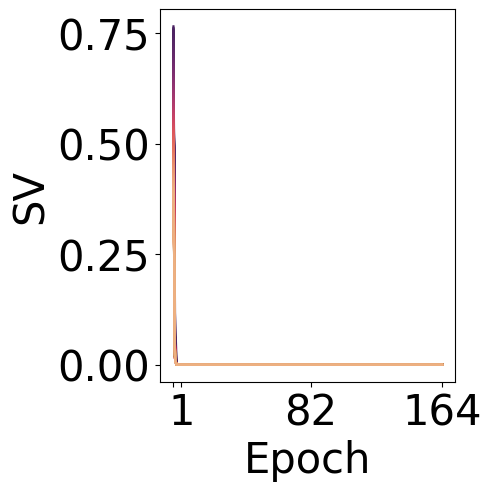}\hfil
    \includegraphics[width=0.5\linewidth]{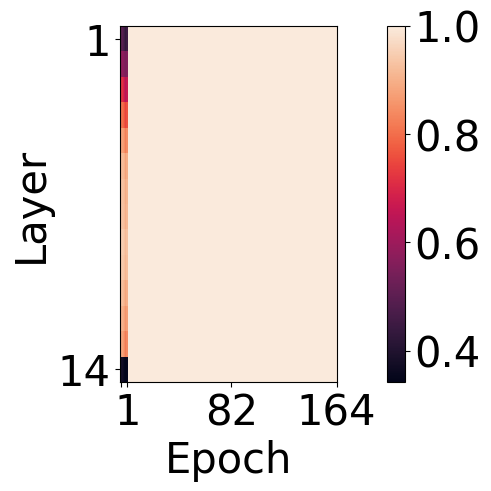}
    \caption{VGG}
  \end{subfigure}
  \begin{subfigure}[b]{0.24\linewidth}
    \centering
    \includegraphics[width=0.5\linewidth]{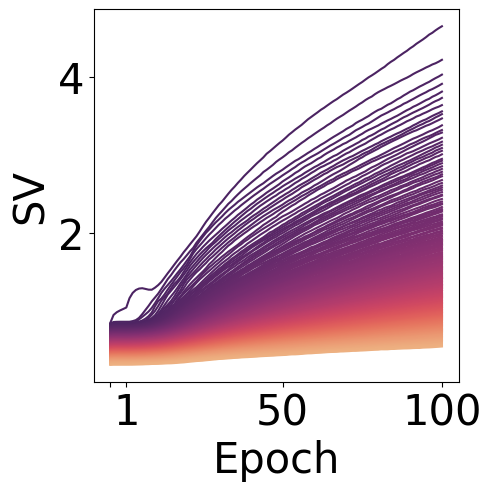}\hfil
    \includegraphics[width=0.5\linewidth]{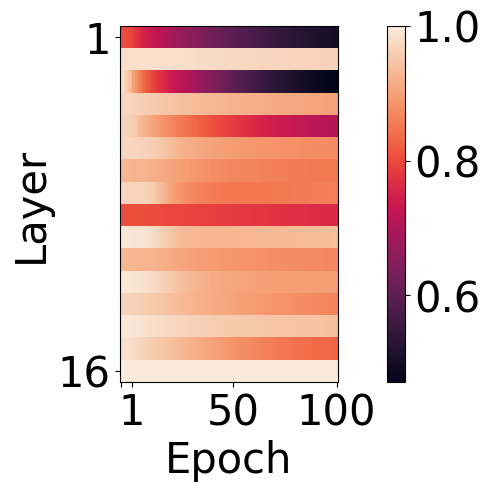}
    \\
    \includegraphics[width=0.5\linewidth]{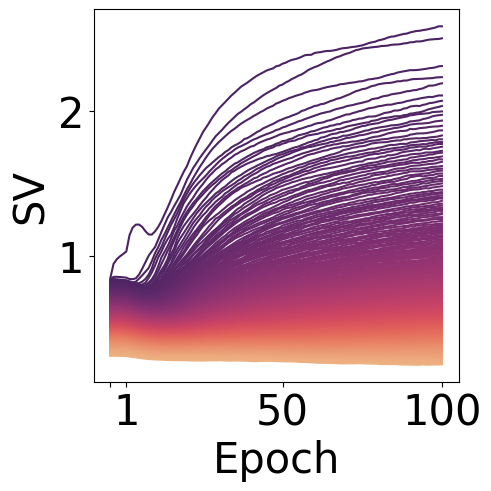}\hfil
    \includegraphics[width=0.5\linewidth]{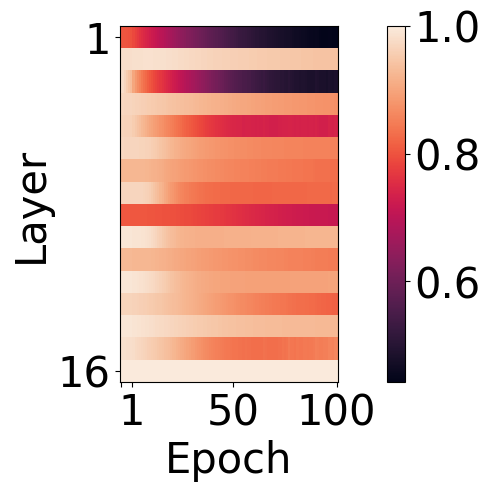}
    \\
    \includegraphics[width=0.5\linewidth]{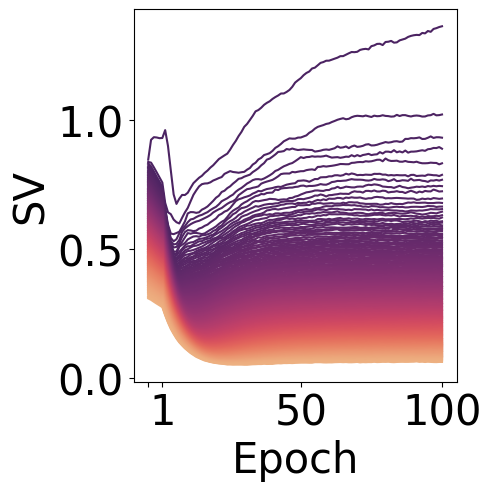}\hfil
    \includegraphics[width=0.5\linewidth]{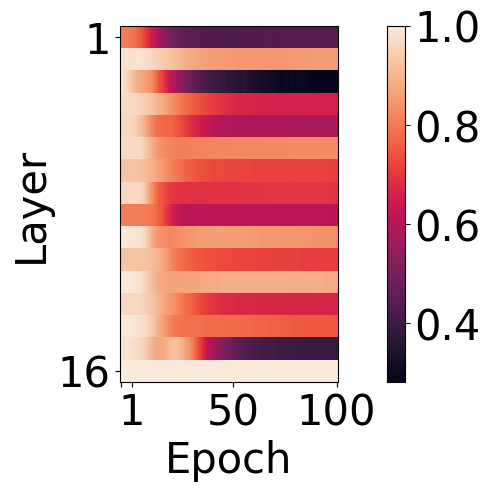}
    \\
    \includegraphics[width=0.5\linewidth]{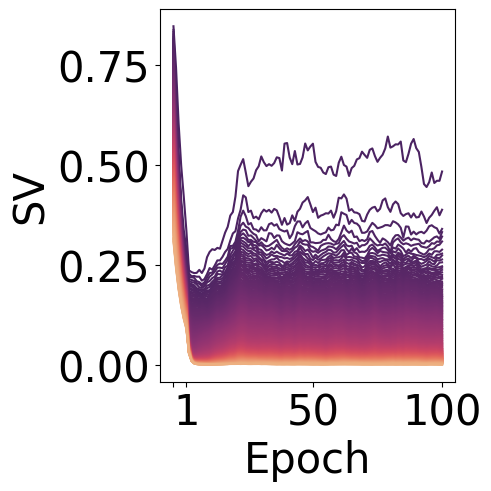}\hfil
    \includegraphics[width=0.5\linewidth]{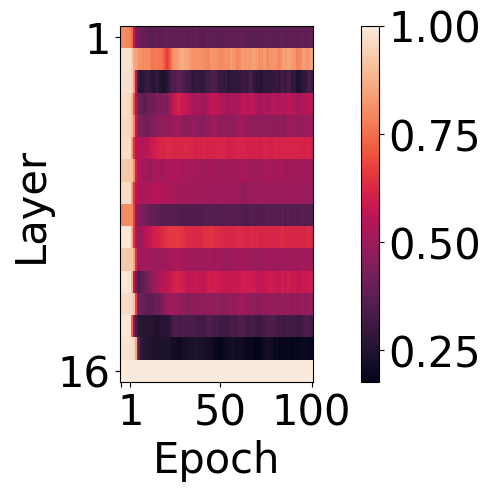}
    \caption{UNet}
  \end{subfigure}
  \begin{subfigure}[b]{0.24\linewidth}
    \centering
    \includegraphics[width=0.5\linewidth]{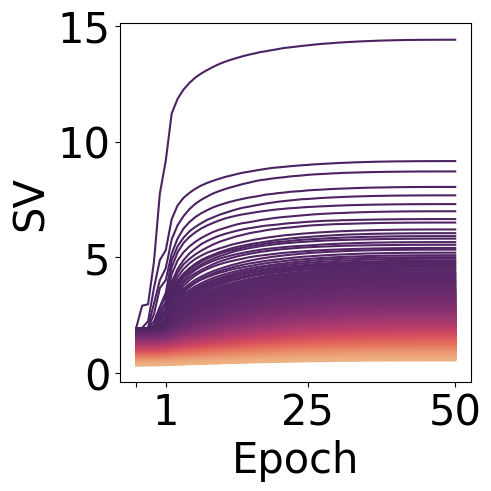}\hfil
    \includegraphics[width=0.5\linewidth]{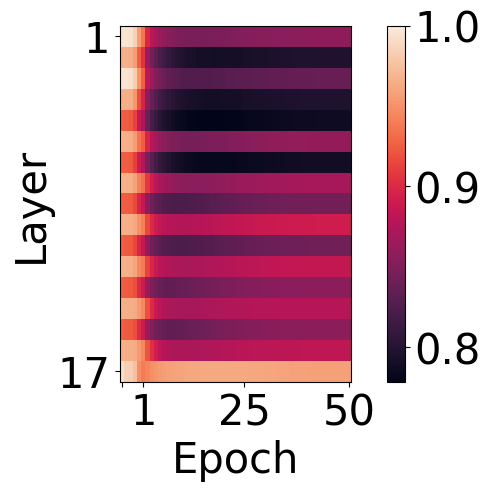}
    \\
    \includegraphics[width=0.5\linewidth]{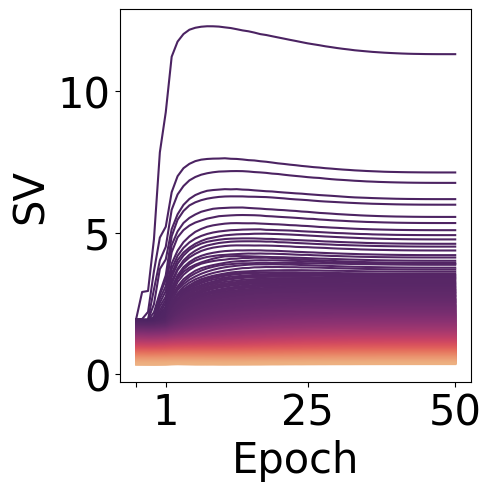}\hfil
    \includegraphics[width=0.5\linewidth]{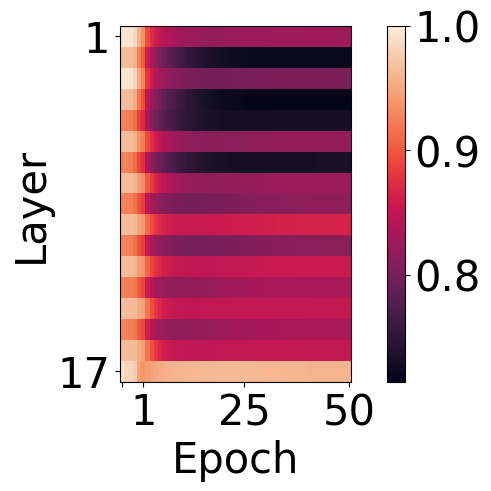}
    \\
        \includegraphics[width=0.5\linewidth]{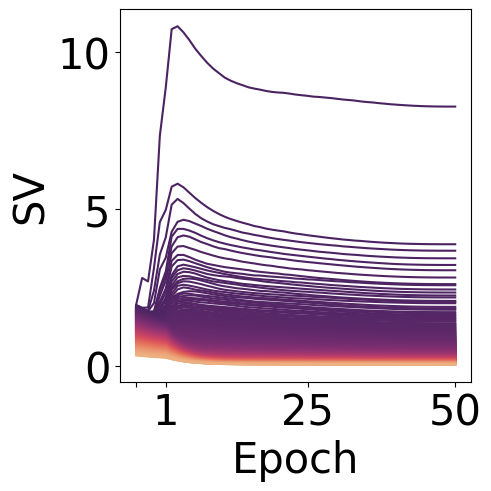}\hfil
    \includegraphics[width=0.5\linewidth]{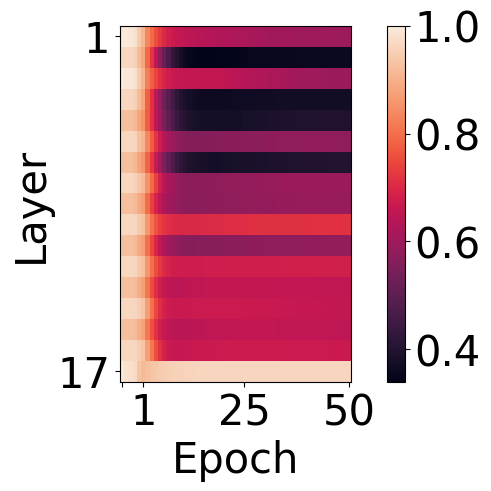}
    \\
    \includegraphics[width=0.5\linewidth]{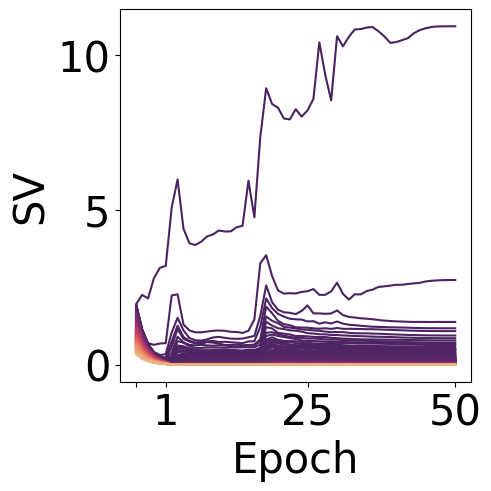}\hfil
    \includegraphics[width=0.5\linewidth]{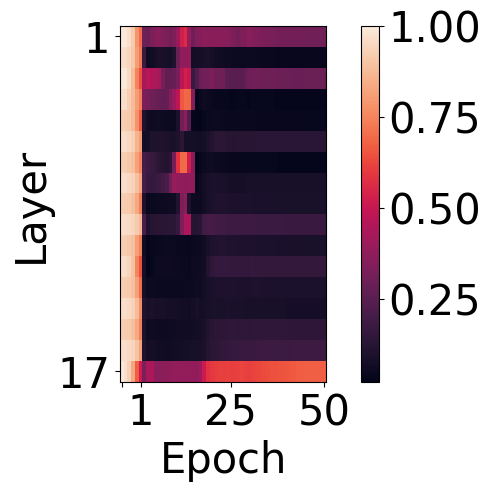}
    \caption{LSTM}
  \end{subfigure}
  \begin{subfigure}[b]{0.24\linewidth}
    \centering
    \includegraphics[width=0.5\linewidth]{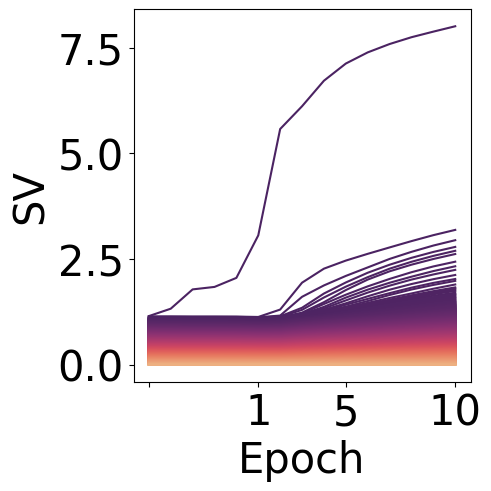}\hfil
    \includegraphics[width=0.5\linewidth]{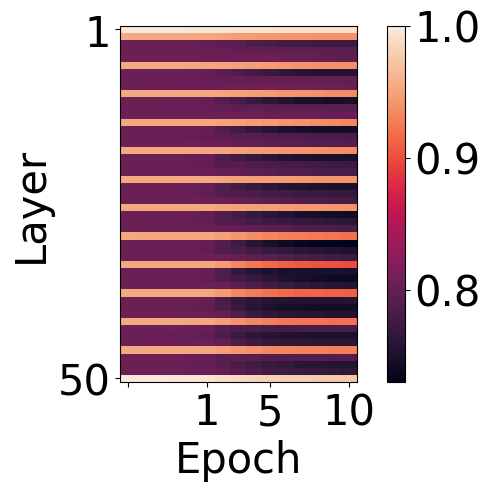}
    \\
    \includegraphics[width=0.5\linewidth]{figs/tfmr/weight_decay_0.1-sv-model.transformer_encoder.layers.7.linear1.weight_sv.png}\hfil
    \includegraphics[width=0.5\linewidth]{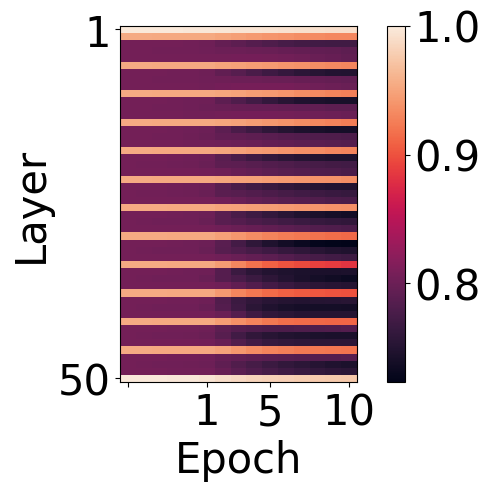}
    \\
    \includegraphics[width=0.5\linewidth]{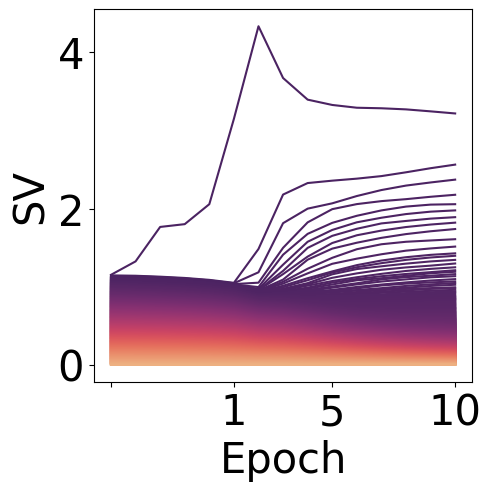}\hfil
    \includegraphics[width=0.5\linewidth]{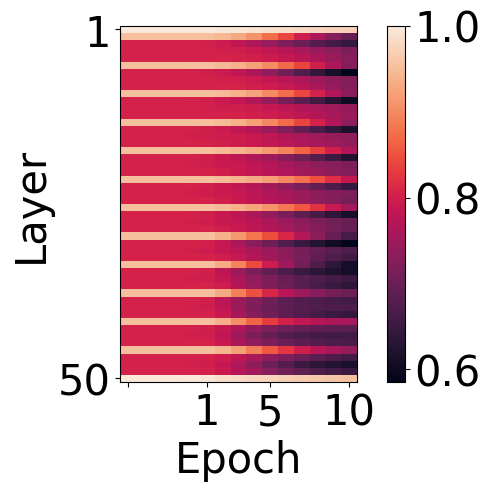}
    \\
    \includegraphics[width=0.5\linewidth]{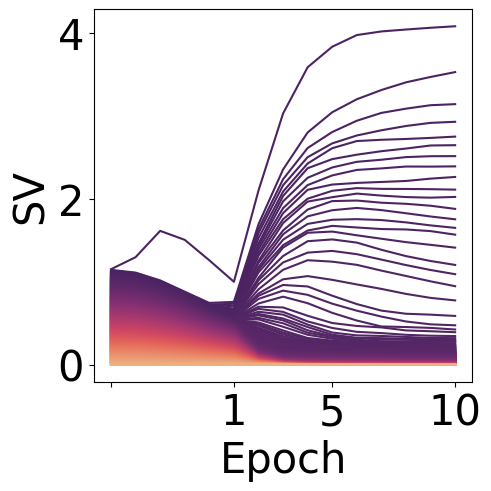}\hfil
    \includegraphics[width=0.5\linewidth]{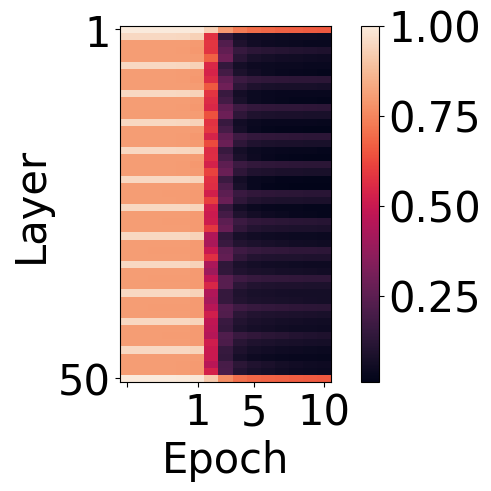}
    \caption{Transformer}
  \end{subfigure}
  \caption{SV evolution for a single matrix and normalized effective rank (Eqn.~\ref{eqn:normalized-effective-rank}) across matrices over time, where the rows use differing amounts of weight decay. From top to bottom, for VGG we use coefficients $\{ 0, 0.001, 0.01, 0.1 \}$, while for other networks we use coefficients $\{0, 0.1, 1, 10\}$. Higher weight decay coefficients promote more aggressive rank minimization. VGG uses SGD and momentum, while the rest use AdamW~\citep{loshchilov2018fixing}, which may explain the earlier norm collapse.}
  \label{fig:wd-effective-rank}
\end{figure*}

We speculate on the intuition of the mechanism in more practical settings. Notice in its simplest form that weight decay asks for $\text{arg min}_{W} \, \mathcal{L}(W) + \lambda \lVert W \rVert_F^2$, where $\lVert W \rVert_F^2 = \sum_{i=1}^R \sigma_i^2$ with singular values $\sigma_i$ of weight matrix $W$ with rank $R$. We saw that larger singular values of neural networks grow faster (Fig.~\ref{fig:normalized-effective-rank} top row) and that the top singular vectors are much more useful for minimizing task loss than the bottom ones (Fig.~\ref{fig:pruned-performance}). Thus, with minor weight decay regularization, one straightforward solution for the network may be to minimize the rank of a given weight matrix while preserving the top singular values to minimize $\mathcal{L}(W)$. \citet{timor2023implicit} argue a similar effect as if all singular values are less than 1, the norm of activations will shrink with depth, so it will be impossible to pass signals with sufficiently deep networks. Thus it is important for a few singular values to be sufficiently large.

Figure~\ref{fig:wd-effective-rank} shows that adding weight decay produces this exact low-rank behavior, while too much weight decay leads to complete norm collapse. The exact choice of ``too much'' varies across architectures and tasks, though it may be due to the use of SGD for VGG~\citep{simonyan2014very} instead of AdamW~\citep{loshchilov2018fixing} for the rest.

\begin{figure*}[t!]
  \centering
  \begin{subfigure}[b]{0.24\linewidth}
    \centering
    \includegraphics[width=0.5\linewidth]{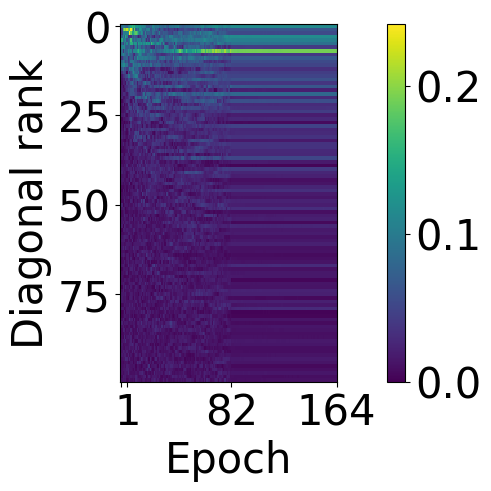}\hfil
    \includegraphics[width=0.5\linewidth]{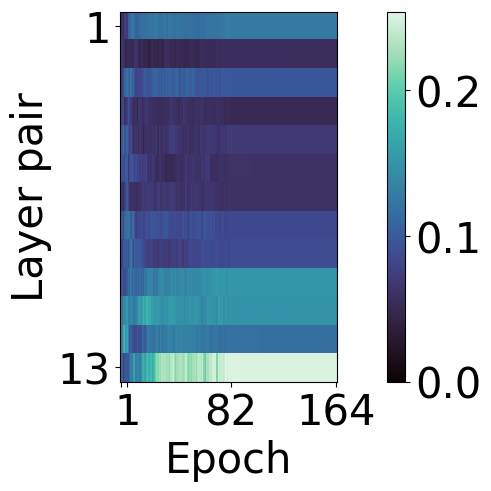}
    \\
    \includegraphics[width=0.5\linewidth]{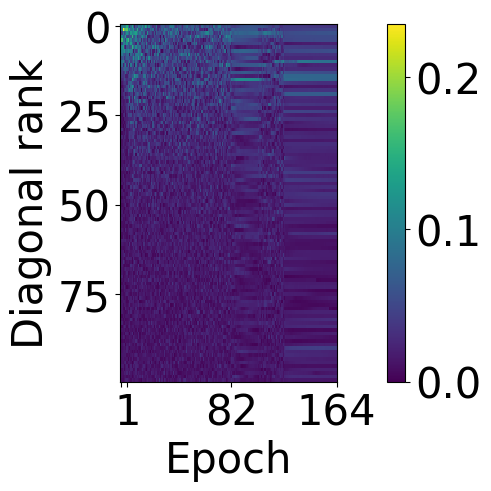}\hfil
    \includegraphics[width=0.5\linewidth]{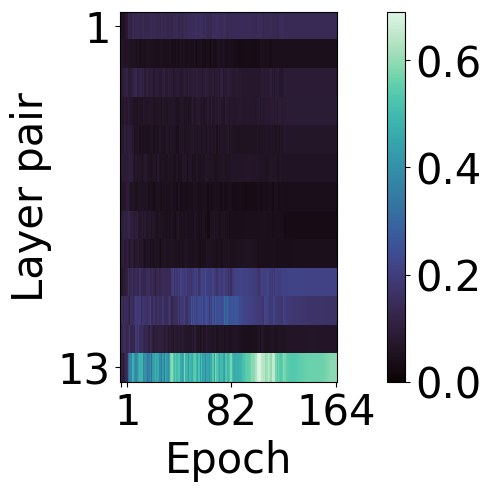}
    \\
    \includegraphics[width=0.5\linewidth]{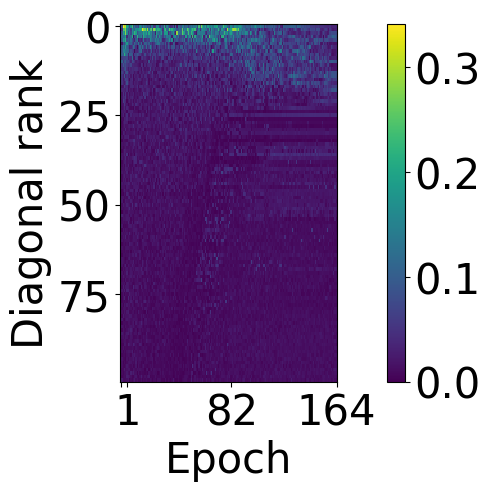}\hfil
    \includegraphics[width=0.5\linewidth]{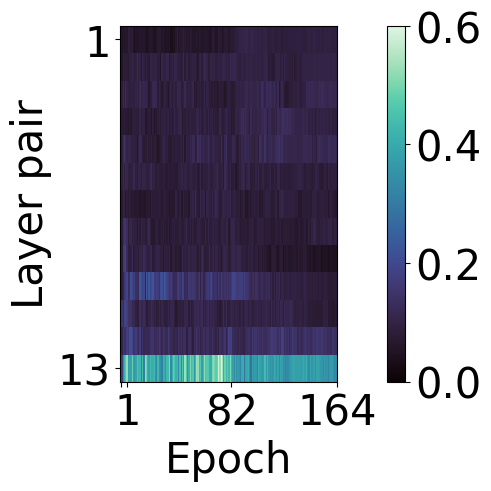}
    \\
    \includegraphics[width=0.5\linewidth]{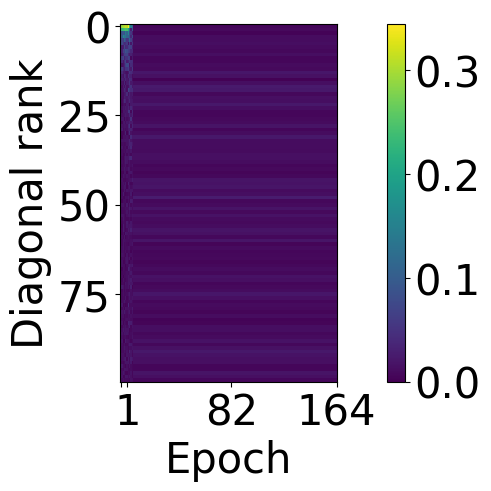}\hfil
    \includegraphics[width=0.5\linewidth]{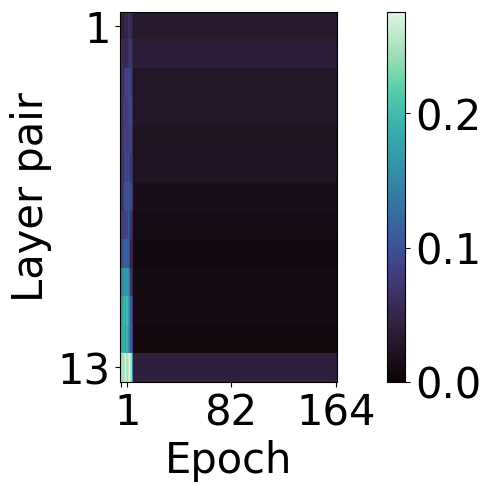}
    \caption{VGG}
  \end{subfigure}
  \begin{subfigure}[b]{0.24\linewidth}
    \centering
    \includegraphics[width=0.5\linewidth]{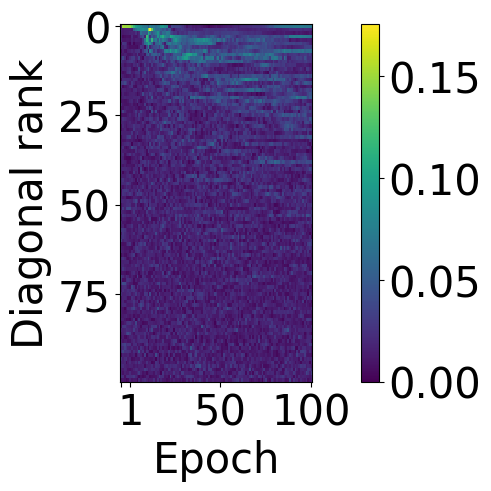}\hfil
    \includegraphics[width=0.5\linewidth]{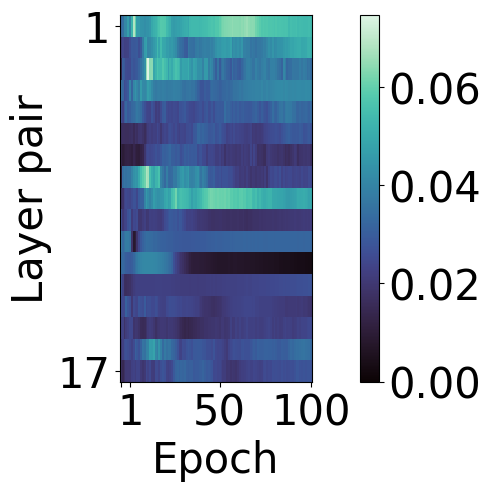}
    \\
    \includegraphics[width=0.5\linewidth]{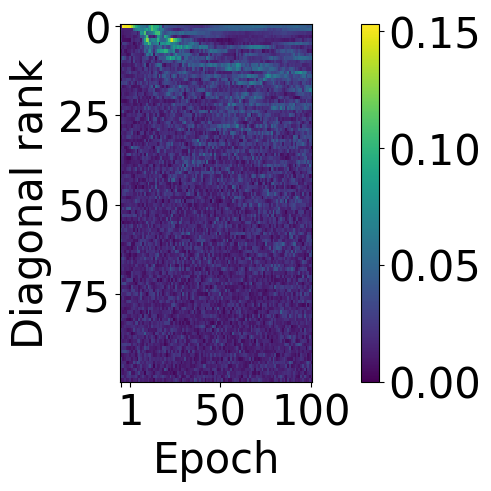}\hfil
    \includegraphics[width=0.5\linewidth]{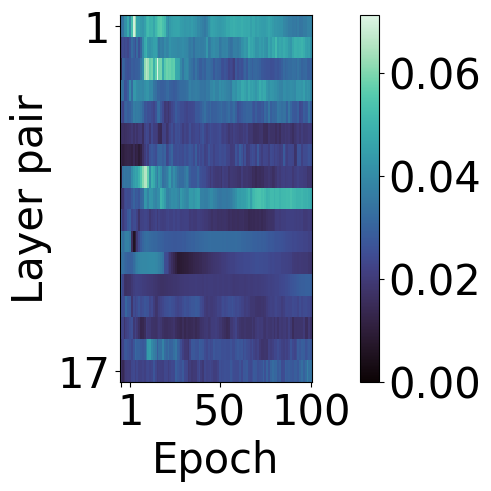}
    \\
    \includegraphics[width=0.5\linewidth]{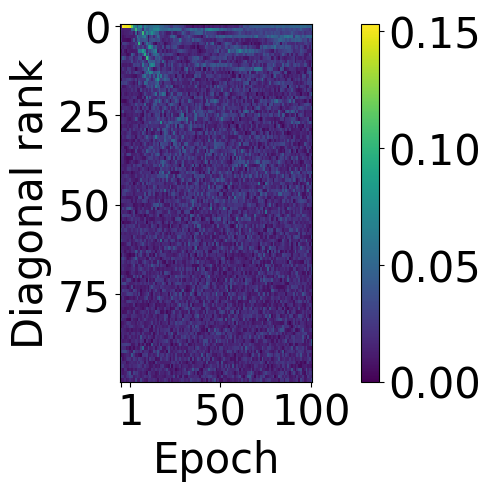}\hfil
    \includegraphics[width=0.5\linewidth]{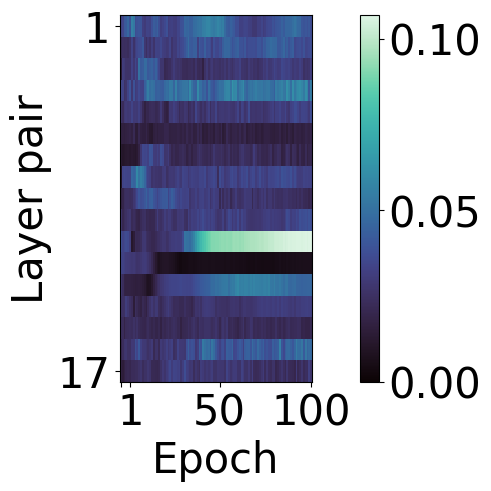}
    \\
    \includegraphics[width=0.5\linewidth]{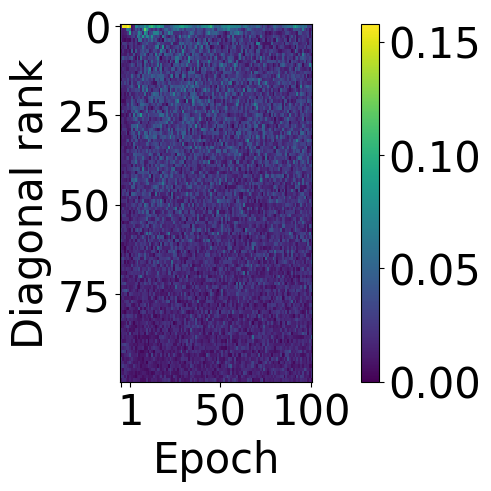}\hfil
    \includegraphics[width=0.5\linewidth]{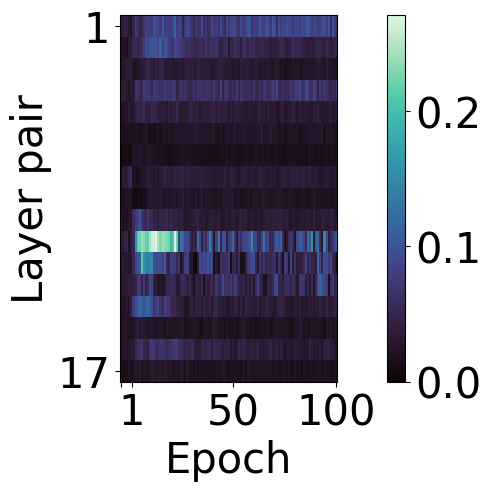}
    \caption{UNet}
  \end{subfigure}
  \begin{subfigure}[b]{0.24\linewidth}
    \centering
    \includegraphics[width=0.5\linewidth]{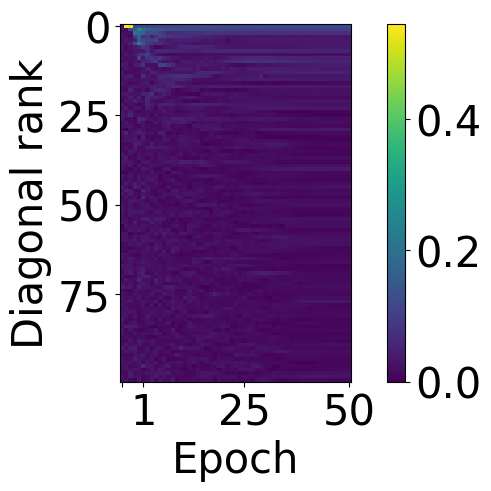}\hfil
    \includegraphics[width=0.5\linewidth]{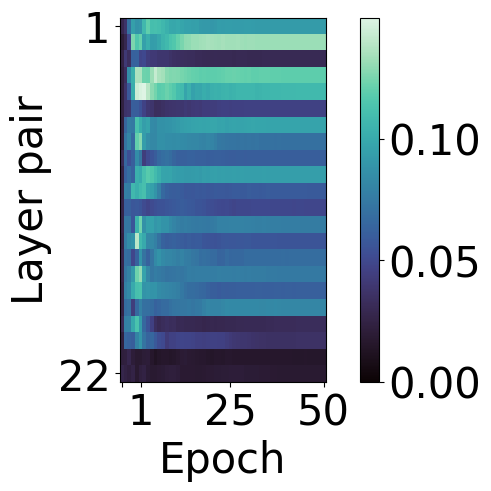}
    \\
    \includegraphics[width=0.5\linewidth]{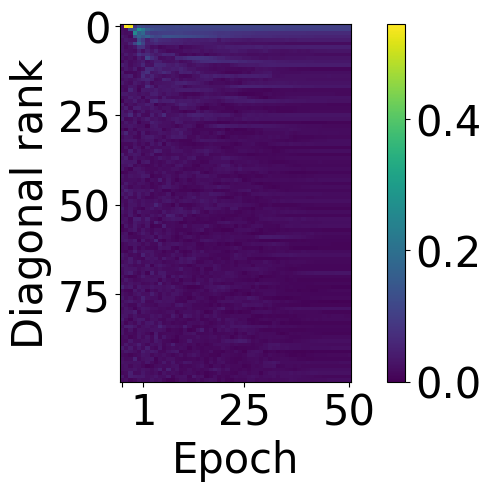}\hfil
    \includegraphics[width=0.5\linewidth]{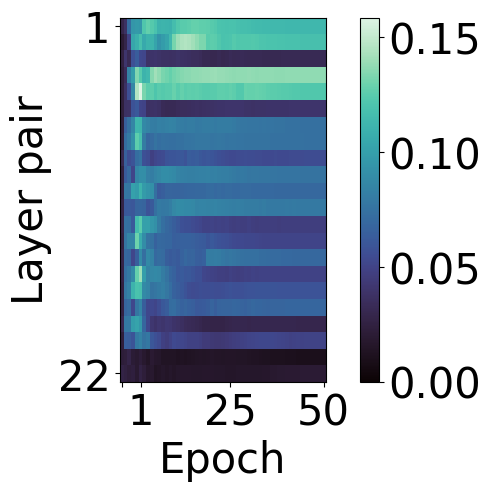}
    \\
    \includegraphics[width=0.5\linewidth]{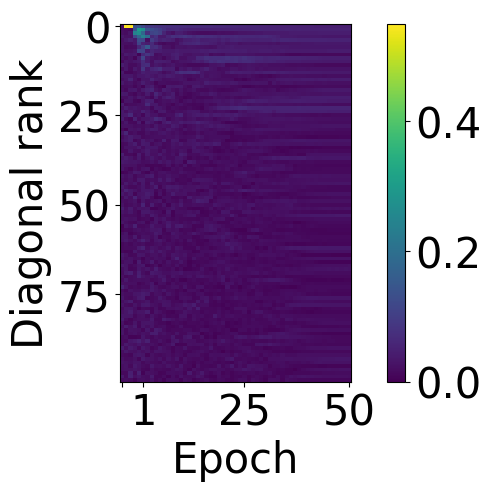}\hfil
    \includegraphics[width=0.5\linewidth]{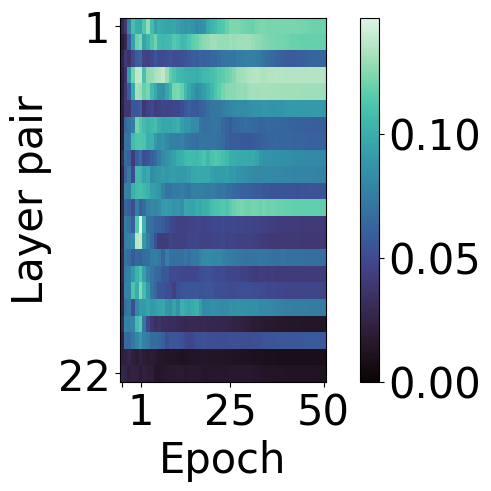}
    \\
    \includegraphics[width=0.5\linewidth]{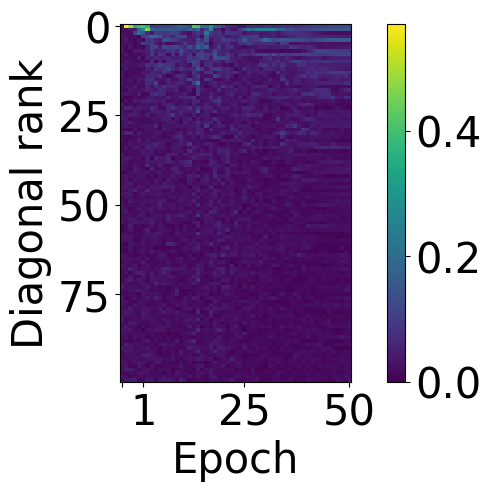}\hfil
    \includegraphics[width=0.5\linewidth]{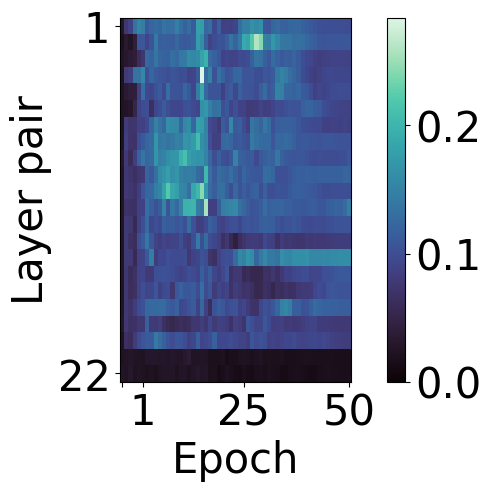}
    \caption{LSTM}
  \end{subfigure}
  \begin{subfigure}[b]{0.24\linewidth}
    \centering
    \includegraphics[width=0.5\linewidth]{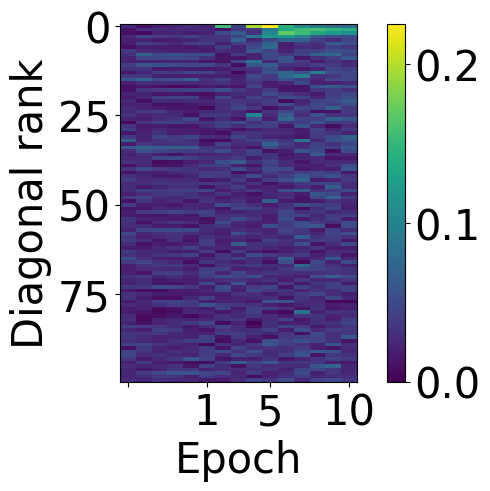}\hfil
    \includegraphics[width=0.5\linewidth]{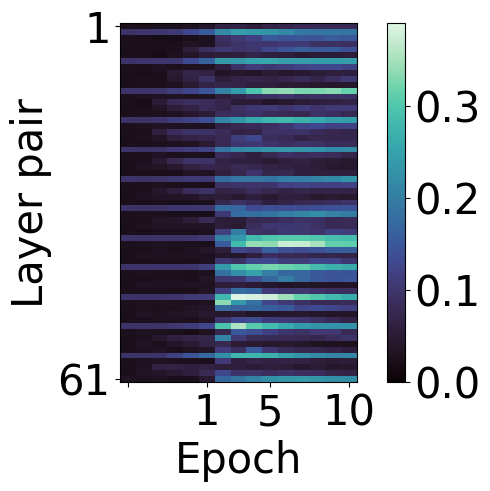}
    \\
    \includegraphics[width=0.5\linewidth]{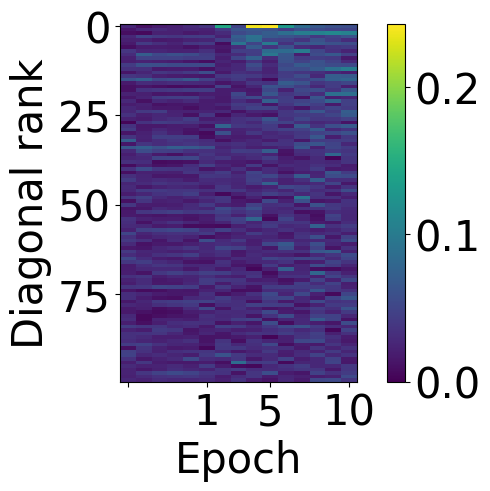}\hfil
    \includegraphics[width=0.5\linewidth]{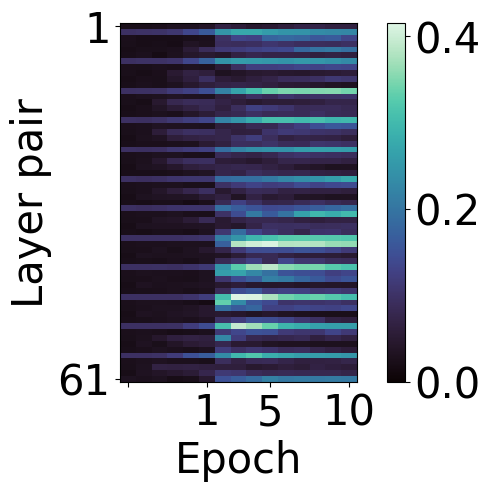}
    \\
    \includegraphics[width=0.5\linewidth]{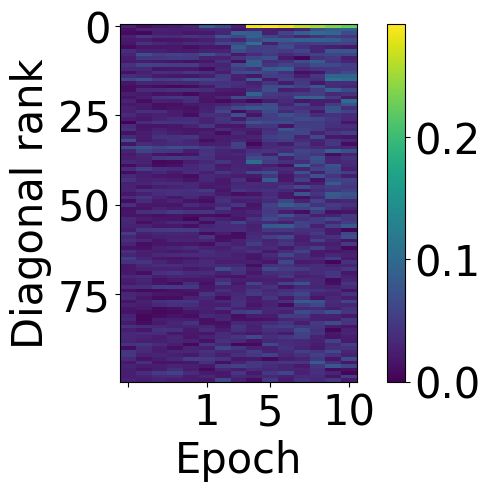}\hfil
    \includegraphics[width=0.5\linewidth]{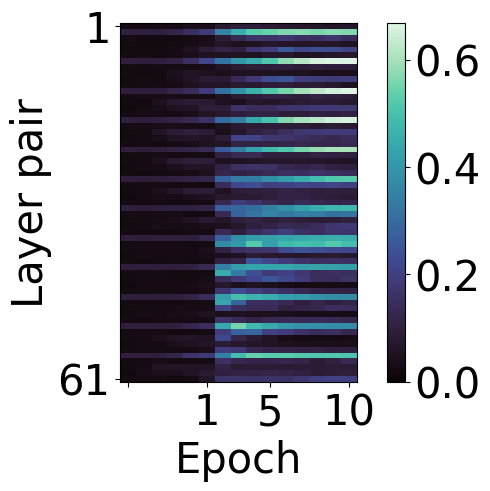}
    \\
    \includegraphics[width=0.5\linewidth]{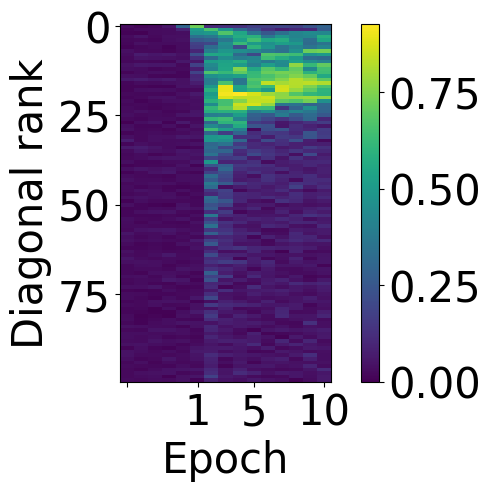}\hfil
    \includegraphics[width=0.5\linewidth]{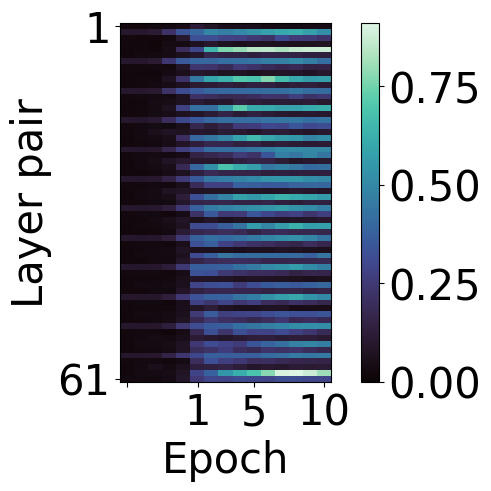}
    \caption{Transformer}
  \end{subfigure}
  \caption{Diagonal of alignment for a single pair over time (Eqn.~\ref{eqn:alignment-matrix}) and alignment metric across pairs of matrices over time (Eqn.~\ref{eqn:alignment-measure}) where the y-axis represents depth.  From top to bottom, for VGG we use coefficients $\{ 0, 0.001, 0.01, 0.1\}$, while for other networks we use coefficients $\{0, 0.1, 1, 10\}$. We see that the maximum alignment magnitude is higher with large weight decay, and in particular, the Transformer has the strongest alignment even when nonlinearities separate the MLP layers.}
  \label{fig:wd-alignment-score}
\end{figure*}

In addition, even more surprisingly, large amounts of weight decay promote a tighter alignment in the top singular vectors of consecutive layers of the Transformer, but not other networks. We see this in Figure~\ref{fig:wd-alignment-score}. This behavior is quite reminiscent of the balancedness condition~\citep{arora2018optimization, arora2019implicit, du2018algorithmic}, though the Transformer considered here has nonlinearities and much more complex structures. It is curious that the trend is so strong for only this architecture. We also provide additional evidence in Appendix~\ref{app:experimental-details} where Figure~\ref{fig:weight-decay-performance} shows that the solutions with very high weight decay are still performant, even though they are much lower rank. Though it is difficult to argue as simple a trend as ``lower rank equals better generalization'' because one does not know the minimal rank necessary for a given task, it is notable that the role of weight decay for improving generalization is tied up with its function as a rank regularizer. In addition, although we lack precise tools to entirely interpret complex models, when there are only a few ranks per matrix it may become possible to extend analysis efforts~\citep{nanda2023progress, praggastis2022svd} to more complex domains.

\begin{figure*}[!t]
  \centering
  \begin{subfigure}[b]{0.18\linewidth}
    \centering
    \includegraphics[width=\linewidth]{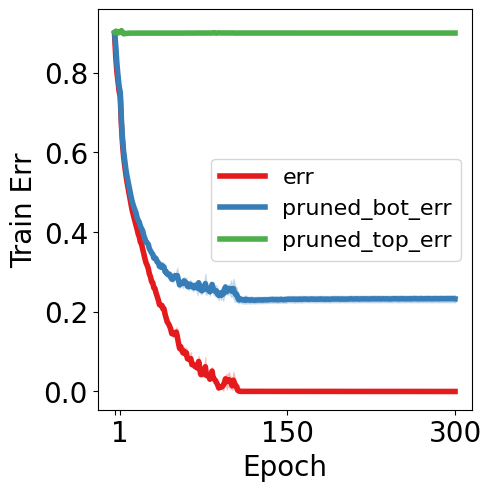}
    \\
    \includegraphics[width=\linewidth]{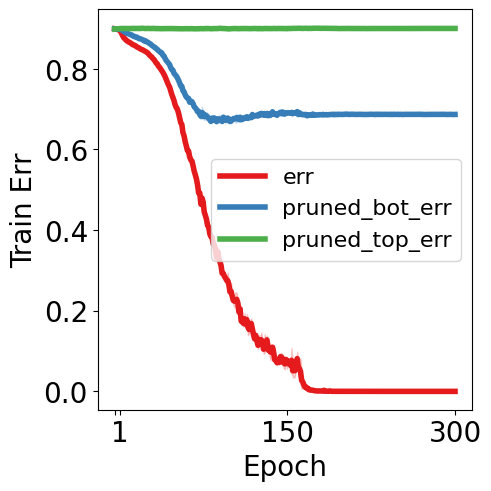}
    \caption{Train Err.}
  \end{subfigure}
  \begin{subfigure}[b]{0.18\linewidth}
    \centering
    \includegraphics[width=\linewidth]{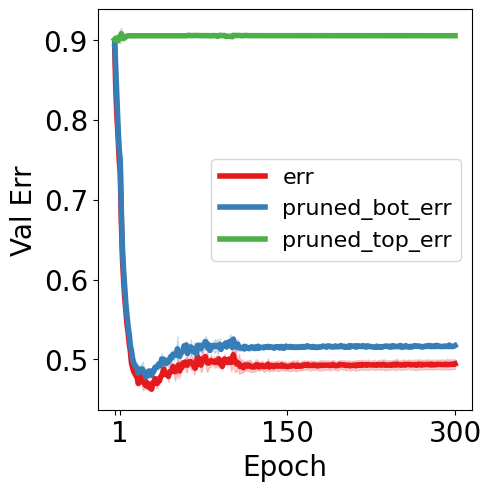}
    \\
    \includegraphics[width=\linewidth]{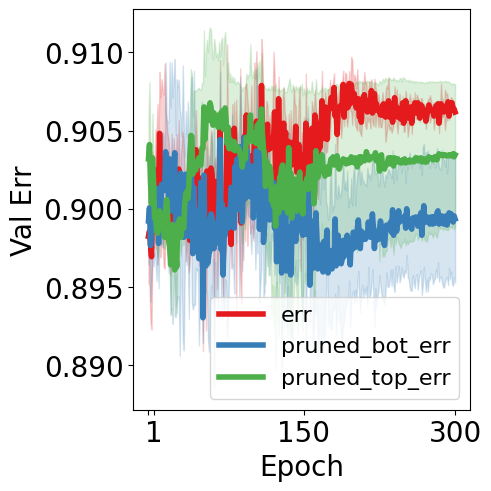}
    \caption{Val. Err.}
  \end{subfigure}
  \begin{subfigure}[b]{0.18\linewidth}
    \centering
    \includegraphics[width=\linewidth]{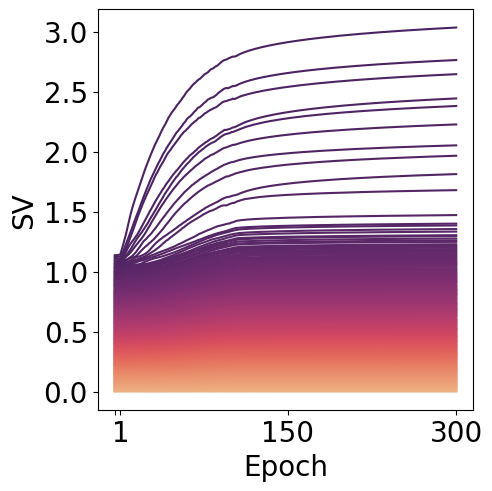}
    \\
    \includegraphics[width=\linewidth]{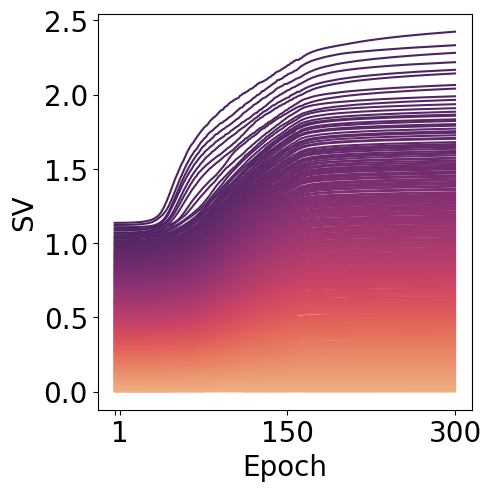}
    \caption{SVs}
  \end{subfigure}
  \begin{subfigure}[b]{0.18\linewidth}
    \centering
    \includegraphics[width=\linewidth]{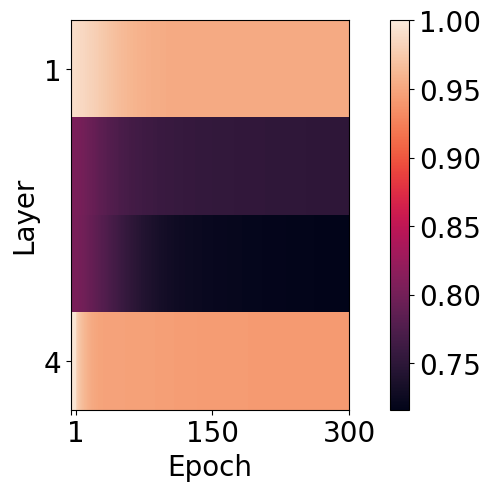}
    \\
    \includegraphics[width=\linewidth]{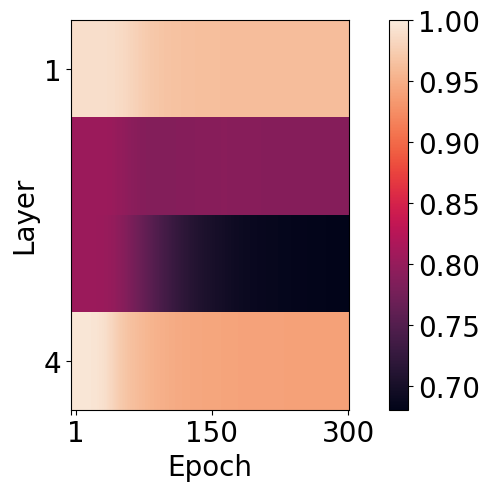}
    \caption{Eff. Rank}
  \end{subfigure}
  \begin{subfigure}[b]{0.18\linewidth}
    \centering
    \includegraphics[width=\linewidth]{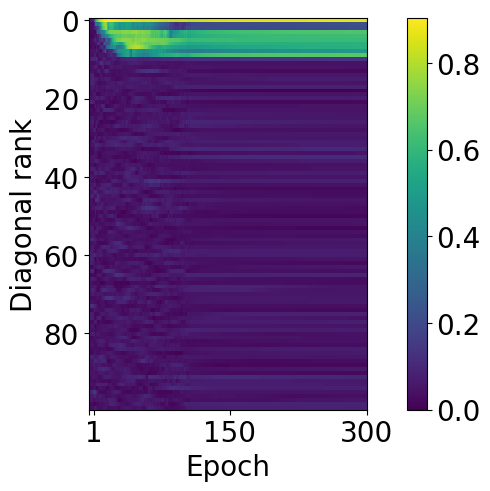}
    \\
    \includegraphics[width=\linewidth]{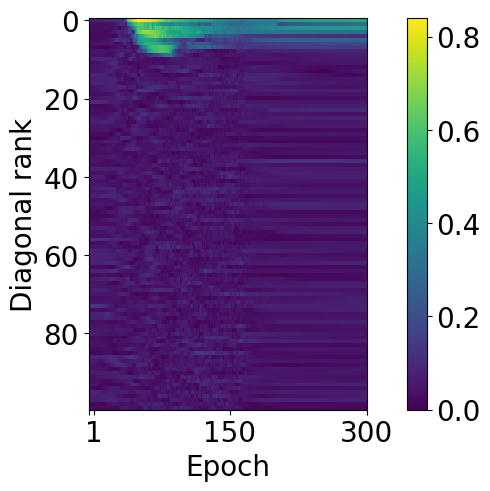}
    \caption{Alignment}
  \end{subfigure}
  \caption{\textbf{Top row:} results with true labels. \textbf{Bottom row:} results with random labels. We see that the middle layers have a lower effective rank when using true labels and that alignment in the middle layers persists throughout training, unlike in the random label case. We emphasize this alignment occurs despite the nonlinearities.}
  \label{fig:random-labels}
\end{figure*}

\section{Spectral Dynamics with Random Labels}\label{sec:random_labels}

Given the observations connecting generalization and rank thus far, and the enlightening view on the implicit effects of weight decay, we are interested in seeing whether the perspective developed sheds any light on the classic random label memorization experiments of \citet{zhang2021understanding}.

Similar to \citet{zhang2021understanding}, we train a simple MLP to fit random or true labels on CIFAR10. Please see Appendix~\ref{app:experimental-details} for the details regarding the experimental setup. \citet{zhang2021understanding} decay the learning rate to zero, and the random label experiments only converge late in training. Consequently, we use a constant learning rate to control this phenomenon. We see in Figure~\ref{fig:random-labels} that both cases are able to achieve zero error, though with different singular value evolution and alignment in the middle layer.

Surprisingly, we see that with true labels the inner layers are low rank, while with random labels they are much higher rank. This may be explained by the shared structure in the true classes of the dataset, which manifests in the parameters. Even more surprisingly, we find here that even without weight decay, inner layers align with true labels, while with random labels, this alignment occurs and then disappears with more training. This is particularly intriguing as there are non-linearities that could theoretically separate the network from the linear case, and yet strong alignment occurs despite that. Such alignment has not yet been leveraged by existing theory, and might provide structured assumptions for new understanding. In summary, these results suggest that viewing generalization through the lens of rank may be fruitful.

\section{Beyond Generalization}\label{sec:beyond-generalization}

We have seen over the course of many experiments that deep models are biased toward low rank, and that there is a tempting connection between rank minimization and generalization. Still, the lens of spectral dynamics can be applied more broadly. In the following subsections, we explore two phenomena: lottery tickets~\citep{frankle2018lottery} and linear mode connectivity~\citep{frankle2020linear}. Beyond shedding further light on neural networks, these phenomena have implications for more efficient inference and storage, as well as understanding the importance of pretraining~\citep{neyshabur2020being}. We find that lottery tickets are a sparse approximation of final-checkpoint top singular vectors. The ability to linearly interpolate between faraway checkpoints and improve performance coincides strongly with top singular vector sharing between checkpoints. Such observations may form a foundation for a better understanding compression and model averaging~\citep{wortsman2022model, ilharco2022editing}.

\subsection{Top Singular Vectors Become Stable Earlier}

\begin{figure}
  \centering
  \begin{subfigure}[b]{0.24\columnwidth}
    \centering
    \includegraphics[width=\columnwidth]{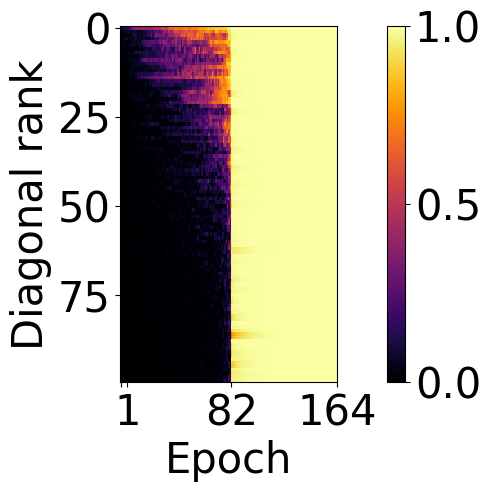}
    \\
    \includegraphics[width=\columnwidth]{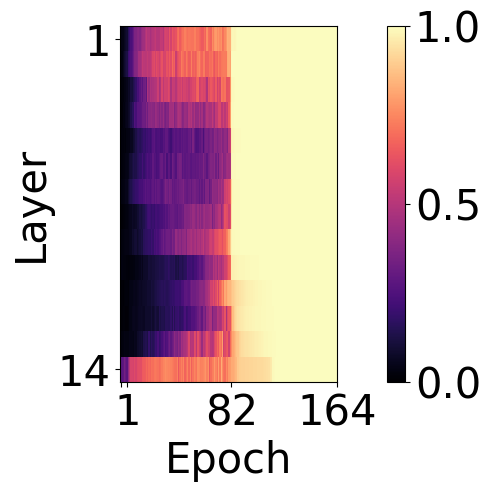}
    \caption{VGG}
  \end{subfigure}
  \begin{subfigure}[b]{0.24\columnwidth}
    \centering
    \includegraphics[width=\columnwidth]{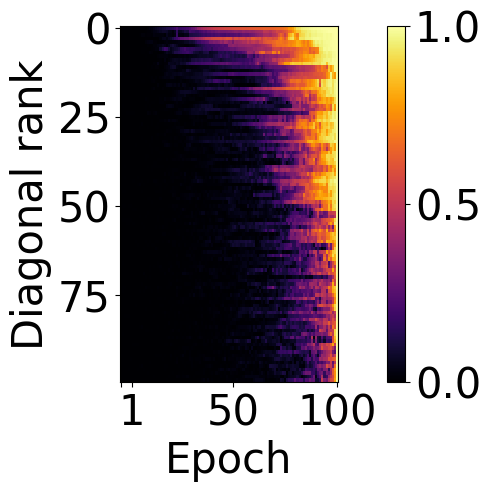}
    \\
    \includegraphics[width=\columnwidth]{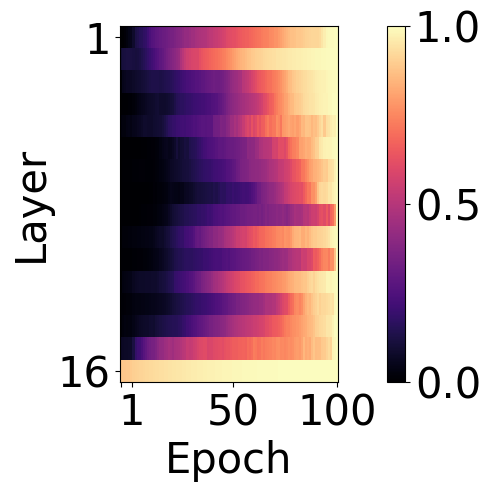}
    \caption{UNet}
  \end{subfigure}
  \begin{subfigure}[b]{0.24\columnwidth}
    \centering
    \includegraphics[width=\columnwidth]{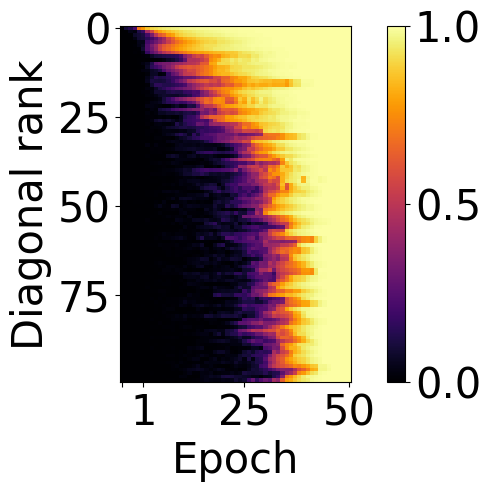}
    \\
    \includegraphics[width=\columnwidth]{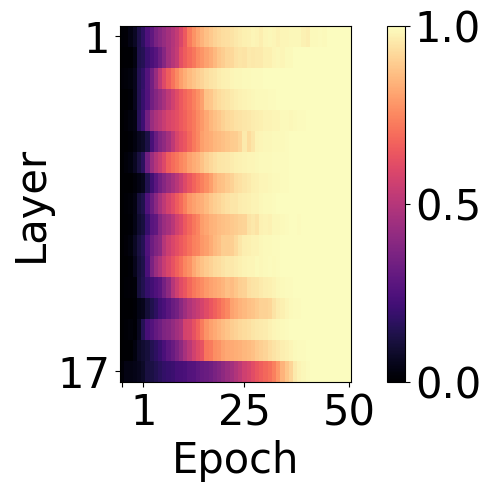}
    \caption{LSTM}
  \end{subfigure}
  \begin{subfigure}[b]{0.24\columnwidth}
    \centering
    \includegraphics[width=\columnwidth]{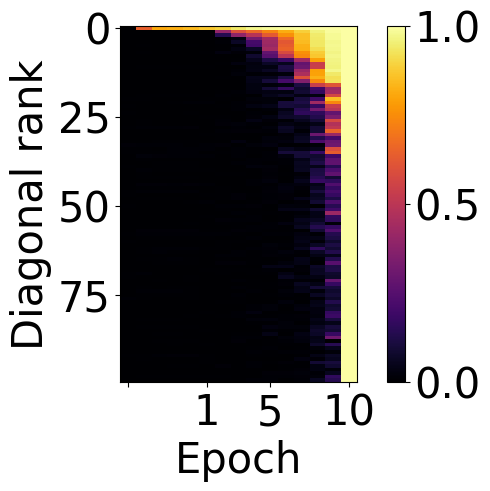}
    \\
    \includegraphics[width=\columnwidth]{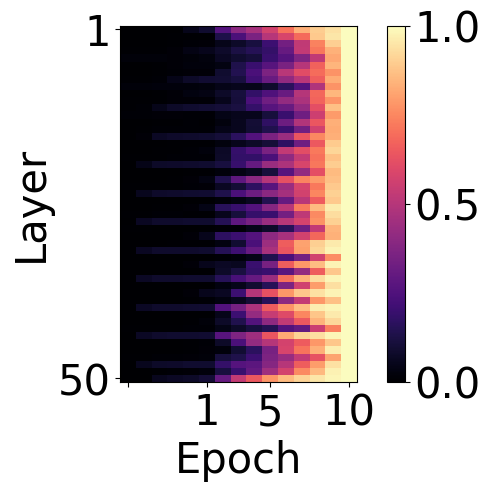}
    \caption{Transformer}
  \end{subfigure}
  \caption{\textbf{Top row:} Singular vector agreement for a single matrix in the middle of each model (diagonal of Eqn.~\ref{eqn:sva-matrix}). Notice top singular vectors become stable in direction earlier. \textbf{Bottom row:} Summary score for each matrix across architectures. As we move down the $y$-axis, the depth of the parameters in the model increases, while the $x$-axis tracks training time. The sharp transition midway through training in the VGG case is likely due to a 10x learning rate decay.}
  \label{fig:singular-vector-agreement}
\end{figure}

Before we explore the phenomena, we first make another observation that will be helpful. As top singular values grow disproportionately large, it would be natural that top singular vectors become stable in direction as the gradients remain small. To demonstrate this, for a given matrix in the network $W_i(t) = \sum_{j=1}^R \sigma_j(t) u_j(t) v_j(t)^\top$ at training time t, we compute
\begin{equation} \label{eqn:sva-matrix}
    S(t)_{jk} = \lvert \langle u_j(t)v_j(t)^\top, u_k(T)v_k(T)^\top \rangle \rvert,
\end{equation}
where $T$ is the final step of training, and the absolute value is taken to ignore sign flips in the SVD computation. We then plot the diagonal of this matrix $S(t)_{ii}~\forall~i \leq 100$ over time. We also use a scalar measure of the diagonal to summarize like in the alignment case: $s(t) = \frac{1}{10} \sum_i S(t)_{ii}$. In Figure~\ref{fig:singular-vector-agreement}, we see that top singular vectors converge in direction earlier than bottom vectors.

\subsection{Lottery Tickets Preserve Final Top Singular Vectors}

As large singular vectors will become stable late in training, we wonder about the connection to magnitude pruning and the lottery ticket hypothesis. \citet{frankle2018lottery} first showed evidence for the lottery ticket hypothesis, the idea that there exist sparse subnetworks of neural networks that can be trained to a comparable performance as the full network, where the sparse mask is computed from the largest magnitude weights of the network at the end of training. \citet{frankle2020linear} build further on this hypothesis and notice that, for larger networks, the masking cannot begin at initialization, but rather at some point early in training. Still, the mask must come from the end of training.

The reason for this particular choice of mask may be connected to the dynamics we previously observed. Specifically, at the end of training large singular values are disproportionately larger, so high-magnitude weights may correspond closely to weights in the top singular vectors at the end of training. If magnitude masks were computed at the beginning, the directions that would become the top singular vectors might be prematurely masked as they have not yet stabilized, which may prevent learning on the task.

\begin{figure*}[!t]
  \centering
   \begin{subfigure}[b]{0.18\linewidth}
    \centering
    \includegraphics[width=\linewidth]{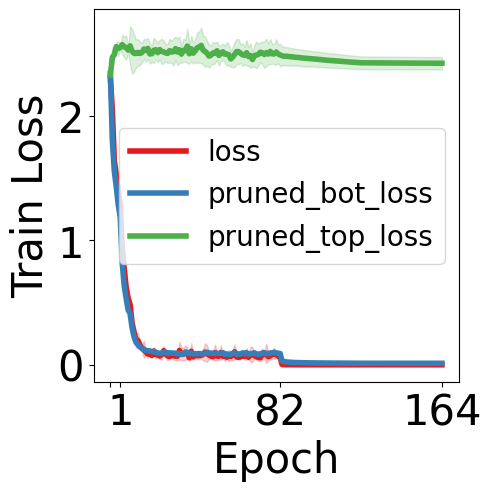}
    \\
    \includegraphics[width=\linewidth]{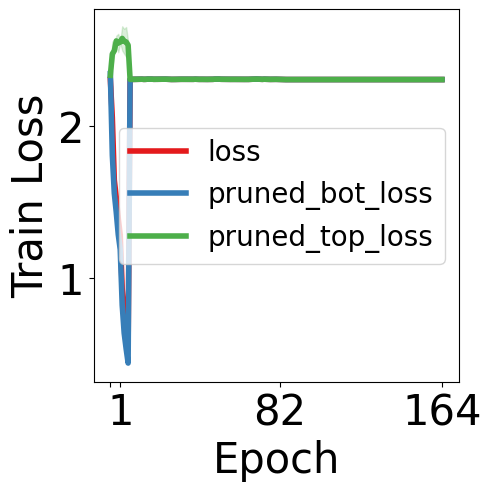}
    \caption{Loss}
  \end{subfigure}
  \begin{subfigure}[b]{0.18\linewidth}
    \centering
    \includegraphics[width=\linewidth]{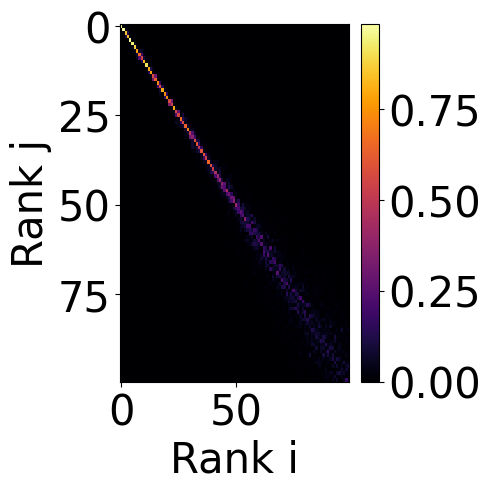}
    \\
    \includegraphics[width=\linewidth]{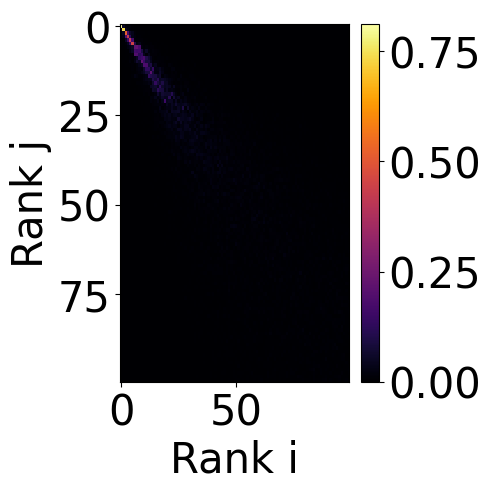}
    \caption{Pruned SVA}
  \end{subfigure}
  \begin{subfigure}[b]{0.18\linewidth}
    \centering
    \includegraphics[width=\linewidth]{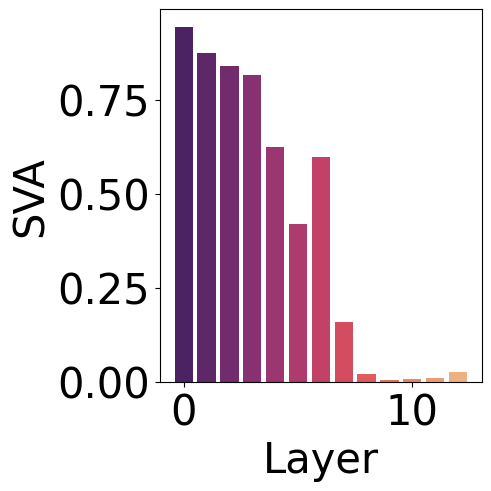}
    \\
     \includegraphics[width=\linewidth]{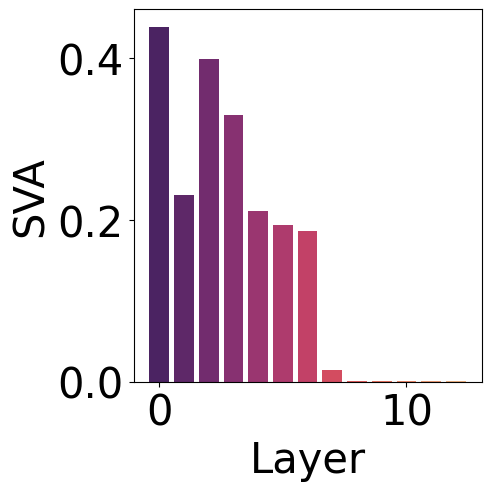}
    \caption{All Layers}
  \end{subfigure}
  \begin{subfigure}[b]{0.18\linewidth}
    \centering
    \includegraphics[width=\linewidth]{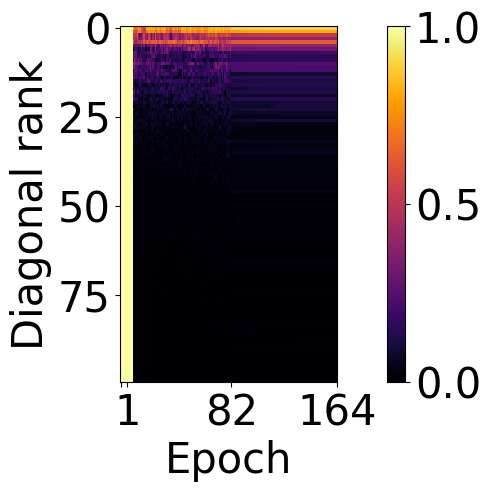}
    \\
    \includegraphics[width=\linewidth]{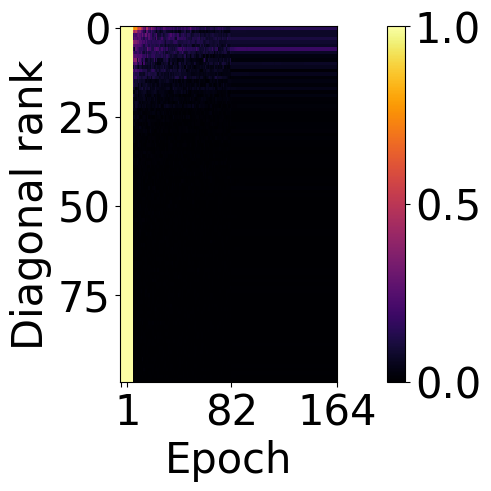}
    \caption{SVA evol.}
  \end{subfigure}
  \begin{subfigure}[b]{0.18\linewidth}
    \centering
    \includegraphics[width=\linewidth]{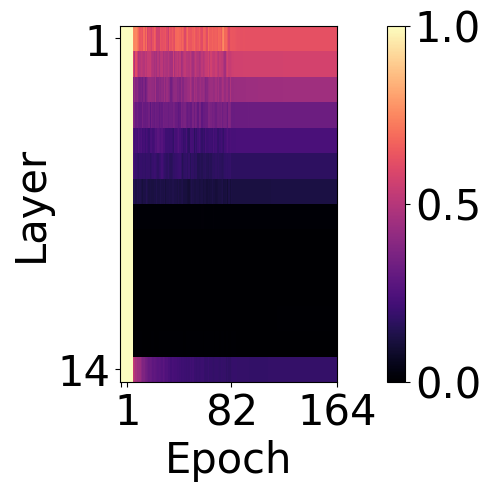}
    \\
    \includegraphics[width=\linewidth]{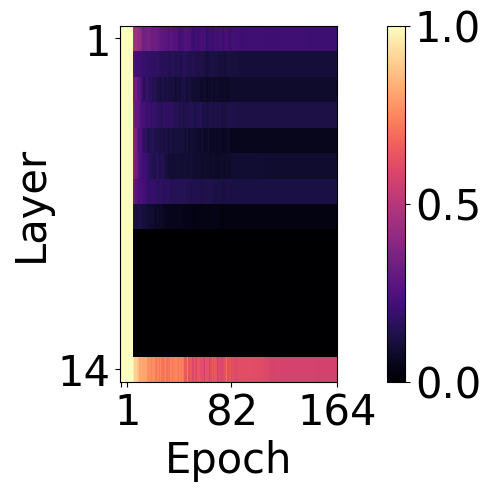}
    \caption{All Layers}
  \end{subfigure}
  \caption{\textbf{Top row:} Magnitude pruning. \textbf{Bottom row:} random pruning. \textbf{First column:} Training loss. We see that at 5\% sparsity magnitude pruning is significantly better than random pruning of the same layerwise sparsity. \textbf{2nd column:} Singular vector alignment pre- and post-pruning at the end of training for a single layer (the 3rd convolution). We see that magnitude pruning approximates the top singular vectors, while random pruning at the same level does not. \textbf{3rd column:} Singular vector alignment score pre- and post-pruning across all layers. Agreement is higher across all layers for magnitude pruning, though later layers do not agree, likely as later layers are wider so weights are lower magnitude. \textbf{4th column:} Singular vector alignment between the pruned and unpruned models along the training trajectory. We see that the magnitude pruning still has similar dynamics in its top singular vectors, while random pruning does not. \textbf{Last column:} Singular vector alignment score between pruned and unpruned models across layers and time. Again evolution is similar for early layers with magnitude pruning, and completely different for random pruning.}
  \label{fig:lottery-tickets}
\end{figure*}

Here we train an unmasked VGG-16~\citep{simonyan2014very} on CIFAR10, then compute either a random mask, or a global magnitude mask from the end of training, and rewind to an early point~\citep{frankle2020linear} to start sparse retraining. Please see Appendix~\ref{app:experimental-details} for details. In Figure~\ref{fig:lottery-tickets}, we plot the singular vector agreement (SVA, Eqn.~\ref{eqn:sva-matrix}) between the final model, masked and unmasked, where we see exactly that magnitude masks preserve the top singular vectors of parameters, and with increasing sparsity fewer directions are preserved. Even though prior work has remarked that it is possible to use low-rank approximations for neural networks~\citep{yu2017compressing}, and others have explicitly optimized for low-rank lottery tickets~\citep{wang2021pufferfish, schotthofer2022low}, we rather are pointing out that the magnitude pruning procedure seems to recover a low-rank approximation.

We also compute the singular vector agreement (SVA) between the masked model trajectory and the original unmasked model trajectory (diagonal of Eqn.~\ref{eqn:sva-matrix}). We see in Figure~\ref{fig:lottery-tickets} that there is no agreement between the bottom singular vectors at all, but there is still loose agreement in the top singular vectors. Thus, it seems the mask allows the dynamics of only the top singular vectors to remain similar, which we know are most important from the pruning analysis in Figure~\ref{fig:pruned-performance}.

Preserving top singular vectors by pruning seems like a natural outcome of large matrices, so as a control, we follow exactly the same protocol except we generate the mask randomly with the same layerwise sparsity. We can see in Figure~\ref{fig:lottery-tickets} that this results in much lower preservation of top singular vector dynamics, and also performs worse, as in \citep{frankle2020linear}. It would not be surprising that random pruning is worse if simply evaluated at the end of training, but masking is applied quite early in training at epoch 4 of 164 long before convergence, so it's striking that the network now fails to learn further even though it is far from convergence. We interpret this as evidence that the mask has somehow cut signal flow between layers, so it is now impossible for the network to learn further, while magnitude pruning and rewinding still allows signals to pass that eventually become important.

\subsection{Spectral Dynamics and Linear Mode Connectivity}\label{sec:lmc}

We come to the final phenomenon that we seek to describe: linear mode connectivity. Linear mode connectivity (LMC) is the property that one can interpolate linearly between two different minima in weight space and every parameter set along that path performs well, which gives the impression that the loss surface of neural networks is somehow convex despite its theoretical nonconvexity. This was first demonstrated in small networks with the same initialization~\citep{nagarajan2019uniform}, then expanded to larger networks and connected to lottery tickets~\citep{frankle2020linear, paul2022unmasking}. \citet{entezari2021role} first conjecture that two arbitrary minima show LMC up to permutation, and demonstrate it in simple models. This was expanded to wide models~\citep{ainsworth2022git, jordan2022repair, qu2024rethink}, and can be proven in various ways~\citep{kuditipudi2019explaining, brea2019weight, simsek2021geometry, ferbach2023proving}, but it does not hold for standard models~\citep{qu2024rethink}. LMC has also been exploited for model-averaging and performance gains~\citep{wortsman2022model, ilharco2022editing, rame2022diverse}. Still despite all of this work, we lack a description for why LMC occurs. In particular: why is there a convex, high dimensional~\citep{yunis2022convexity} basin that models find shortly in training~\citep{frankle2020linear}, or after pretraining~\citep{neyshabur2020being, sadrtdinov2023stay}? We do not answer this question in full, but find an interesting view through the singular vectors.

\subsubsection{Linear Mode Connectivity Correlates with Top Singular Vector Agreement}

As we saw earlier directional convergence of top singular vectors in Figure~\ref{fig:singular-vector-agreement}, it suggests the dynamics of those components are more stable, so we might expect mode-connected solutions to share these components. To examine this, we plot agreement between the singular vectors of the weight matrices at either endpoint of branches: 
\begin{align*}
 W^{(1)}(T) &= \sum_{j}^R \sigma_j(T) u_j(T) v_j(T)^\top\enspace,\\
 W^{(2)}(T) &= \sum_{k}^R \sigma'_k(T) u'_k(T) {v'_k}(T)^\top\enspace,   
\end{align*}
spawned from the same initialization in training. If the branches are split from an initialization on a trunk trajectory $W(t)$, we call $t$ the split point or epoch. We visualize the diagonal of $\lvert \langle u_j(T) v_j(T)^\top, u'_k(T) {v'_k}(T)^\top \rangle \rvert_{jk}$ vs.\ split epoch, where the absolute value is taken to ignore sign flips in SVD computation.

\begin{figure}[!t]
  \centering
  \begin{subfigure}[b]{0.24\linewidth}
    \centering
    \includegraphics[width=\linewidth]{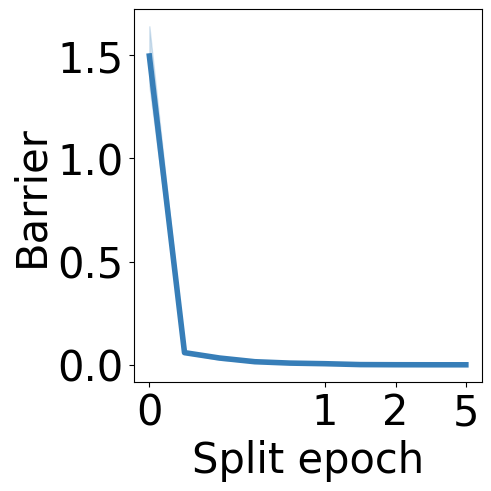}
    \\
    \includegraphics[width=\linewidth]{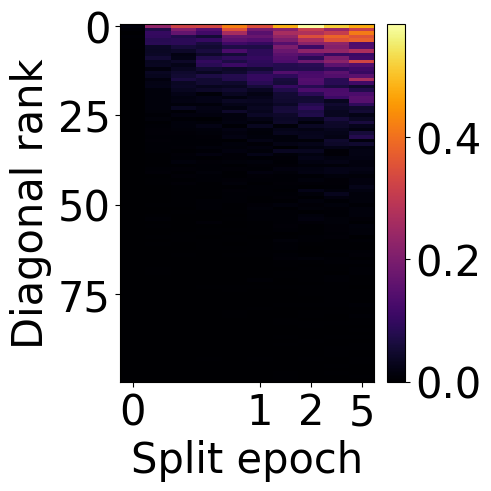}
    \\
    \includegraphics[width=\linewidth]{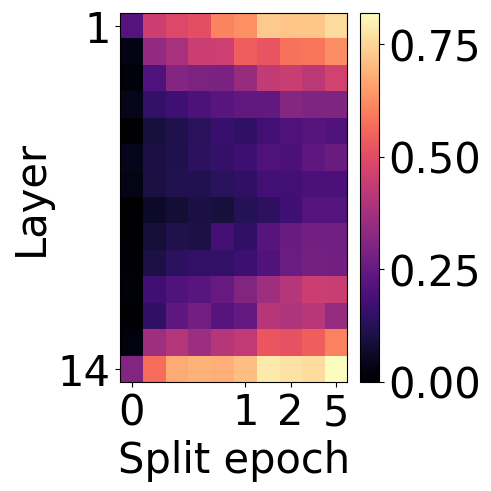}
    \caption{VGG}
  \end{subfigure}
  \begin{subfigure}[b]{0.24\linewidth}
    \centering
    \includegraphics[width=\linewidth]{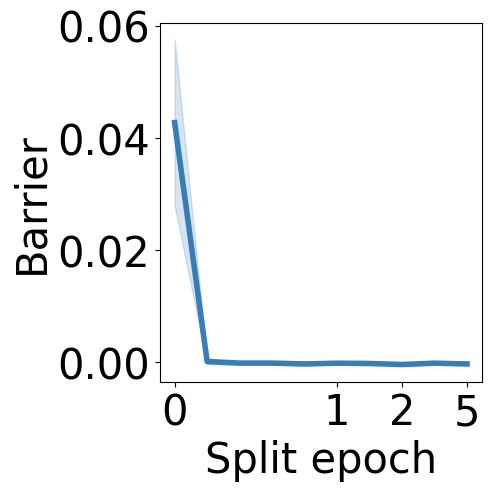}
    \\
    \includegraphics[width=\linewidth]{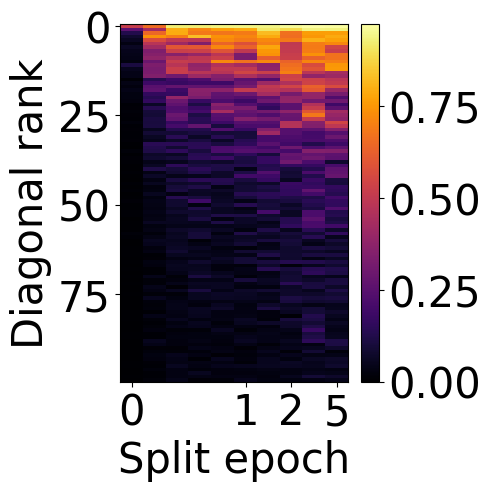}
    \\
    \includegraphics[width=\linewidth]{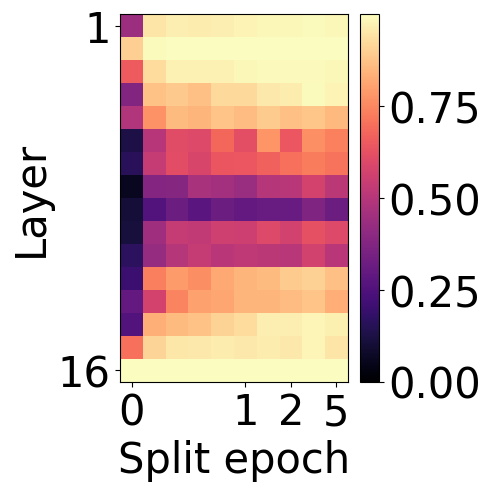}
    \caption{UNet}
  \end{subfigure}
  \begin{subfigure}[b]{0.24\linewidth}
    \centering
    \includegraphics[width=\linewidth]{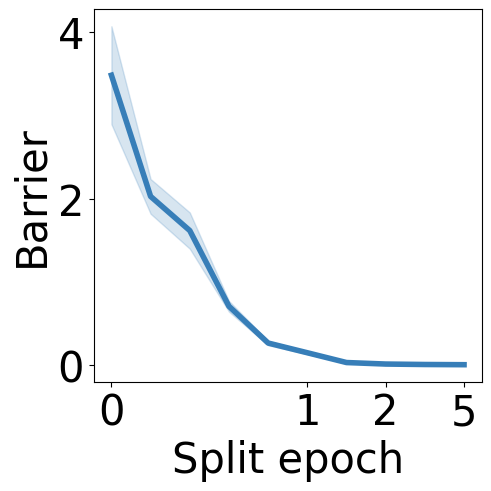}
    \\
    \includegraphics[width=\linewidth]{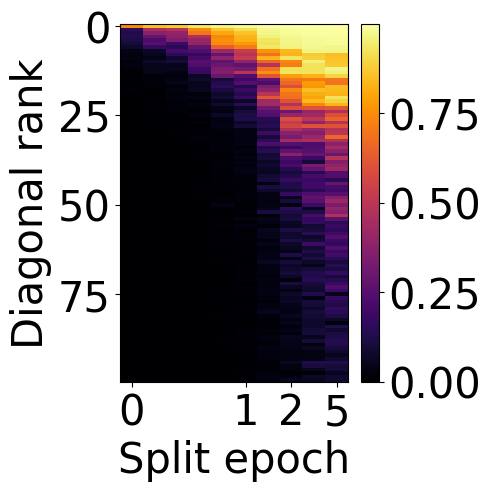}
    \\
    \includegraphics[width=\linewidth]{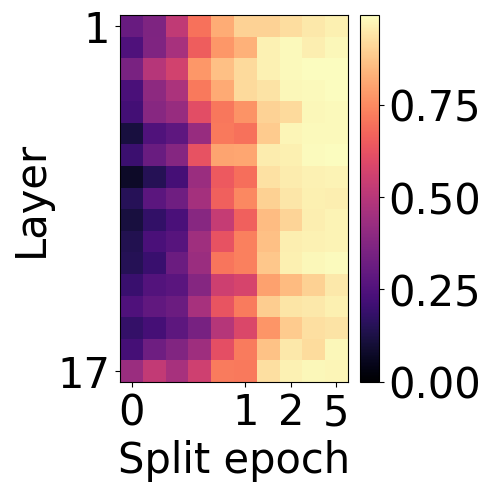}
    \caption{LSTM}
  \end{subfigure}
  \begin{subfigure}[b]{0.24\linewidth}
    \centering
    \includegraphics[width=\linewidth]{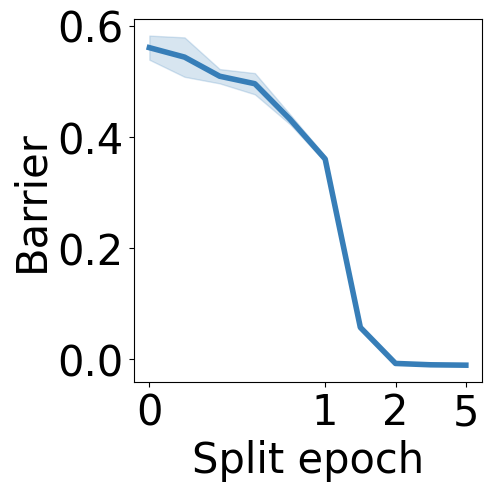}
    \\
    \includegraphics[width=\linewidth]{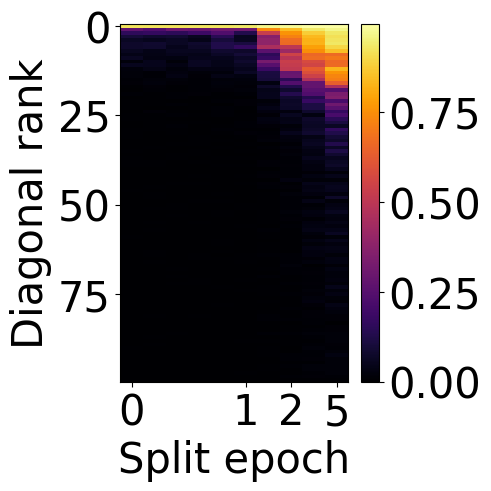}
    \\
    \includegraphics[width=\linewidth]{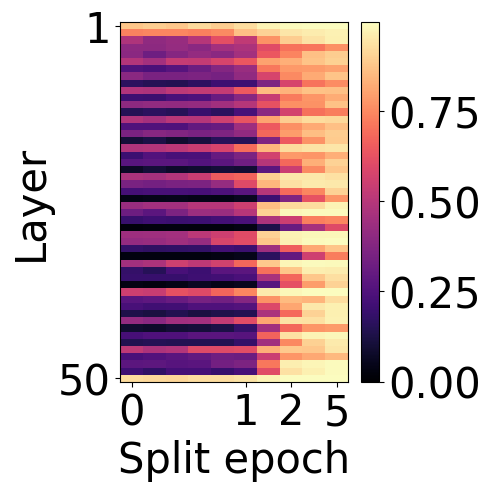}
    \caption{Transformer}
  \end{subfigure}
  \caption{\textbf{Top row:} Barrier size vs.\ split step. \textbf{Middle row:} singular vector agreement for a single matrix parameter between branch endpoints that share a common trunk. \textbf{Bottom row:} summary statistic for singular vector agreement across layers vs.\ split step. We see that as models exhibit LMC, they also share top singular vectors.}
  \label{fig:lmc-agreement}
\end{figure}

\begin{figure}[!t]
  \centering
  \begin{subfigure}[b]{0.24\linewidth}
    \centering
    \includegraphics[width=\linewidth]{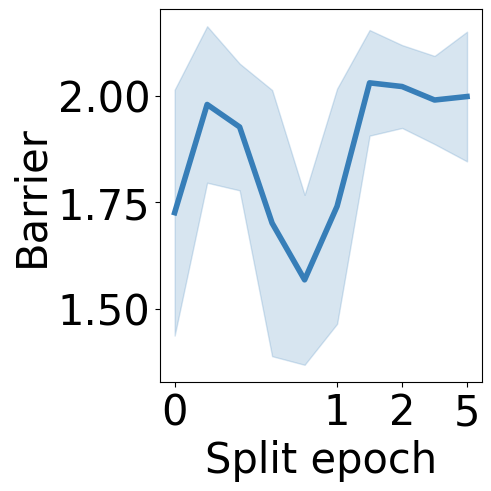}
    \\
    \includegraphics[width=\linewidth]{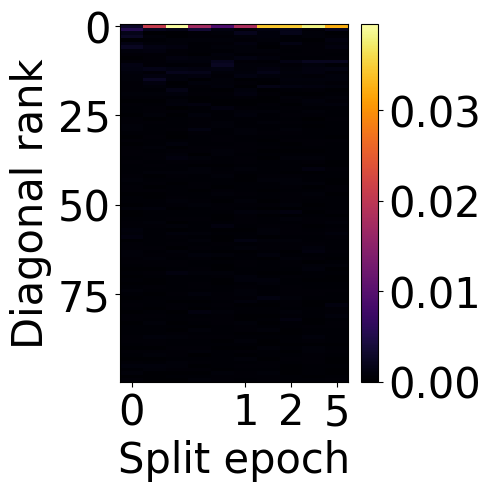}
    \\
    \includegraphics[width=\linewidth]{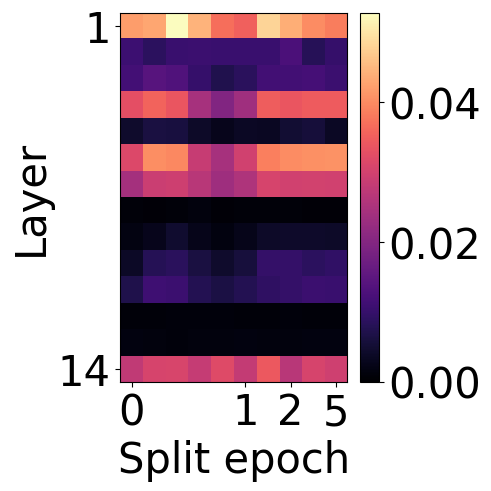}
    \caption{VGG}
  \end{subfigure}
  \begin{subfigure}[b]{0.24\linewidth}
    \centering
    \includegraphics[width=\linewidth]{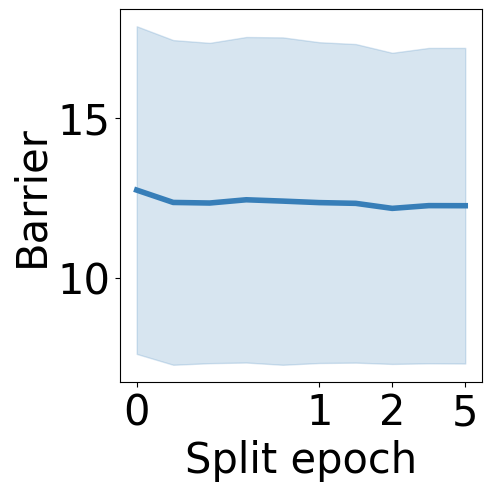}
    \\
    \includegraphics[width=\linewidth]{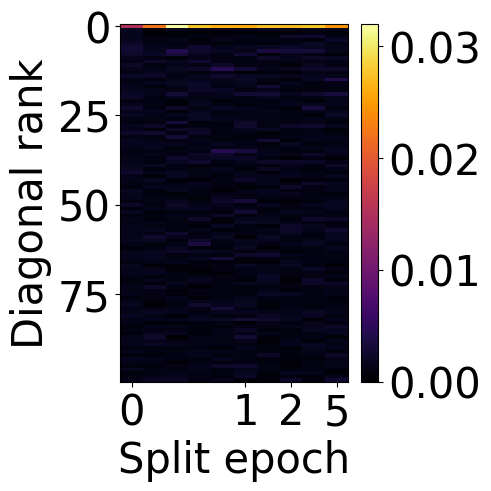}
    \\
    \includegraphics[width=\linewidth]{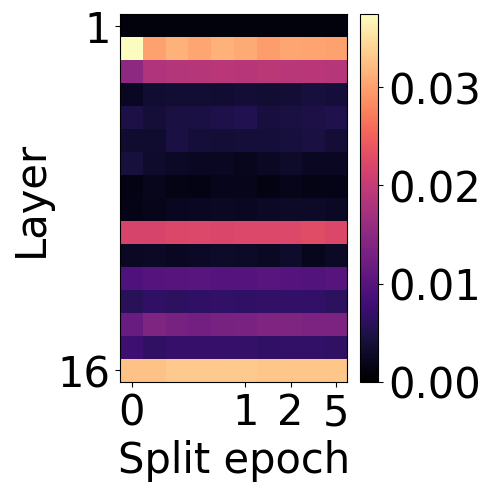}
    \caption{UNet}
  \end{subfigure}
  \begin{subfigure}[b]{0.24\linewidth}
    \centering
    \includegraphics[width=\linewidth]{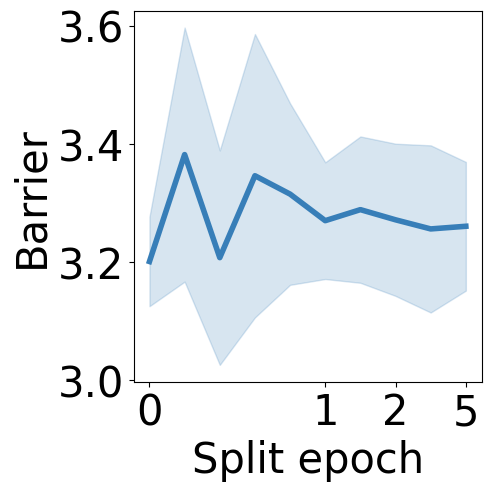}
    \\
    \includegraphics[width=\linewidth]{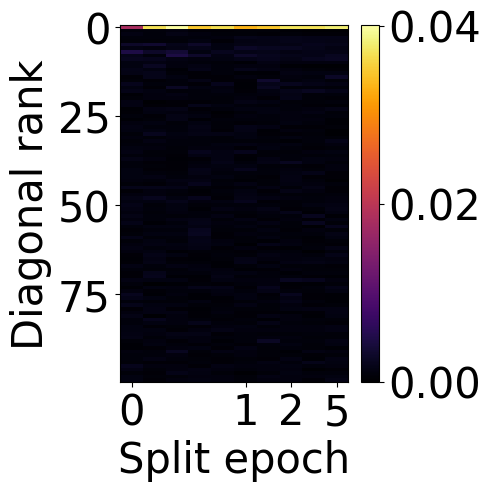}
    \\
    \includegraphics[width=\linewidth]{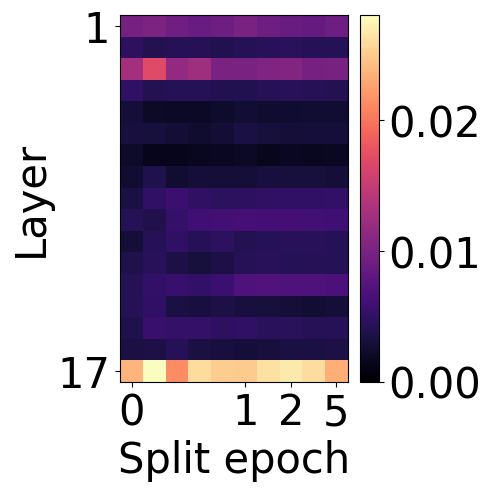}
    \caption{LSTM}
  \end{subfigure}
  \begin{subfigure}[b]{0.24\linewidth}
    \centering
    \includegraphics[width=\linewidth]{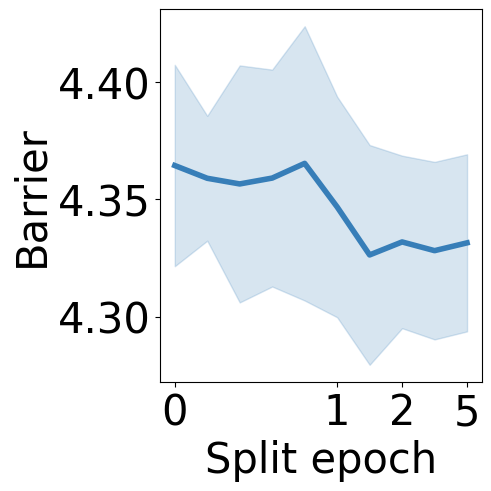}
    \\
    \includegraphics[width=\linewidth]{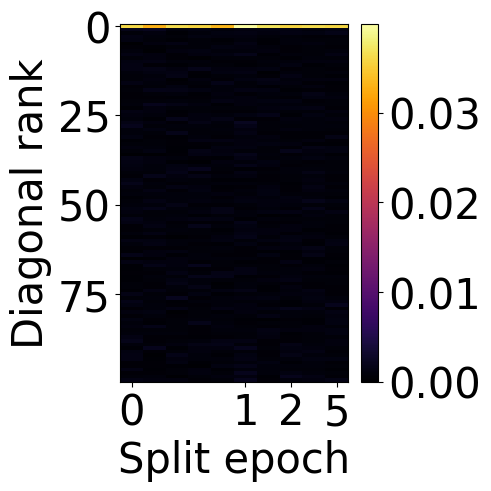}
    \\
    \includegraphics[width=\linewidth]{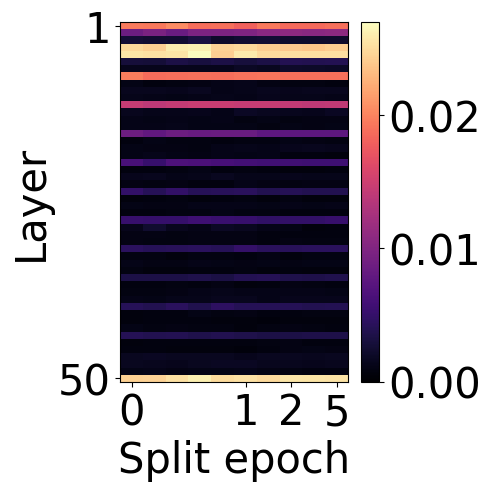}
    \caption{Transformer}
  \end{subfigure}
  \caption{\textbf{Top row:} Barrier size vs.\ split step. \textbf{Middle row:} singular vector agreement for a single matrix parameter between branch endpoints that do not share a common trunk, but do share split time and branch data order. \textbf{Bottom row:} summary statistic for singular vector agreement across layers. We see that when branches do not share a common trunk, there is neither LMC nor singular vector agreement, even though the optimization is otherwise the same.}
  \label{fig:lmc-cross-agreement}
\end{figure}

To remind the reader, LMC only occurs after a small amount of training time has passed. Too early and the final models of each branch will show a bump, or barrier, in the loss surface along the linear interpolation \citep{frankle2020linear}. To measure this precisely, we use the definition from \citet{neyshabur2020being}, which is the maximum deviation from a linear interpolation in the loss, an empirical measure for convexity in this linear direction. When this deviation is 0, we consider the checkpoints to exhibit LMC. Please see Appendix~\ref{app:lmc-details} for details on the calculation. Given evidence in Figure~\ref{fig:pruned-performance} that top components are the most important for prediction, and that top components become stable before training has finished, it is plausible that LMC is connected to the stability of top singular vectors in the later portion of training.

This would mean that checkpoints that do not exhibit the LMC property should not share top singular vectors, while checkpoints that do exhibit the LMC property should share top singular vectors. We see in Figure~\ref{fig:lmc-agreement} that this is the case across models and tasks, where the alignment between endpoints is much stronger in top singular vectors. We also see no LMC and poor agreement in top components between branches that have initializations from different trunk trajectories, but with the same split epoch $t$ and the same branch data order in Figure~\ref{fig:lmc-cross-agreement}. Thus, these top directions are not a unique property of the architecture and data, but rather are dependent on initialization. It is notable that concurrent work~\citep{ito2024analysis} arrives at a similar conclusion: permutation solvers between optima match top singular vectors. Though the conclusions are similar, their experiments are primarily conducted on smaller scale settings, and only for permutation matching at the end of training. Here we connect these observations to the optimization behavior of networks throughout training.

\subsubsection{Perturbing Breaks Linear Mode Connectivity and Singular Vector Agreement Simultaneously}

\begin{figure}[!t]
  \centering
  \begin{subfigure}[b]{0.24\linewidth}
    \centering
    \includegraphics[width=\linewidth]{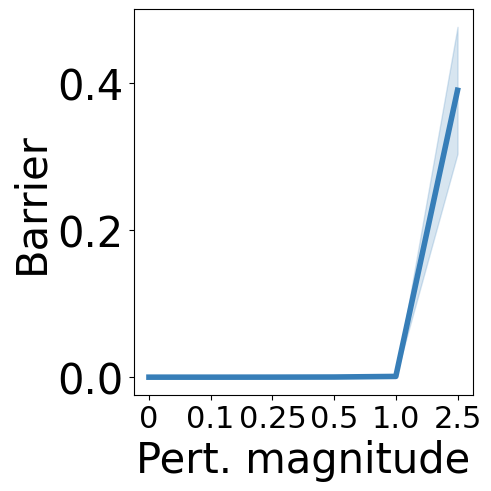}
    \\
    \includegraphics[width=\linewidth]{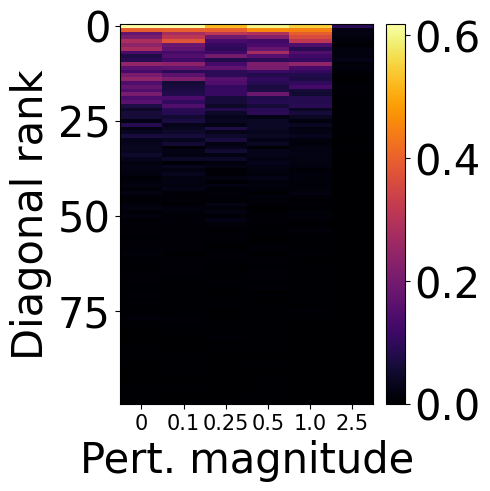}
    \\
    \includegraphics[width=\linewidth]{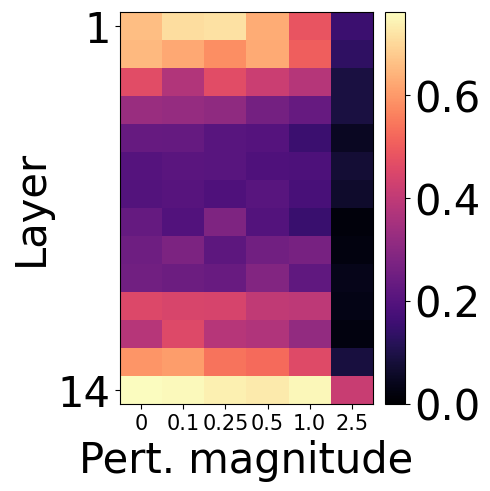}
    \caption{VGG}
  \end{subfigure}
  \begin{subfigure}[b]{0.24\linewidth}
    \centering
    \includegraphics[width=\linewidth]{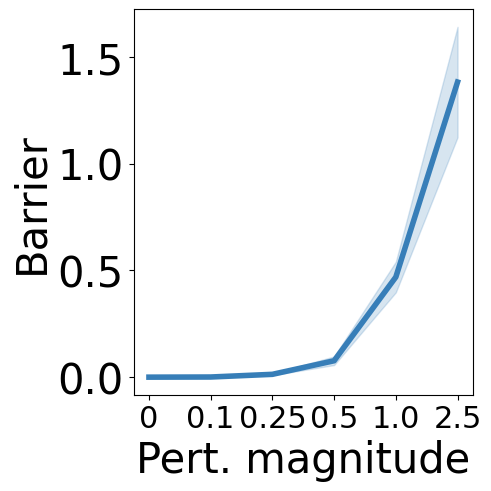}
    \\
    \includegraphics[width=\linewidth]{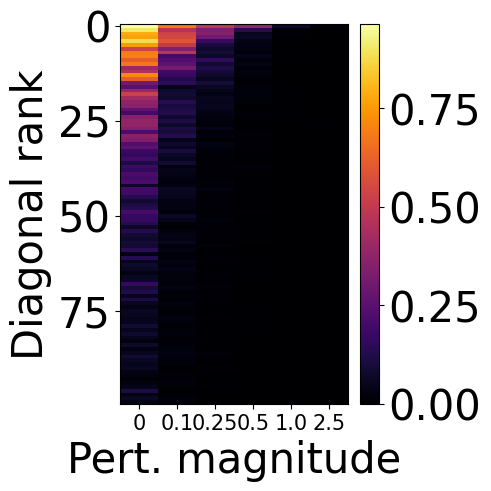}
    \\
    \includegraphics[width=\linewidth]{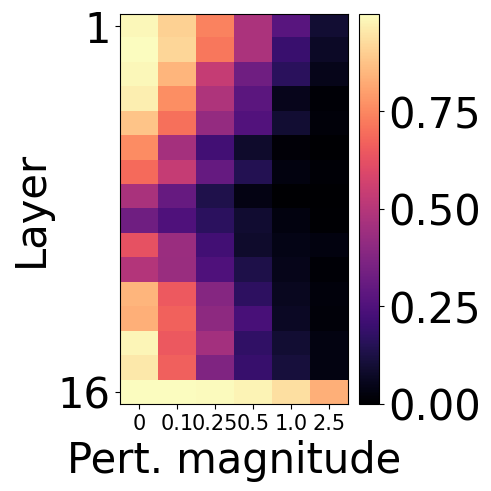}
    \caption{UNet}
  \end{subfigure}
  \begin{subfigure}[b]{0.24\linewidth}
    \centering
    \includegraphics[width=\linewidth]{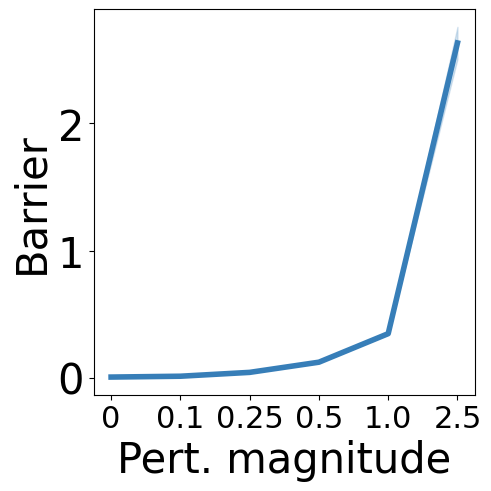}
    \\
    \includegraphics[width=\linewidth]{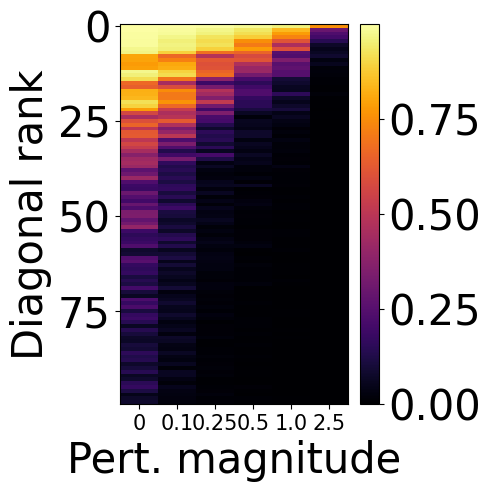}
    \\
    \includegraphics[width=\linewidth]{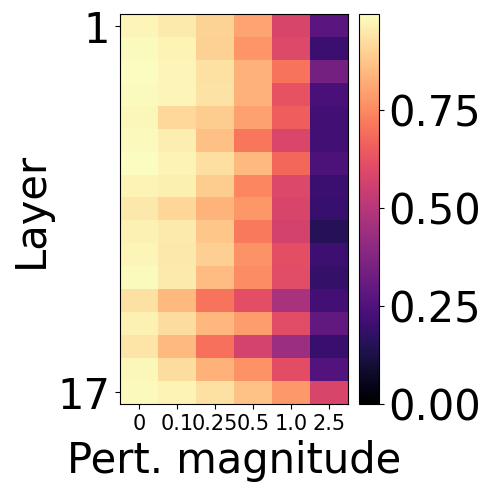}
    \caption{LSTM}
  \end{subfigure}
  \begin{subfigure}[b]{0.24\linewidth}
    \centering
    \includegraphics[width=\linewidth]{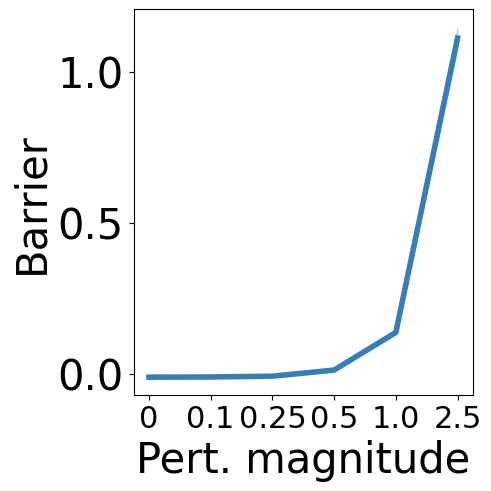}
    \\
    \includegraphics[width=\linewidth]{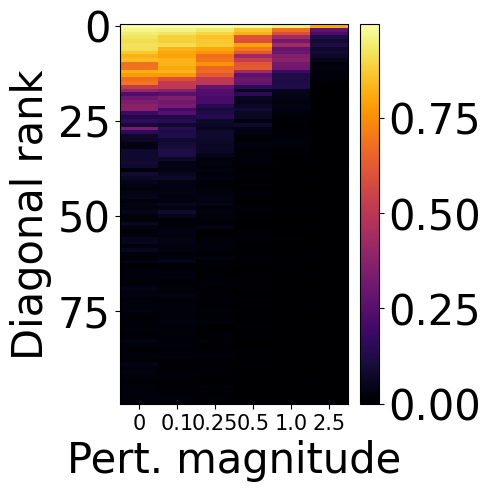}
    \\
    \includegraphics[width=\linewidth]{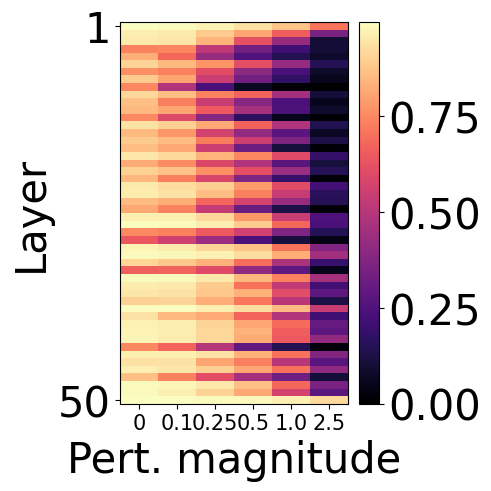}
    \caption{Transformer}
  \end{subfigure}
  \caption{\textbf{Top row:} Barrier size vs.\ perturbation magnitude. \textbf{Middle row:} singular vector agreement for a single matrix parameter between branch endpoints vs.\ perturbation magnitude. \textbf{Bottom row:} summary statistic for singular vector agreement across layers with perturbation magnitude. We see that whereas without perturbation models would exhibit LMC after training, with increasing perturbations the LMC property disappears simultaneously with the agreement in top singular vectors.}
  \label{fig:lmc-pert-agreement}
\end{figure}

To make the connection between top singular vectors and LMC even tighter, we intervene in the normal training process. If we add random perturbations to destabilize the components that will become the top components long before they have converged, and if singular vector agreement is tied to LMC, we would like to see that final models no longer exhibit the LMC property. Indeed this is the case. In Figure~\ref{fig:lmc-pert-agreement}, when increasingly large random perturbations are applied, the barrier between final checkpoints increases and the LMC behavior disappears. Please see Appendix~\ref{app:experimental-details} for details. In addition, the previously-strong singular vector agreement disappears simultaneously. Thus it seems this agreement is tied to linear mode connectivity.

We speculate that, due to the results in Figure~\ref{fig:pruned-performance} that show the top half of the SVDs are much more critical for performance, if these components are shared then interpolating will not affect performance much. Rather, interpolation will eliminate the orthogonal bottom components which may only make a minor impact on performance. If however the top components are not shared, then interpolating between two models will remove these components, leading to poor performance in between. Such observations may help in explaining the utility of pretraining~\citep{neyshabur2020being}, weight averaging~\citep{rame2022diverse, wortsman2022model, ilharco2022editing} or the use of LoRA~\citep{huh2022low} to replace full finetuning.

\section{Discussion}

We provide an empirical perspective to understand deep learning through the lens of SVD dynamics. We first note a tendency toward rank minimization on a small scale in grokking, then expand these findings to practical networks and tasks. In addition we find that weight decay, though it explicitly penalizes norm, implicitly promotes this low-rank bias. We also show through the developed analysis that generalization and memorization differ in the rank and alignment of solutions found by optimization.

We go beyond remarks on generalization and show that magnitude pruning for lottery tickets acts similarly to low-rank pruning, and linear mode connectivity coincides with the sharing of top singular vectors between checkpoints. Low rank models have more efficient inference, and finetuning from pretrained models always operates in the LMC regime~\citep{neyshabur2020being}, thus this understanding can lead to more efficient inference and compression and possibly stronger optimization practices.

While a comprehensive theory for our results remains elusive, these observations can act as a platform for a deeper understanding of deep learning. Notably, the observed spectral dynamics appear consistent across diverse settings, even without restrictive assumptions like balanced initialization, linearity, or small weight scales. This suggests a common underlying mechanism.

On the empirical side, several interesting problems present themselves. Interpretability of neural networks is a growing area of research~\citep{nanda2023progress}, and there already exist early efforts to interpret singular vectors of convolutional weights~\citep{praggastis2022svd}. There may also be connections to other unexplained phenomena such as double descent~\citep{belkin2019reconciling, nakkiran2021deep, davies2022unifying} or adversarial examples~\citep{szegedy2013intriguing, ilyas2019adversarial, hendrycks2021natural}. The solutions to these problems may help us diagnose deployment risks in the wild, and ideally help to design better algorithms. We believe our results contribute another step along this path.

\section*{Acknowledgements}

DY would like to thank the following people for helpful discussion at many different stages: Wei Hu for a conversation on connecting LMC to rank dynamics in deep matrix factorization; Jonathan Frankle for very early support and periodic discussions on aiming to understand the source of LMC; Shubham Toshniwal and Shane Settle for support in following this line of work in the early stages; Mikhail Belkin, Tengyu Ma, and Rob Nowak for early conversations and encouragement on studying rank minimization in small-scale cases; Sudarshan Babu, Haochen Wang, Takuma Yoneda, and Xiao Zhang for discussions and enthusiasm on resuming this work when it had been paused; Sam Buchanan for similar enthusiasm on resuming the work and very helpful comments on presentation; Anmol Kabra for suggestions on additional experiments and visualizations to round out the work; and David McAllester for very helpful comments on presentation. DY would also like to thank Adam Bohlander for his work in maintaining the TTIC cluster, which made the many experiments in this paper possible. This material is based upon work supported by the National Science Foundation Graduate Research Fellowship Program under Grant No. 1754881.  KKP was supported through the NSF TRIPOD Institute on Data, Economics, Algorithms and Learning (IDEAL) and other awards from DARPA and NSF.

\newpage
\bibliography{main}
\bibliographystyle{icml2024}
\flushcolsend

\newpage
\appendix
\onecolumn

\section{Experimental Details}\label{app:experimental-details}

For all experiments, we use 3 random seeds and average all plots over those 3. This is relatively small, but error bars tend to be very tight, and due to the high volume of runs required for this work we lack the resources to run much more.

In order to compute alignment we consider only pairs of layers that directly feed into each other, and ignore the influence of residual connections so as to cut down on the number of comparisons. Specifics on individual architectures are given below.

\subsection{Image Classification with VGG}\label{app:imgclass-details}

We train a VGG-16~\citep{simonyan2014very} on CIFAR-10~\citep{krizhevsky2009learning} for 164 epochs, following hyperparameters and learning rate schedule in~\citep{frankle2020linear}, but without data augmentation. For the optimizer we use SGD with batch size 128, initial learning rate 0.1 and momentum of 0.9. We also decay the learning rate 3 times by a factor of 10 at epoch 82, epoch 120, and finally at epoch 160. We also use a minor amount of weight decay with coefficient 0.0001.

VGG-16 uses ReLU activations and batch normalization~\citep{ioffe2015batch}, and includes both convolutional and linear layers. For linear layers we simply compute the SVD of the weight matrix. For convolutional layers, the parameters are typically stored as a 4D tensor of shape $(c_\text{out}, c_\text{in}, h, w)$ for the output channels, input channels, height and width of the filters respectively. As the filters compute a transformation from each position and input channel to an output channel, we compute the SVD of the flattened tensor $(c_\text{out}, c_\text{in} \cdot h \cdot w)$, which maps all inputs to outputs, similar to \citet{praggastis2022svd}. This is not the SVD of the entire transformation of the feature map to the next feature map, but rather the transformation from a set of adjacent positions to a particular position in the next layer. For the individual SV evolution plot, we use the 12th convolutional layer.

In order to compute alignment of bases between consecutive convolutional layers, $V_{i+1}^\top U_i$ we need to match the dimensionality between $U_i$ and $V_{i+1}$. For convolutional layers we are presented with a question as to how to handle the spatial dimensions $h$ and $w$ as naively the input dimension of the next layer will be a factor of $h\cdot w$ larger dimension. We experimented with multiple cases, including aligning at each spatial position individually or averaging over the alignment at all spatial positions, and eventually settled at aligning the output of one layer to the center spatial input of the next layer. That is, for a 3x3 convolution mapping to a following 3x3 convolution, we compute the alignment only for position (1,1) of the next layer. This seemed reasonable to us as on average the edges of the filters showed poorer alignment overall. For the individual alignment plot, we use the alignment between the 11th and 12th convolutional layers at the center spatial position of the 12th convolutional layer.

\subsection{Image Generation with UNets}\label{app:imggen-details}

We train a UNet~\citep{ronneberger2015u} diffusion model~\citep{sohl2015deep, ho2020denoising} on MNIST~\citep{lecun1998mnist} generation. We take model design and hyperparameters from~\citep{wang2020ml}. In particular we use a 4-layer residual UNet and train with AdamW~\citep{loshchilov2018fixing} with batch size 128, and learning rate of 0.0003 for 100 epochs. This model uses swish~\citep{ramachandran2017searching} activations and a combination of linear and convolutional, as well as transposed convolutional layers.

Computing SVDs and alignment is similar to the image classification case described above, except in the case of the transposed convolutions where an extra transpose of dimensions is needed as parameters are stored with the shape $(c_\text{in}, c_\text{out}, h, w)$. For the individual SV evolution plot, we use the 3rd convolutional layer. For the alignment plot, we use the alignment between the 3rd and 4th convolutional layers at the center spatial position of the 4th convolutional layer.

\subsection{Speech Recognition with LSTMs}\label{app:speech-details}

We train a bidirectional LSTM~\citep{hochreiter1997flat} for automatic speech recognition on LibriSpeech~\citep{panayotov2015librispeech}. We tune for a simple and well-performing hyperparameter setting. We use AdamW~\citep{loshchilov2018fixing} with batch size 32, learning rate 0.0003 and weight decay 0.1 for 50 epochs. We also use a cosine annealing learning rate schedule from 1 to 0 over the entire 50 epochs.

The LSTM only has matrix parameters and biases, so it is straightforward to compute SVDs of the matrices. For individual SV evolution plots, we plot the 3rd layer input parameter. In the case of alignment, we make a number of connections: first down depth for the input parameters, then connecting the previous input parameter to the current hidden parameter in both directions, then connecting the previous hidden parameter to the current input parameter. In particular the LSTM parameters are stored as a stack of 4 matrices in PyTorch, and we find alignment is highest for the "gate" submatrix, so we choose that for all plots. For the individual layer alignment, we plot alignment between the 3rd and 4th layer input parameters.

\subsection{Language Modeling with Transformers}\label{app:language-details}

We train a Transformer~\citep{vaswani2017attention} language model on Wikitext-103~\citep{merity2016pointer}. We base hyperparameter choices on the Pythia suite~\citep{biderman2023pythia}, specifically the 160 million parameter configuration with sinusoidal position embeddings, 12 layers, model dimension 768, 12 attention heads per layer, and hidden dimension 768. We use AdamW~\citep{loshchilov2018fixing} with batch size 256, learning rate 0.0006 and weight decay 0.1. We use a context length of 2048 and clip gradients to a maximum norm of 1. We also use a learning rate schedule with a linear warmup and cosine decay to 10\% of the learning rate, like \citet{biderman2023pythia}.

For SVDs, for simplicity we take the SVD of the entire $(3d_\text{model}, d_\text{model})$ parameter that computes queries, keys and values from the hidden dimension inside the attention layer, without splitting into individual heads. This is reasonable as the splitting is done after the fact internally. We also take the SVD of the output parameters, and linear layers of the MLPs, which are 2 dimensional matrices. For the individual SV evolution plot, we plot the SVs of $W_1$ of the 8th layer MLP

For alignment, we consider the alignment of $W_Q$ and $W_K$ matrices, $W_V$ and $W_O$ matrices, computing alignment between heads individually then averaging over all heads. We also consider the alignment between $W_O$ and $W_1$ of the MLP block, between $W_1$ and $W_2$ of the MLP block, and between $W_2$ and the next attention layer. For the individual layer alignment, we plot alignment between $W_1$ and $W_2$ of the 8th layer MLP.

\subsection{Weight Decay Experiments}

\begin{figure}[!t]
  \centering
  \begin{subfigure}[b]{0.24\linewidth}
    \centering
    \includegraphics[width=\linewidth]{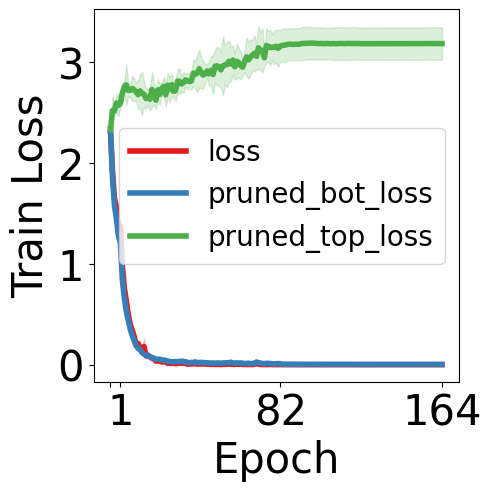}
    \\
    \includegraphics[width=\linewidth]{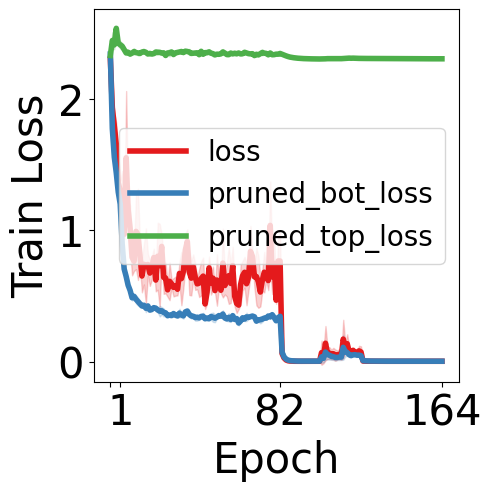}
    \\
    \includegraphics[width=\linewidth]{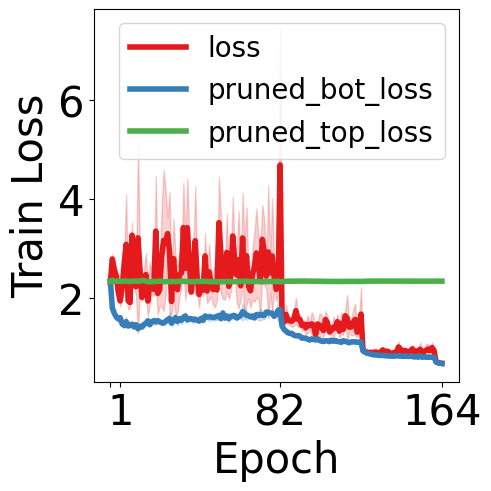}
    \\
    \includegraphics[width=\linewidth]{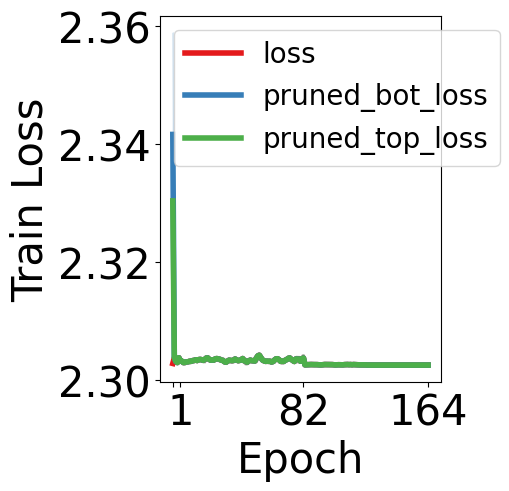}
    \caption{VGG}
  \end{subfigure}
  \begin{subfigure}[b]{0.24\linewidth}
    \centering
    \includegraphics[width=\linewidth]{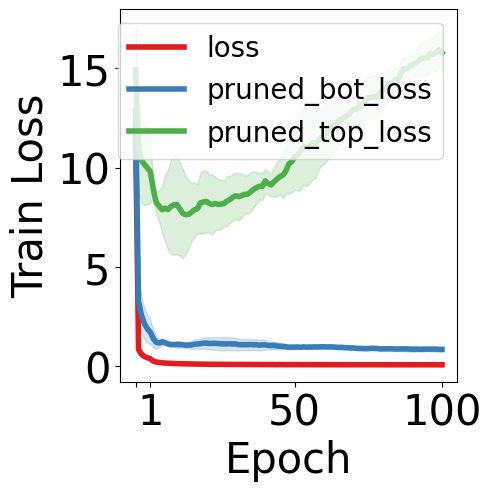}
    \\
    \includegraphics[width=\linewidth]{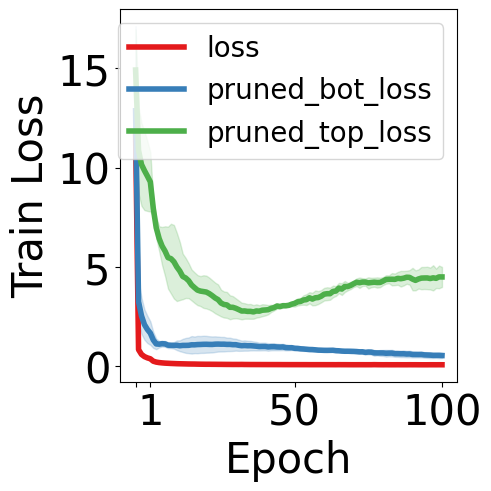}
    \\
    \includegraphics[width=\linewidth]{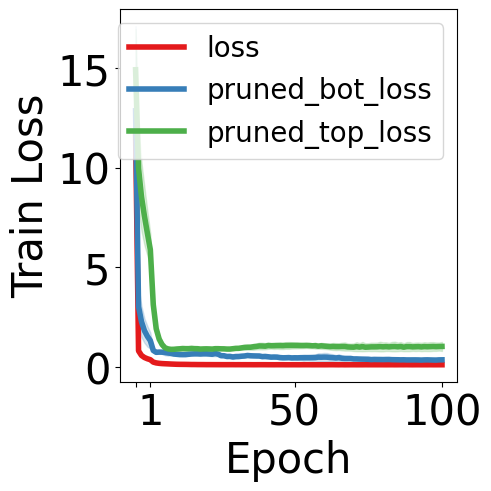}
    \\
    \includegraphics[width=\linewidth]{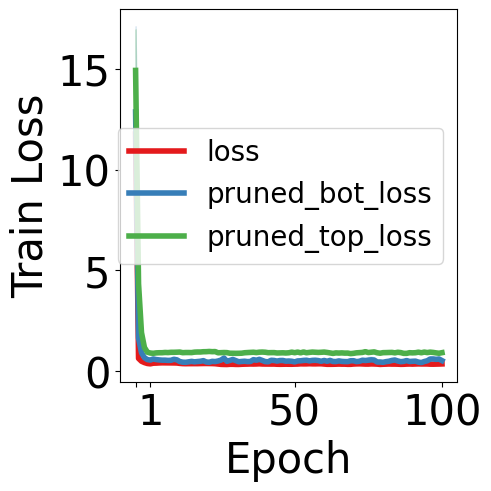}
    \caption{UNet}
  \end{subfigure}
  \begin{subfigure}[b]{0.24\linewidth}
    \centering
    \includegraphics[width=\linewidth]{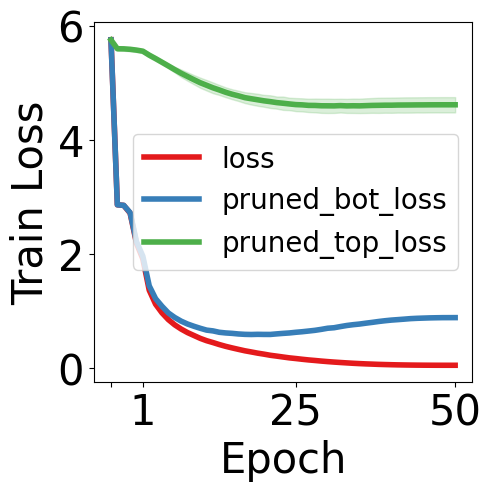}
    \\
    \includegraphics[width=\linewidth]{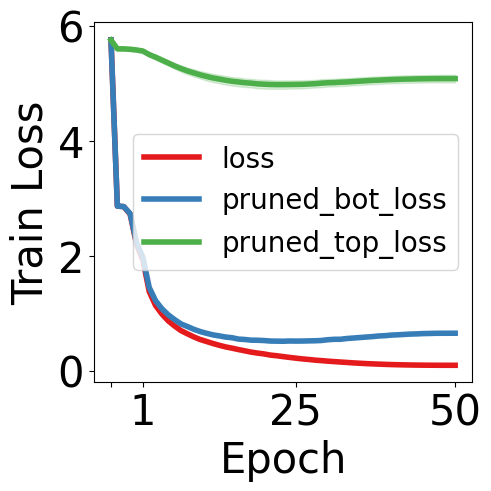}
    \\
    \includegraphics[width=\linewidth]{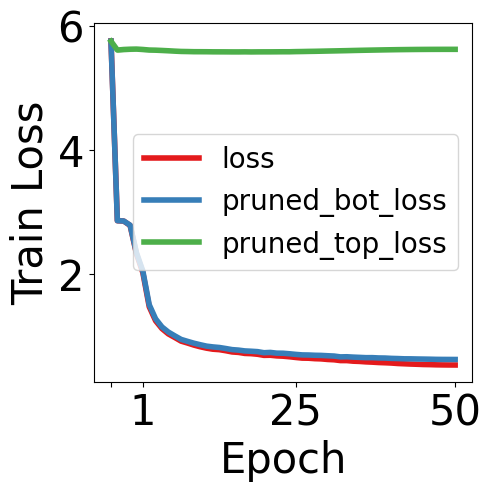}
    \\
    \includegraphics[width=\linewidth]{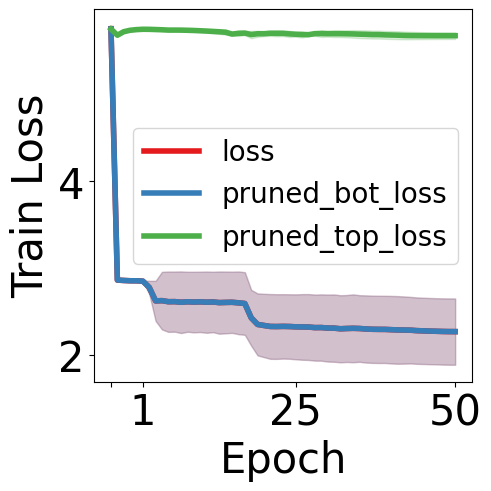}
    \caption{LSTM}
  \end{subfigure}
  \begin{subfigure}[b]{0.24\linewidth}
    \centering
    \includegraphics[width=\linewidth]{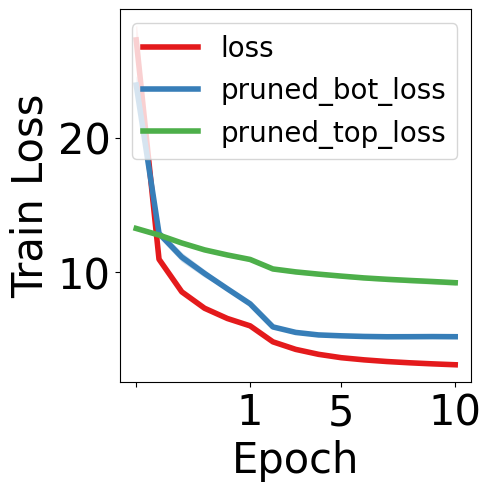}
    \\
    \includegraphics[width=\linewidth]{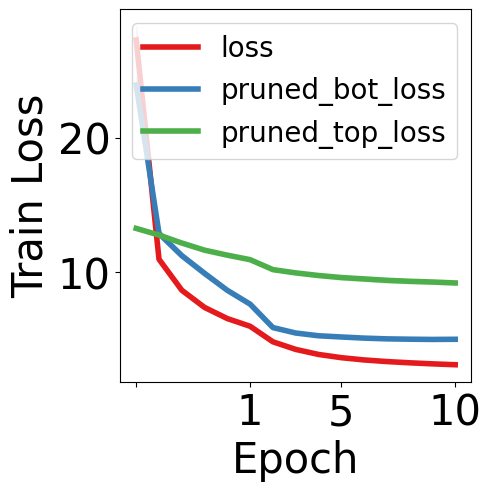}
    \\
    \includegraphics[width=\linewidth]{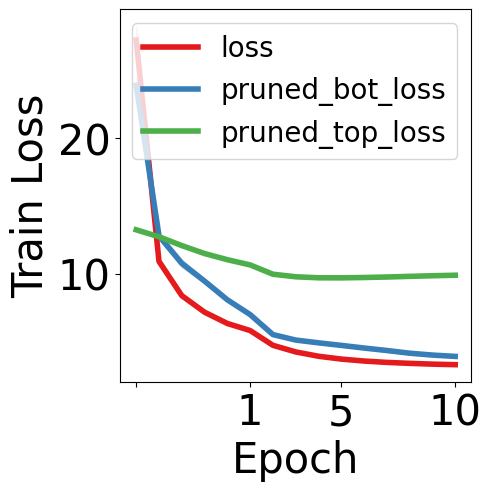}
    \\
    \includegraphics[width=\linewidth]{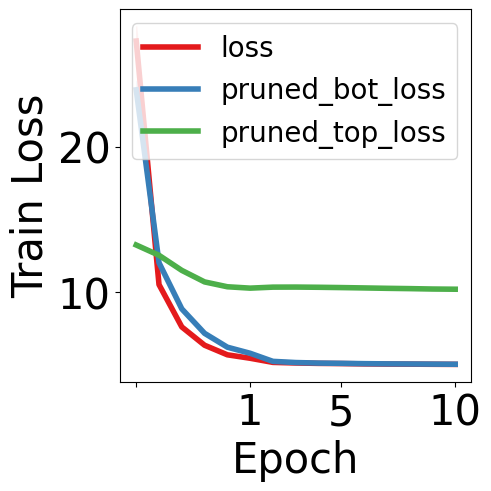}
    \caption{Transformer}
  \end{subfigure}
  \caption{Training loss over time, where the rows use differing amounts of weight decay. From top to bottom, for VGG we use coefficients $\{ 0, 0.001, 0.01, 0.1 \}$, while for other networks we use coefficients $\{0, 0.1, 1, 10\}$. We see that it is still possible to achieve low training loss under high weight decay, and as we increase the amount of weight decay, the gap between pruned and unpruned parameters closes, lending support to the idea that the parameters become lower rank.}
  \label{fig:weight-decay-performance}
\end{figure}

All tasks are trained in exactly the same fashion as mentioned previously, with increasing weight decay in the set $\{0, 0.0001, 0.001, 0.01, 0.1, 1.0, 10.0\}$. For ease of presentation we consider a subset of settings across tasks. In Figure~\ref{fig:weight-decay-performance} we include trained model performance and pruned model performance to show that, even with high levels of weight decay, models do not entirely break down. More so, the approximation of the pruned model to the full model gets better with higher weight decay.

\subsection{Grokking Experiments}

We mostly follow the settings and architecture of \citet{nanda2023progress}, except we use sinusoidal positional encodings instead of learned.

For the slingshot case we follow hyperparameter settings in \citet{thilak2022slingshot}, Appendix B except with the 1-layer architecture from \citet{nanda2023progress} instead of the 2-layer architecture specified. W perform addition modulo 97. The original grokking plot in \citet{thilak2022slingshot} appears much more dramatic as it log-scales the x-axis, which we do not do here for clarity.

\subsection{Random Label Experiments}

We train a 4-layer MLP on CIFAR10~\citep{krizhevsky2009learning} with either completely random labels, or the true labels. We use SGD with momentum of 0.9 and constant learning rate of 0.001, and train for 300 epochs to see the entire trend of training. The major difference to the setting of \citet{zhang2021understanding} is the use of a constant learning rate, as their use of a learning rate schedule might conflate the results.

\subsection{Magnitude Pruning Experiments}

We use the same VGG setup as described previously. In this case we train til the end, then compute a global magnitude mask. To do this we flatten all linear and convolutional weights into a single vector, except for the last linear layer, and sort by magnitude. Then we keep the top 5\% of weights globally, and reshape back to the layerwise masks. This results in different sparsity levels for different layers, so when generating the random masks, we use the per-layer sparsities that resulted from the global magnitude mask.

To retrain the network, we rewind to epoch 4, then continue training with the mask, always setting other weights and their gradients to 0. We average all results over 3 random seeds.

\subsection{LMC Experiments}\label{app:lmc-details}

We save 5 evenly-spaced checkpoints in the first epoch, as well as at the end of the next 4 epochs for 10 intializations in total. We train 3 trunks, and split 3 branches from each trunk for a total of 9 branches which we average all plots over.

Following \citet{neyshabur2020being}, we compute the barrier between checkpoints as follows: given $W^{(1)}(T)$ and $W^{(2)}(T)$ that were branched from $W(t)$ we compute
\begin{equation}
    b(t) = (\max_{\alpha \in [0, 1]} \mathcal{L}((1-\alpha)W^{(1)}(T) + \alpha W^{(2)}(T)) - ((1-\alpha)\mathcal{L}(W^{(1)}(T)) + \alpha \mathcal{L}(W^{(2)}(T)))
\end{equation}
when this quantity is 0, we consider the checkpoints to exhibit LMC.

We recompute batch normalization parameters after interpolating for VGG-16, and group normalization parameters for the UNet, as these do not necessarily interpolate well~\citep{frankle2020linear}. We also compute singular vector agreement for the same parameter between either branch endpoint.

To plot the singular vector (dis)agreement and LMC between different modes, we make 11 evenly spaced measurements interpolating between branch endpoints that had the same split epoch, and the same branch seed, but different trunk initializations.

\subsection{Perturbed LMC Experiments}

We perturb all weights $W$ after the point of dynamics stability where we expect to see LMC at the end of training (epoch 4 is sufficiently late in all cases) using randomly sampled normal perturbations $\epsilon \sim \mathcal{N}(0, I)$ with $\lVert \epsilon \rVert = \eta \lVert W \rVert$ where $\eta \in \{0.0, 0.1, 0.25, 0.5, 1.0, 2.5\}$. We do not perturb the output layer, as this has a very substantial effect on the optimization. We also do not perturb the input layer for the Transformer as it is too computationally expensive for our resources.
\section{Limitations}

There are a few key limitations to our study. As mentioned, we lack the computational resources to run more than 3 random seeds per experiment, though we do find error bars to be quite tight in general (except for the generalization epoch in the grokking experiments). In addition, as discussed we ignore 1D parameters in the neural networks, which may be particularly crucial (especially normalization). In addition, due to computational constraints we do not consider alignment of layers across residual connections as this quickly becomes combinatorial in depth, thus there may be other interesting interactions that we do not observe. Finally, due to computational constraints we are unable to investigate results on larger models than the 12 layer Transformer, which may have different behavior.

\section{Compute Resources}

All experiments are performed on an internal cluster with on the order of 100 NVIDIA 2080ti GPUs or newer. All experiments run on a single GPU in less than 8 hours, though it is extremely helpful to parallelize across machines. We estimate that end-to-end it might take a few days on these resources to rerun all of the experiments in this paper. Additionally, the storage requirements for all of the checkpoints will take on the order of 5 terabytes.

\section{Code Sources}

We use PyTorch~\citep{paszke2019pytorch} and NumPy~\citep{harris2020array} for all experiments and Weights \& Biases~\citep{biewald2020experiment} for experiment tracking. We make plots with Matplotlib~\citep{Hunter:2007} and Seaborn~\citep{Waskom2021}. We also use HuggingFace Datasets~\citep{lhoest2021datasets} for Wikitext-103~\citep{merity2016pointer}.


\end{document}